\documentclass[acmsmall]{acmart}

\usepackage[utf8]{inputenc} % allow utf-8 input
\usepackage[T1]{fontenc}    % use 8-bit T1 fonts
\usepackage{hyperref}       % hyperlinks
\usepackage{url}            % simple URL typesetting
\usepackage{booktabs}       % professional-quality tables
\usepackage{amsfonts}       % blackboard math symbols
\usepackage{nicefrac}       % compact symbols for 1/2, etc.
\usepackage{microtype}      % microtypography
\usepackage{subcaption}
\usepackage{graphicx}
\usepackage{makecell}
\usepackage[ruled,vlined,linesnumbered]{algorithm2e}
\usepackage{multirow}
\usepackage{amsmath}

\newtheorem{problem}{Problem}
\usepackage{color,pifont}

\usepackage{amsmath}

\usepackage{wrapfig}
\usepackage{setspace}
\usepackage{color,colortbl}
\usepackage{tablefootnote}

\usepackage{booktabs}

\AtBeginDocument{%
  \providecommand\BibTeX{{%
    \normalfont B\kern-0.5em{\scshape i\kern-0.25em b}\kern-0.8em\TeX}}}

\setcopyright{acmcopyright}
\acmJournal{CSUR}
\acmYear{2022} \acmVolume{1} \acmNumber{1} \acmArticle{1}
\acmMonth{1} \acmPrice{15.00}\acmDOI{10.1145/3512467}

\acmJournal{CSUR}
\acmVolume{1}
\acmNumber{1}
\acmArticle{1}
\acmMonth{1}

\begin{document}

%% allowing the author to define a "short title" to be used in page 

% \title{A Survey of Knowledge-Enhanced Text Generation}

\title[A Survey of Knowledge-Enhanced Text Generation]{A Survey of Knowledge-Enhanced Text Generation \\ 
\vspace{0.2in}
\small \hfill Accepted to ACM Computing Survey (CUSR)
\vspace{-0.3in}
}


\author{Wenhao Yu}
\email{wyu1@nd.edu}
\affiliation{%
  \institution{University of Notre Dame}
  \city{Notre Dame}
  \state{Indiana}
  \country{USA}
  \postcode{46556}
}

\author{Chenguang Zhu}
\email{chezhu@microsoft.com}
\affiliation{%
  \institution{Microsoft Research}
  \city{Redmond}
  \state{Washington}
  \country{USA}
  \postcode{98052}
}

\author{Zaitang Li}
\email{1155107739@link.cuhk.edu.hk}
\affiliation{%
  \institution{The Chinese University of Hong Kong}
  \city{Hong Kong}
  \country{China}
  \postcode{999077}
}

\author{Zhiting Hu}
\email{zhitinghu@gmail.com}
\affiliation{%
  \institution{University of California at San Diego}
  \city{San Diego}
  \state{California}
  \country{USA}
  \postcode{92092}
}

\author{Qingyun Wang}
\email{qingyun4@illinois.edu}
\affiliation{%
  \institution{University of Illinois at Urbana-Champaign}
  \city{Urbana}
  \state{Illinois}
  \country{USA}
  \postcode{61801}
}

\author{Heng Ji}
\email{hengji@illinois.edu}
\affiliation{%
  \institution{University of Illinois at Urbana-Champaign}
  \city{Urbana}
  \state{Illinois}
  \country{USA}
  \postcode{61801}
}

\author{Meng Jiang}
\email{mjiang2@nd.edu}
\affiliation{%
  \institution{University of Notre Dame}
  \city{Notre Dame}
  \state{Indiana}
  \country{USA}
  \postcode{46556}
}

% \blfootnote{Acknowledgements: We thank all anonymous reviewers for valuable comments. We also appreciate the suggestions from readers of our pre-print version.
% We thank Dr. Michael Zeng (Microsoft Research) and Dr. Nazneen Rajani (Saleforce Research) for their constructive comments and suggestions.
% Wenhao Yu and Dr. Meng Jiang's research is supported by National Science Foundation grants IIS-1849816, CCF-1901059, and IIS-2119531.
% Qingyun Wang and Dr. Heng Ji's research is based upon work supported by Agriculture and Food Research Initiative (AFRI) grant no. 2020-67021-32799/project accession no.1024178 from the USDA National Institute of Food and Agriculture, U.S. DARPA SemaFor Program No. HR001120C0123, DARPA AIDA Program No. FA8750-18-2-0014, and DARPA KAIROS Program No. FA8750-19-2-1004. The views and conclusions contained herein are those of the authors and should not be interpreted as necessarily representing the official policies, either expressed or implied, of DARPA, or the U.S. Government. The U.S. Government is authorized to reproduce and distribute reprints for governmental purposes notwithstanding any copyright annotation therein.}



\begin{abstract}
The goal of text-to-text generation is to make machines express like a human in many applications such as conversation, summarization, and translation. It is one of the most important yet challenging tasks in natural language processing (NLP). Various
neural encoder-decoder models have been proposed to achieve the goal by learning to map input text to output text.
However, the input text alone often provides limited knowledge to generate the desired output, so the performance of text generation is still far from satisfaction in many real-world scenarios. To address this issue, researchers have considered incorporating (i) internal knowledge embedded in the input text and (ii) external knowledge from outside sources such as knowledge base and knowledge graph into the text generation system. This research topic is known as \textit{knowledge-enhanced text generation}. In this survey, we present a comprehensive review of the research on this topic over the past five years. The main content includes two parts: (i) general methods and architectures for integrating knowledge into text generation; (ii) specific techniques and applications according to different forms of knowledge data.
This survey can have broad audiences, researchers and practitioners, in academia and industry.
\end{abstract}



%
% Keywords. The author(s) should pick words that accurately describe
% the work being presented. Separate the keywords with commas.

\maketitle

\vspace{-0.05in}
\textbf{$\S\S$ Some useful materials related to this survey:}
\begin{itemize}
    \item A tutorial entitled ``Knowledge-enriched Natural Language Generation'', at EMNLP 2021. Tutorial abstract, slides, videos can be found at \textcolor{blue}{\url{https://kenlg-tutorial.github.io}}.
    \item A tutorial entitled ``Knowledge-augmented Methods for NLP'', to appear at ACL 2022.
    \item A Github repository with a more complete collection of papers and codes can be found at \\ \textcolor{blue}{\url{https://github.com/wyu97/KENLG-Reading}}. It will be frequently updated with new papers.
\end{itemize}

\vspace{-0.1in}
\section{Introduction}
Text generation, which is often formally referred as natural language generation (NLG), is one of the most important yet challenging tasks in natural language processing (NLP)~\cite{garbacea2020neural}.
NLG aims at producing understandable text in human language from linguistic or non-linguistic data in a variety of forms such as textual data, numerical data, image data, structured knowledge bases, and knowledge graphs.
Among these, text-to-text generation is one of the most important applications and thus often shortly referred as ``text generation''. Researchers have developed numerous technologies for this task in a wide range of applications~\cite{gatt2018survey,wang2020towards,iqbal2020survey}.
Text generation takes text (e.g., a sequence, keywords) as input, processes the input text into semantic representations, and generates desired output text.
For example, machine translation generates text in a different language based on the source text; summarization generates an abridged version of the source text to include salient information; question answering (QA) generates textual answers to given questions; dialogue system supports chatbots to communicate with humans with generated responses.

With the recent resurgence of deep learning technologies~\cite{lecun2015deep}, deep neural NLG models have achieved remarkable performance in enabling machines to understand and generate natural language. A basic definition of the text generation task is to generate an expected \emph{output sequence} from a given \emph{input sequence}, called sequence-to-sequence (Seq2Seq). 
The Seq2Seq task and model were first introduced in 2014~\cite{sutskever2014sequence}. It maps an input text to an output text under encoder-decoder schemes. The encoder maps the input sequence to a fixed-sized vector, and the decoder maps the vector to the target sequence.
Since then, developing NLG systems has rapidly become a hot topic. Various text generation models have been proposed under deep neural encoder-decoder architectures. Popular architectures include recurrent neural network (RNN) encoder-decoder~\cite{sutskever2014sequence}, convolutional neural network (CNN) encoder-decoder~\cite{gehring2017convolutional}, and Transformer encoder-decoder~\cite{vaswani2017attention}.
% The attention mechanism \cite{bahdanau2015neural} and copy/pointing mechanism \cite{gu2016incorporating,see2017get} are two widely used mechanisms to improve the performance of generation models.
% Encoder-decoder models have been used for various NLG tasks such as dialogue system \cite{zhang2020grounded} and summarization \cite{lin2019abstractive}.

Nevertheless, the input text alone contains limited knowledge to support neural generation models to produce the desired output.
Meanwhile, the aforementioned methods generally suffer from an inability to well comprehend language, employ memory to retain and recall knowledge, and reason over complex concepts and relational paths; as indicated by their name, they involve encoding an input sequence, providing limited reasoning by transforming their hidden state given the input, and then decoding to an output.
Therefore, the performance of generation is still far from satisfaction in many real-world scenarios. For example, in dialogue systems, conditioning on only the input text, a text generation system often produces trivial or non-committal responses of frequent words or phrases in the corpus~\cite{xing2017topic,zhou2018commonsense}, such as \emph{``Me too.''} or \emph{``Oh my god!''} given the input text \emph{``My skin is so dry.''} These mundane responses lack meaningful content, in contrast to human responses rich in knowledge. In comparison, humans are constantly acquiring, understanding, and storing knowledge from \emph{broader sources} so that they can be employed to understand the current situation in communicating, reading, and writing. For example, in conversations, people often first select \emph{concepts from related topics} (e.g., sports, food), then organize those topics into understandable content to respond; for summarization, people tend to write summaries containing \emph{keywords} used in the input document and perform necessary modifications to ensure grammatical correctness and fluency; in question answering (QA), people use \emph{commonsense} or \emph{professional knowledge} pertained to the question to infer the answer. Therefore, it is often the case that knowledge beyond the input sequence is required to produce informative output text.

\vspace{-0.05in}
\subsection{What is Knowledge-enhanced Text Generation?}

\begin{figure}[t]
  \begin{center}
    \includegraphics[width=1.0\textwidth]{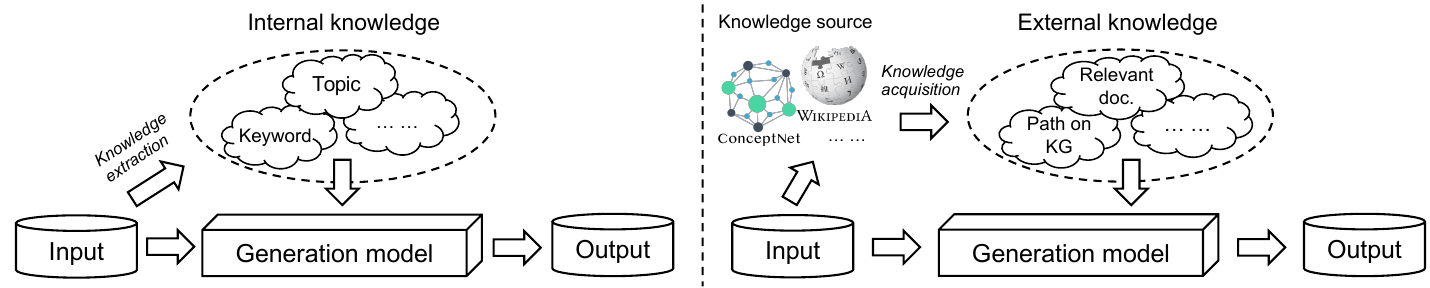}
  \end{center}
  \vspace{-0.1in}
  \caption{We divide different knowledge sources into internal knowledge and external knowledge. Internal knowledge creation takes place within the input text(s), while external knowledge acquisition occurs when knowledge is provided from outside sources (e.g., Wikipedia, ConceptNet~\cite{speer2017conceptnet}).}
  \label{fig:int-ent-framework}
\vspace{-0.05in}
\end{figure}

In general, knowledge is the familiarity, awareness, or understanding that coalesces around a particular subject. 
In NLG systems, knowledge is an awareness and understanding of the input text and its surrounding context.
These knowledge sources can be categorized into internal knowledge and external knowledge (see Figure~\ref{fig:int-ent-framework}).
\textit{Internal knowledge} creation takes place within the input text(s), including but not limited to keyword, topic, linguistic features, and internal graph structure. \textit{External knowledge} acquisition occurs when knowledge is provided from outside sources, including but not limited to knowledge base, external knowledge graph, and grounded text. 
% This knowledge can be learnt by many different methods and from various information sources, including but not limited to keywords, topics, linguistic features, knowledge bases, knowledge graphs, and grounded texts (see Figure~\ref{fig:overall}). 
These sources provide information (e.g., commonsense triples, topic words, reviews, background documents) that can be used as knowledge through various neural representation learning methods, and then applied to enhance the process of text generation. 
In addition, knowledge introduces interpretability for models with explicit semantics.
This research direction of incorporating knowledge into text generation is named as \textit{knowledge-enhanced text generation}.
%~\cite{guan2020knowledge,wang2020towards} \hzt{The two references here are probably not necessary and not appropriate, since we are bringing up the general concept here while the two papers are narrowly about specific applications/methods}. 

\vspace{-0.05in}
\begin{problem}[Knowledge-enhanced Text Generation] 
%Suppose we have 
Given a text generation problem where the system is given an input sequence $X$,
% $X = \{x_1, x_2, \cdots , x_n \} \in \mathcal{X}$ \hzt{Probably do not restrain the definition to a ``sequence'' to sequence setting.}, 
and aims to generate an output sequence $Y$.
% $Y = \{y_1, y_2, \cdots , y_m\} \in \mathcal{Y}$. Here $n$ and $m$ are the number of words in the input and output sequences, respectively. 
Assume we also have access to additional knowledge denoted as $K$. Knowledge-enhanced text generation aims to incorporate the knowledge $K$ to enhance the generation of $Y$ given $X$, through leveraging the dependencies among the input text, knowledge, and output text.
%The system should be able to learn the dependencies among the input, knowledge, and output. It can be written as a predictive function: $f: (\mathcal{X}, \mathcal{K}) \rightarrow \mathcal{Y}$, which maps the input sequence $X \in \mathcal{X}$ to the output sequence $Y \in \mathcal{Y}$.
\label{prob:seq2seq}
\end{problem}

Many existing knowledge-enhanced text generation systems have demonstrated promising performance on generating informative, logical, and coherent texts. In dialogue systems, a topic-aware Seq2Seq model helped understand the semantic meaning of an input sequence and generate a more informative response such as \emph{``Then hydrate and moisturize your skin.''} to the aforementioned example input \emph{``My skin is so dry.''} 
%The responses by the topic aware systems are two to three times more informative than those by a vanilla Seq2Seq~\cite{xing2017topic}.
%Another example is that summarization models could not generate coherent summaries for original documents due to their greatly imbalanced length~\cite{narayan2018don,wang2019topic,huang2020knowledge}.
%One way to evaluate the coherence is to use the summary to answer questions designed for the original document.
In summarization, knowledge graph produced a structured summary and highlight the proximity of relevant concepts, when complex events related with the same entity may span multiple sentences. A knowledge graph enhanced Seq2Seq model generated summaries that were able to correctly answer 10\% more topically related questions~\cite{huang2020knowledge}.
% The evaluation result shows adding structured representation from knowledge graph enables the generated summary to correctly answer 10\% more questions.
In question answering (QA) systems, facts stored in knowledge bases completed missing information in the question and elaborate details to facilitate answer generation \cite{he2017generating,fan2019using}.
%For example, to answer the question \emph{``What is the most famous landmark in Japan?''}, one needs to reason and infer based on relevant facts and commonsense in knowledge bases.
In story generation, using commonsense knowledge acquired from knowledge graph facilitated understanding of the storyline and better
narrate following plots step by step, so each step could be reflected as a link on the knowledge graph and the whole story would be a path~\cite{guan2019story}.

\vspace{-0.05in}
\subsection{Why a Survey of Knowledge-enhanced Text Generation?}

Recent years have witnessed a surge of interests in
% Researchers have proposed to 
developing methods for incorporating knowledge in NLG beyond input text. However, there is a lack of comprehensive survey of this research topic.
% Existing surveys in this area have only partially reviewed some related topics.
Related surveys have laid the foundation of discussing this topic.
% only partially review some work in the field.
For example, Garbacea et al.~\cite{garbacea2020neural} and Gatt et al.~\cite{gatt2018survey} reviewed model architectures for core NLG tasks but did not discuss knowledge-enhanced methods.
Ji et al.~\cite{ji2020survey} presented a review on
% conducted a review on
knowledge graph techniques which could be used for enhancing NLG.
% , some of  which have been applied to enhance NLG performance.
Wang et al.~\cite{wang2020towards} summarized how to represent structural knowledge such as knowledge base and knowledge graph for reading comprehension and retrieval.

To the best of our knowledge, this is the first survey that presents a comprehensive review of knowledge-enhanced text generation. It aims to provide NLG researchers a synthesis and pointer to related research. Our survey includes a detailed discussion about how NLG can benefit from recent progress in deep learning and artificial intelligence, including technologies such as graph neural network, reinforcement learning, and neural topic modeling.

\vspace{-0.05in}
\subsection{What are the Challenges in Knowledge-enhanced Text Generation?}

%Many researchers in academia and industry have been developing knowledge enhanced text generation systems.
% Due to the diversity of knowledge sources, the primary challenge of knowledge enhanced NLG is how to extract the most relevant knowledge.
To start with, we note that the \textit{first challenge} in knowledge-enhanced NLG is to \emph{obtain} useful related knowledge from diverse sources.
% Existing works incorporate different types of knowledge ranging from including topic, keyword, knowledge base, knowledge graph to knowledge grounded text.
There has been a rising line of work that discovers knowledge from topic, keyword, knowledge base, knowledge graph and knowledge grounded text.
The \textit{second challenge} is how to effectively \emph{understand} and \emph{leverage} the acquired knowledge to facilitate text generation. Multiple methods have been explored to improve the encoder-decoder architecture (e.g., attention mechanism, copy and pointing mechanism).

\begin{figure}[t]
  \begin{center}
    \includegraphics[width=1.0\textwidth]{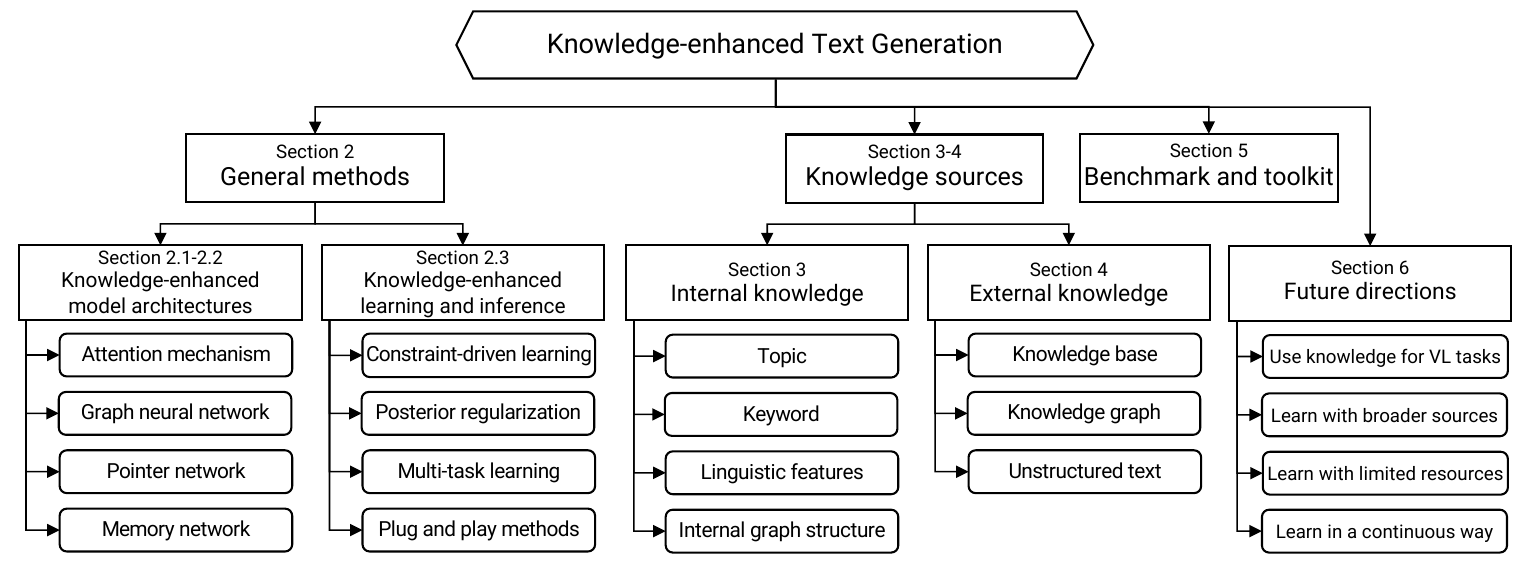}
  \end{center}
  \vspace{-0.2in}
  \caption{Categorization of information sources and methods for  knowledge-enhanced text generation. Knowledge can be learnt from various information sources, and then integrated into the generation process.}
  \label{fig:overall}
\vspace{-0.15in}
\end{figure}

% \begin{wrapfigure}{R}{0.55\textwidth}
%   \vspace{-0.1in}
%   \begin{center}
%     \includegraphics[width=0.55\textwidth]{figures/paper-trend.pdf}
%   \end{center}
%   \vspace{-0.2in}
%   \label{fig:paper-num}
%   \caption{Knowledge-enhanced text generation has been gaining emerging interests in the recent five years.}
%   \vspace{-0.2in}
% \end{wrapfigure}

Based on the first challenge, the main content of our survey is divided into two parts: (1) general methods of integrating knowledge into text generation (Section 2); (2) specific methods and applications according to different sources of knowledge enhancement (Sections 3--4). 
More concretely, since knowledge can be obtained from different sources, we first divide existing knowledge enhanced text generation work into two categories: internal knowledge enhanced and external knowledge enhanced text generation. The division of internal and external knowledge is widely adopted by management science~\cite{menon2003valuing}, which can be analogous with knowledge enhanced text generation.
% {Internal knowledge} creation takes place within the input text(s), including but not limited to keyword, topic, linguistic features, and internal graph structures. {External knowledge} acquisition occurs when knowledge is provided from outside sources, including but not limited to knowledge base, external knowledge graph, and grounded text.
Based on the second challenge, we categorize recent knowledge-enhanced text generation  methods evolved from how knowledge is extracted and incorporated into the process of text generation in each section (named as M1, M2, and etc).
Furthermore, we review methods for a variety of natural language generation applications in each section to help practitioners choose, learn, and use the methods.
In total, we discuss seven mainstream applications presented in more than 80 papers that were published or released in or after the year of 2016.

As shown in Figure \ref{fig:overall}, the remainder of this survey is organized as follows.
Section 2 presents basic NLG models and general methods of integrating knowledge into text generation. 
Sections 3 reviews internal knowledge-enhanced NLG methods and applications. The internal knowledge is obtained from topic, keyword, linguistic features and internal graph structures.
Sections 4 reviews external knowledge-enhanced NLG methods and applications. The external knowledge sources include knowledge bases, knowledge graphs, and grounded text.
Section 5 presents knowledge-enhanced NLG benchmarks.
Section 6 discusses future work and concludes the survey.

% \begin{table}[h]
% \begin{center}
% \vspace{-0.1in}
% \caption{Basic symbols used in this survey and their descriptions.}
% \vspace{-0.15in}
% \setlength{\tabcolsep}{1mm}{
% \scalebox{0.7}{%
% \begin{tabular}{l|l||l|l}
% \toprule
% \textbf{Symbol} & \textbf{Description} & \textbf{Symbol} & \textbf{Description} \\ 
% \hline
% $X, \mathcal{X}$ & input sequence item, and the set of all input sequences &
% $Y, \mathcal{Y}$ & output sequence item, and the set of all output sequences \\
% $x_i, \textbf{e}(x_i)$ & $i$-th word in the input sequence and its word embedding &
% $y_i, \textbf{e}(y_i)$ & $i$-th word in the output sequence and its word embedding \\
% $y_{<t}$ & a sequence of words from $y_0$ to $y_{t-1}$, i.e., $(y_0, \cdots, y_{t-1})$ &
% $\textbf{h}_i/\textbf{H}_i$ & encoding hidden state/matrix at $i$-th step \\
% $\textbf{s}_i/\textbf{S}_i$ & decoding hidden state/matrix at $i$-th step &
% $\textbf{c}_i$ & decoding context state at $i$-th step \\
% $\eta(\cdot)$ & a nonlinear, potentially multi-layered, function (e.g., MLP) &
% $\sigma(\cdot)$ & a nonlinear activation function (e.g., tanh, ReLU) \\
% $\mathcal{V}, \mathcal{V}_X$ & pre-defined vocabulary and source sequence vocabulary &
% $\mathcal{J}, \mathcal{L}$ & objective function, and loss function \\
% \bottomrule
% \end{tabular}}}
% \end{center}
% \vspace{-0.2in}
% \end{table}

\vspace{-0.05in}
\section{General Methods of Integrating Knowledge into NLG}

\subsection{The Basic Text Generation Models}

Early encoder-decoder frameworks are often based on recurrent neural network (RNN) such as RNN-Seq2Seq~\cite{sutskever2014sequence}. Convolutional neural network (CNN) based encoder-decoder~\cite{gehring2017convolutional} and Transformer encoder-decoder~\cite{vaswani2017attention} have been increasingly widely used. From a probabilistic perspective, the encoder-decoder frameworks learn the conditional distribution over a variable length sequence conditioned on yet another variable length sequence:
\begin{equation}
    \setlength\abovedisplayskip{2pt} 
    \setlength\belowdisplayskip{2pt}
    P(Y|X) = P(y_1, \cdots, y_m|x_1, \cdots, x_n) = \prod_{t=1}^{m}p(y_t|X, y_1, \cdots, y_{t-1}).
\end{equation}

\vspace{-0.05in}
\paragraph{Encoder.}
The encoder learns to encode a variable length sequence into a fixed length vector representation. RNN encoder reads the input sentence $X$ \textit{sequentially}. CNN encoder performs convolutional operations on a word and its surrounding word(s) in a sequential window.
Transformer encoder eschews recurrence and instead relying entirely on the self-attention mechanism to draw global dependencies between different tokens in the input $X$. We denote them uniformly as:
\begin{equation}
    (\textbf{h}_1, \textbf{h}_2, \cdots, \textbf{h}_n)=\textsc{Encoder}(\textbf{e}(x_1), \textbf{e}(x_2), \cdots, \textbf{e}(x_n)),
\end{equation}
where $\textbf{e}(x_i)$ is the word embedding of word $x_i$, $\textbf{h}_i$ is the contextualized hidden representation of $x_i$.
% In RNN encoder, the last hidden state $\textbf{h}_n $ is seen as the representation of the whole input sequence, denoted as $\textbf{c} = \textbf{h}_n$. 

\vspace{-0.05in}
\paragraph{Decoder.}
The decoder is to decode a given fixed length vector representation into a variable length sequence~\cite{sutskever2014sequence}. Specially, the decoder generates an output sequence one token at each time step. At each step the model is auto-regressive, consuming the previously generated tokens as additional input when generating the next token. Formally, the decoding function is represented as:
\begin{equation}
    \setlength\abovedisplayskip{2pt} 
    \setlength\belowdisplayskip{2pt}
    \textbf{s}_t = \textsc{Decoder} (\textbf{s}_{t-1}, \textbf{e}(y_{t-1})), \label{eq:Seq2Seq-decoder}
\end{equation}
\begin{equation}
    \setlength\abovedisplayskip{2pt} 
    \setlength\belowdisplayskip{2pt}
    p(y_t|y_{t-1}, y_{t-2}, \cdots , y_1) = \textsc{Readout}(\textbf{s}_t),
\end{equation}
where $\textsc{Readout}(\cdot)$ is a nonlinear multi-layered function that outputs the probability of $y_t$.

\vspace{-0.05in}
\paragraph{Optimization} A generation process is regarded as a sequential multi-label classification problem. It can be directly optimized by the negative log likelihood ({NLL}) loss. Therefore, the objective of a text generation model via maximum likelihood estimation (MLE) is formulated as:
\begin{equation}
    \setlength\abovedisplayskip{2pt} 
    \setlength\belowdisplayskip{1pt}
    \mathcal{L}_{NLL}(\theta) = - \log p_{\theta}(Y|X)  = - \sum^m_{t=1} \log \left( p_{\theta}(y_t|y_{< t}, X) \right).
\label{eq:Seq2Seq-loss}
\end{equation}

\vspace{-0.05in}
\subsection{Knowledge-enhanced Model Architectures}

The most popular idea of incorporating knowledge is designing \emph{specialized architectures} of text generation models that can reflect the particular type of knowledge.
% The perhaps most popular way of enhancing text generation with knowledge is by designing specialized model architectures that reflect the particular knowledge.
In the context of neural networks, several general neural architectures are widely used and customized to bake the knowledge about the problems being tackled into the models.

\vspace{-0.03in}
\subsubsection{\textbf{Attention Mechanism}}
\label{sec:k-attn}

It is useful to capture the weight of each time step in both encoder and decoder~\cite{bahdanau2015neural}.
During the decoding phase, the context vector $\textbf{c}_t$ is added, so the hidden state $\textbf{s}_t$ is:
\begin{equation}
    \setlength\abovedisplayskip{2pt} 
    \setlength\belowdisplayskip{2pt}
   \textbf{s}_t = \textsc{Decoder} (\textbf{s}_{t-1}, \textbf{e}(y_{t-1}), \textbf{c}_t).
\end{equation}
% Note that 
Unlike Eq.(\ref{eq:Seq2Seq-decoder}), here the probability is conditioned on the distinct context vector $\textbf{c}_t$ for target word $y_t$, and $\textbf{c}_t$ depends on a sequence of hidden states $\textbf{H} = \{\textbf{h}_i\}_{i=1}^{n}$ that were mapped from input sequence. 

In RNN-Seq2Seq decoder, the $\textbf{c}_t$ is computed as a weighted sum of $\{\textbf{h}_i\}_{i=1}^{n}$:
\begin{equation}
    \setlength\abovedisplayskip{2pt} 
    \setlength\belowdisplayskip{2pt}
   \textbf{c}_t = \sum^{n}_{i=1} \alpha_{ti} \textbf{h}_i, ~\text{where}~ \alpha_{ti} = \frac{\exp (\eta(\textbf{s}_{t-1}, \textbf{h}_i ))} {\sum^{n}_{k=1} \exp(\eta(\textbf{s}_{t-1}, \textbf{h}_k)) } ,    
   \label{eq:atten-weight}
\end{equation}
where $\eta(\cdot)$ is parametrized as a multi-layer perception to compute a soft alignment. 
$\eta(\cdot)$ enables the gradient of loss function to be backpropagated. There are six alternatives for the $\eta(\cdot)$ function (see Table 2 in \cite{garbacea2020neural}). The probability $\alpha_{ti}$ reflects the importance of the hidden state of input sequence in presence of the previous hidden state $\textbf{s}_{t-1}$ for deciding the next hidden state.

In Transformer decoder, on top of the two sub-layers in the encoder, the decoder inserts a third sub-layer, which performs multi-head attention over the output of the encoder stack $\textbf{H}$. 
Efficient implementations of the transformer use the cached history matrix $\textbf{S}_t$ to generate next token.
To compare with RNN-Seq2Seq, we summarize the Transformer decoder using recurrent notation:
\begin{equation}
    \textbf{S}_t = \textsc{Transformer-Decoder}(\textbf{S}_{t-1}, \textbf{e}(y_{t-1}), \textbf{H}), 
\label{eq:transformer}
\end{equation}
where $\textbf{S}_t$ = $[(\textbf{K}^{(1)}_t, \textbf{V}^{(1)}_t), \cdots, (\textbf{K}^{(l)}_t, \textbf{V}^{(l)}_t)]$, where $(\textbf{K}^{(i)}_t, \textbf{V}^{(i)}_t)$ corresponds to the key-value pairs from the $i$-th layer generated at all time-steps from $0$ to $t$.
Instead of noting a specific name, we will use \textsc{Encoder}($\cdot$) and \textsc{Decoder}($\cdot$) to represent encoder and decoder in the following sections.
% instead of noting a specific kind of encoder and decoder.

\begin{table*}[t]
\caption{NLG methods that incorporates knowledge attention ($\S$\ref{sec:k-attn}) and knowledge mode ($\S$\ref{sec:k-mode}).}
\vspace{-0.15in}
\centering
\scalebox{0.82}{\begin{tabular}{|l|c|c|c|c|c|}
\hline
& Topic & Keyword & Knowledge base & Knowledge graph & Grounded text \\
\hline \hline
Knowledge-related attention & \cite{xing2017topic,wei2019translating,zhang2016topic} & \cite{li2018guiding,li2019coherent,li2020keywords} & \cite{he2017generating,fu2018natural} & \cite{zhang2020grounded,huang2020knowledge,zhou2018commonsense,guan2019story} & \cite{meng2020refnet,bi2020generating} \\
Knowledge-related mode & \cite{xing2017topic} & \cite{li2020keywords} & \cite{he2017generating} & \cite{zhou2018commonsense,zhang2020grounded,ji2020language} & \cite{meng2020refnet,ren2020thinking} \\
Knowledge-related memory & \cite{fu2018natural,zhou2018emotional} & - &  \cite{madotto2018mem2seq,wu2019global} & \cite{yang2019enhancing} & \cite{kim2020sequential} \\
\hline
\end{tabular}}
\vspace{-0.15in}
\label{tab:mul-source-attn}
\end{table*}

\vspace{-0.05in}
\paragraph{Knowledge-related attention}
Attention mechanism has been widely used to incorporate knowledge representation in recent knowledge-enhanced NLG work.
% Leveraging attention mechanism to incorporate knowledge representation into the decoding phase has been widely used in recent knowledge-enhanced NLG work.
The general idea is to learn a knowledge-aware context vector (denoted as $\widetilde{\textbf{c}}_t$) by combining both hidden context vector ($\textbf{c}_t$) and knowledge context vector (denoted as $\textbf{c}^{K}_t$) into decoder update, such as $\widetilde{\textbf{c}}_t = f_{mlp}(\textbf{c}_t \oplus \textbf{c}^{K}_t)$. 
The knowledge context vector ($\textbf{c}^{K}_t$) calculates attentions over knowledge representations (e.g., topic vectors, node vectors in knowledge graph).
Table \ref{tab:mul-source-attn} summarizes a variety of knowledge attentions, including
keyword attention~\cite{li2018guiding,li2020keywords,li2019coherent}, topic attention~\cite{xing2017topic,liu2019generating,wei2019translating,zhang2016topic}, knowledge base attention~\cite{he2017generating,fu2018natural}, knowledge graph attention~\cite{zhang2020grounded,huang2020knowledge,koncel2019text}, and grounded text attention~\cite{meng2020refnet,bi2020generating}. 

\vspace{-0.03in}
\subsubsection{\textbf{Copy and Pointing Mechanisms}}
\label{sec:k-mode}

CopyNet and Pointer-generator (PG) are used to choose subsequences in the input sequence and put them at proper places in the output sequence.

CopyNet and PG have a differentiable network architecture~\cite{gu2016incorporating}. They can be easily trained in an end-to-end manner. In CopyNet and PG, the probability of generating a target token is a combination of the probabilities of two modes, generate-mode and copy-mode. First, they represent unique tokens in the global vocabulary $\mathcal{V}$ and the vocabulary of source sequence $\mathcal{V}_X$. They build an extended vocabulary $\mathcal{V}_{\text{ext}} = \mathcal{V} \cup \mathcal{V}_X \cup \{\text{unk}\}$. The difference between CopyNet and PG is the way to calculate distribution over the extended vocabulary. CopyNet calculates the distribution by
\begin{equation}
    \setlength\abovedisplayskip{2pt} 
    \setlength\belowdisplayskip{2pt}
    p(y_t)  =  p_{g}(y_t) +  p_{c}(y_t),
\label{eq:mode1}
\end{equation}
where $p_{g}(\cdot|\cdot)$ and $p_{c}(\cdot|\cdot)$ stand for the probability of generate-mode and copy-mode.
Differently, PG explicitly calculates a switch probability $p_{m}$ between generate-mode and copy-mode. It recycles the attention distribution to serve as the copy distribution. The distribution over $\mathcal{V}_{\text{ext}}$ is calculated by
\begin{equation}
    \setlength\abovedisplayskip{2pt} 
    \setlength\belowdisplayskip{2pt}
    p(y_t) = p_m{(\mathrm{g})} \cdot p_g(y_t) + (1 - p_m{(\mathrm{g})}) \cdot p_c(y_t),
\end{equation}
where $p_m(\mathrm{g})$ indicates the probability of choosing generate-mode, which is obtained by a nonlinear multi-layered (MLP) function.
Importantly, CopyNet and pointer-generator network have been used as the \textit{base module} for a lot of knowledge-enhanced NLG work.

\vspace{-0.05in}
\paragraph{Knowledge-related mode}
A knowledge-related mode chooses subsequences in the obtained knowledge and puts them at proper places in the output sequence. It helps NLG models to generate words that are not included in the global vocabulary ($\mathcal{V}$) and input sequence ($\mathcal{V}_X$).
For example, by adding the model of knowledge base, the extended vocabulary ($\mathcal{V}_{ext}$) adds entities and relations from the knowledge base, i.e., $\mathcal{V}_{ext} = \mathcal{V} + \mathcal{V}_X + \mathcal{V}_{KB}$. The probability of generating a target token is a combination of the probabilities of three modes: generate-mode, copy-mode and knowledge base-mode. Therefore, knowledge-related mode is not only capable of regular generation of words but also operation of producing appropriate subsequences in knowledge sources. Table \ref{tab:mul-source-attn} summarizes different kinds of knowledge-related modes such as topic mode~\cite{xing2017topic}, keyword mode~\cite{li2020keywords}, knowledge base mode~\cite{he2017generating}, knowledge graph mode~\cite{zhou2018commonsense,zhang2020grounded}, and background mode~\cite{meng2020refnet,ren2020thinking}.

\vspace{-0.03in}
\subsubsection{\textbf{Memory Network}}
Memory networks (MemNNs) are recurrent attention models over a possibly large external memory~\cite{sukhbaatar2015end}. They write external memories into several embedding matrices, and use query (generally speaking, the input sequence $X$) vectors to read memories repeatedly. This approach encodes long dialog history and memorize external information.

Given an input set $\{m_1, \cdots, m_i\}$ to be stored in memory. The memories of MemNN are
represented by a set of trainable embedding matrices $\textbf{C} = \{\textbf{C}^1, \cdots, \textbf{C}^{K+1} \}$, where each $\textbf{C}^k$ maps tokens to vectors, and a query (i.e., input sequence) vector $\textbf{h}_X^{k}$ is used as a reading head. The model loops over $K$ hops and it computes the attention weights at hop $k$ for each memory $m_i$ using:
\begin{equation}
    \setlength\abovedisplayskip{2pt} 
    \setlength\belowdisplayskip{2pt}
    \textbf{p}^k_i = \mathrm{softmax}((\textbf{h}^k_X)^\top \textbf{C}^k_i),   
\end{equation}
where $\textbf{C}^k_i = \textbf{C}^k(m_i) $ is the memory content in $i$-th position, i.e., mapping $m_i$ into a memory vector. Here, $\textbf{p}^k$ is a soft memory selector that decides the memory relevance with respect to the query vector $\textbf{h}_X^{k}$. Then, the model reads out the memory $\textbf{o}^k$ by the weighted sum over $\textbf{C}^{k+1}$,
\begin{equation}
\textbf{o}^k = \sum_i \textbf{p}^k_i \textbf{C}^{k+1}_i.
\end{equation}
Then, the query vector is updated for the next hop by using $\textbf{h}_X^{k+1} = \textbf{h}_X^{k} + \textbf{o}^k$. The result from the encoding step is the memory vector $\textbf{o}^K$ and becomes the input for the decoding step.

\vspace{-0.03in}
\paragraph{Knowledge-related memory} Memory augmented encoder-decoder framework has achieved promising progress for many NLG tasks. For example, MemNNs are widely used for encoding dialogue history in task-oriented dialogue systems~\cite{wu2019global,reddy2019multi}. Such frameworks enable a decoder to retrieve information from a memory during generation. Recent work explored to model external knowledge with memory network such as knowledge base~\cite{madotto2018mem2seq,yang2019enhancing} and topic
~\cite{fu2018natural,zhou2018emotional}.

\vspace{-0.03in}
\subsubsection{\textbf{Graph Network}}
\label{sec:graph-net}

Graph network captures the dependence of graphs via message passing between the nodes of graphs.
Graph neural networks (GNNs)~\cite{wu2020comprehensive} and graph-to-sequence (Graph2Seq)~\cite{beck2018graph} potentiate to bridge up the gap between graph representation learning and text generation. 
Knowledge graph, dependency graph, and other graph structures can be integrated into text generation through various GNN algorithms. Here we denote a graph as $\mathcal{G} = (\mathcal{U}, \mathcal{E})$, where $\mathcal{U}$ is the set of entity nodes and $\mathcal{E}$ is the set of (typed) edges.
Modern GNNs typically follow a neighborhood aggregation approach, which iteratively updates the representation of a node by aggregating information from its neighboring nodes and edges. After $k$ iterations of aggregation, a node representation captures the structural information within its $k$-hop neighborhood. Formally, the $k$-th layer of a node $u \in \mathcal{U}$ is:
\begin{equation}
    \setlength\abovedisplayskip{2pt} 
    \setlength\belowdisplayskip{2pt}
    \textbf{u}^{(k)} = \textsc{Combine}_k (\textbf{u}^{(k-1)},  \textsc{Aggregate}_k(\big{\{} (\textbf{u}^{(k-1)}_{i}, \textbf{e}_{ij}^{(k-1)}, \textbf{u}^{(k-1)}_{j}): \forall (u_i, e_{ij}, u_j) \in \mathcal{N}(u)\big{\}})),
    % \label{eq:gnns}
\end{equation}
where $\mathcal{N}(u)= \{(u_i, e_{ij}, u_j) \in \mathcal{E} | u_i = u ~\text{or}~ u_j = u\}$ denotes the set of edges containing node $u$, $\textbf{u}^{(k)}$ and $\textbf{e}_{ij}^{(k)}$ are feature vectors of a node $u$ and the edge between $u_i$ and $u_j$ at the $k$-th iteration/layer. The choice of $\textsc{Aggregate}(\cdot)$ and $\textsc{Combine}(\cdot)$ in GNNs is crucial. A number of architectures for $\textsc{Aggregate}(\cdot)$ have been proposed in different GNN works such as GAT~\cite{velivckovic2018graph}. Meanwhile, the $\textsc{Aggregate}(\cdot)$ function used in labeled graphs (e.g., a knowledge graph) is often taken as those GNNs for modeling relational graphs~\cite{schlichtkrull2018modeling}.
% , such as a relational GAT~\cite{zhou2018commonsense}. 
% \begin{equation}
%     \textbf{u}^{(k)} = \sigma ~ \Big{(} \sum_{(u_i, e_{ij}, u_j) \in \mathcal{N}(u)} \alpha(u_i, e_{ij}, u_j) \cdot (\textbf{u}^{(k-1)}_i \oplus \textbf{u}^{(k-1)}_j) \Big{)}, 
% \end{equation}
% \begin{equation}
%     \alpha(u_i, e_{ij}, u_j) = \frac{\exp( \eta(\textbf{u}_i, \textbf{e}_{ij}, \textbf{u}_j))}{\sum_{(u_{i^\prime}, e_{ij}, u_{j^\prime}) \in \mathcal{N}(u)} \exp(\eta(\textbf{u}_{i^\prime}, \textbf{e}_{ij}, \textbf{u}_{j^\prime}))},  
% \end{equation}
% \begin{equation}
%     \eta(u_i, e_{ij}, u_j)  = (\textbf{W}_e \cdot \textbf{e}_{ij}^{(k-1)})^\top \tanh(\textbf{W}_i \cdot \textbf{u}^{(k-1)}_i + \textbf{W}_j \cdot \textbf{u}^{(k-1)}_j),
% \end{equation}
% where $\sigma(\cdot)$ denotes a nonlinear activation function (often taken as LeakyReLU), and $\textbf{W}_i, \textbf{W}_j, \textbf{W}_e$ are all trainable weight matrices. In practical, $K$ is usually set as $K=2$~\cite{zhu2020boosting,huang2020knowledge,zhang2020grounded} or
% $K=1$~\cite{zhou2018commonsense,guan2019story,li2019coherent} (see performance comparisons with different K in Figure 3 of~\cite{banerjee2019graph}). Graph attention mechanism could be multi-head attention~\cite{huang2020knowledge}. The attention weight measures the association of a relation $e_{ij}$ between two entity nodes $u_i$ and $u_j$. 
To obtain the representation of graph $\mathcal{G}$ (denoted as $\textbf{h}_{G}$), the $\textsc{Readout}(\cdot)$ function (either a simple permutation invariant function or sophisticated graph-level pooling function) pools node features from the final iteration $K$,
\begin{equation}
    \setlength\abovedisplayskip{2pt} 
    \setlength\belowdisplayskip{2pt}
     \textbf{h}_{G} =
     \textsc{Readout}(\big{\{}\textbf{u}^{(K)}: u \in \mathcal{U}\big{\}}).
\end{equation}
% where \textsc{Readout}$(\cdot)$ can be a simple permutation invariant function or sophisticated graph-level pooling function. 

\vspace{-0.05in}
\paragraph{Applications} Graph network has been commonly used in integrating knowledge in graph structure such as knowledge graph and dependency graph. Graph attention network~\cite{velivckovic2018graph} can be combined with sequence attention and jointly optimized~\cite{zhou2018commonsense,zhang2020grounded}. 
{We will introduce different graph structure knowledge in subsequent sections such as knowledge graph (Section \ref{sec:know-graph}), dependency graph (Section \ref{sec:syn-graph}-\ref{sec:sem-graph}), and open knowledge graph (OpenKG) (Section \ref{sec:openkg}).}

\vspace{-0.03in}
\subsubsection{\textbf{Pre-trained Language Models}}

Pre-trained language models (PLMs) aims to learn universal language representation by conducting self-supervised training on large-scale unlabeled corpora.
Recently, substantial PLMs such as BERT~\cite{devlin2019bert} and T5~\cite{raffel2020exploring} have achieved remarkable performance in various NLP downstream tasks.
However, these PLMs suffer from two issues when performing on knowledge-intensive tasks.
First, these models struggle to grasp structured world knowledge, such as concepts and relations, which are very important in language understanding. For example, BERT cannot deliver great performance on many commonsense reasoning and QA tasks, in which many of the concepts are directly linked on commonsense knowledge graphs~\cite{yu2020jaket}.
Second, due to the domain discrepancy between pre-training and fine-tuning, these models do not perform well on domain-specific tasks. For example, BERT can not give full play to its value when dealing with electronic medical record analysis task in the medical field~\cite{liu2020k}.

Recently, a lot of efforts have been made on investigating how to integrate knowledge into PLMs~\cite{yu2020jaket,liu2021kg,liu2020k,xiong2020pretrained,guan2020knowledge,zhou2021pre}. Specifically, we will introduce some PLMs designed for NLG tasks.
Overall, these approaches can be grouped into two categories:
The first one is to explicitly inject entity representation into PLMs, where the representations is pre-computed from external sources~\cite{zhang2019ernie,liu2021kg}. For example, KG-BART encoded the graph structure of KGs with knowledge embedding algorithms like TransE~\cite{bordes2013translating}, and then took the informative entity embeddings as auxiliary input~\cite{liu2021kg}.
However, the method of explicitly injecting entity representation into PLMs has been argued that the embedding vectors of words in text and entities in KG are obtained in separate ways, making their vector-space inconsistent~\cite{liu2020k}.
The second one is to implicitly modeling knowledge information into PLMs by performing knowledge-related tasks, such as concept order recovering~\cite{zhou2021pre}, entity category prediction~\cite{yu2020jaket}.
% For example, JAKET jointly pre-trained both the KG representation and language representation by adding two self-supervised learning objectives (i.e., entity category prediction, relation type prediction) on KGs~\cite{yu2020jaket}.
For example, CALM proposed a novel contrastive objective for packing more commonsense knowledge into the parameters, and jointly pre-trained both generative and contrastive objectives for enhancing commonsense NLG tasks~\cite{zhou2021pre}.

\vspace{-0.05in}
\subsection{Knowledge-enhanced Learning and Inference}

Besides specialized model architectures, one common way of injecting knowledge to generation models is through the supervised knowledge learning. For example, one can encode knowledge into the objective function that guides the model training to acquire desired model behaviors~\cite{dinan2019wizard,kim2020sequential}. Such approaches enjoy the flexibility of integrating diverse types of knowledge by expressing them as certain forms of objectives. In general, knowledge-enhanced learning is agnostic to the model architecture, and can be combined with the aforementioned architectures.

\begin{figure}[t]
  \begin{center}
    \includegraphics[width=1.0\textwidth]{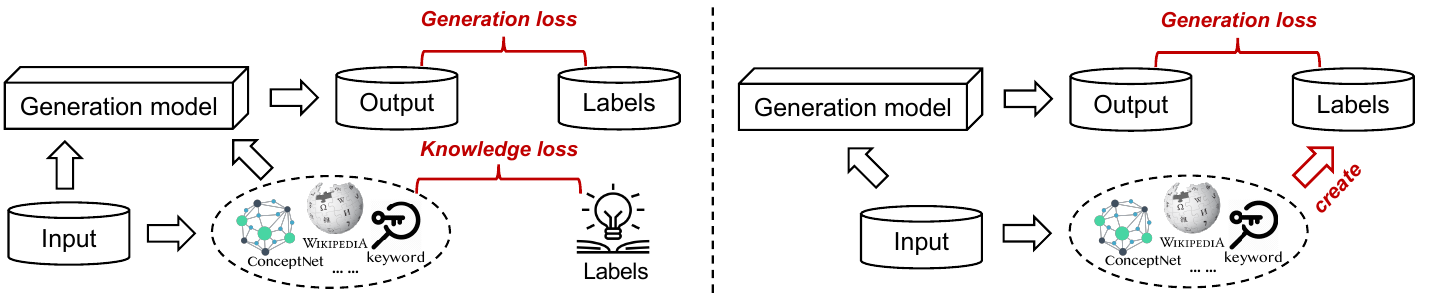}
  \end{center}
  \vspace{-0.1in}
  \caption{Incorporating knowledge into text generation by treating knowledge as the target. The first category of methods (left) combine knowledge-related tasks as auxiliary into the text generation task, resulting in a \textit{multi-task learning} setting. The second category of methods (right) create \textit{weakly-supervised} labels from knowledge, enforcing the relevancy between the knowledge and the target sequence.}
  \label{fig:knowledge-related-task}
\vspace{-0.15in}
\end{figure}

\vspace{-0.05in}
\subsubsection{\textbf{Learning with knowledge-related tasks}}
One could devise learning tasks informed by the knowledge so that the model is trained to acquire the knowledge information.

\vspace{-0.05in}
\paragraph{\textbf{Knowledge as target}}

The methods can be mainly divided into two categories as shown in Figure \ref{fig:knowledge-related-task}.
The first category of knowledge-related tasks creates learning targets based on the knowledge, and the model is trained to recover the targets. These tasks can be combined as auxiliary tasks with the text generation task, resulting in a \emph{multi-task learning} setting.
%is designed with external feedbacks.
%The general idea is to improve the accuracy of acquiring knowledge from external sources. 
For example, knowledge loss is defined as the cross entropy between the predicted and true knowledge sentences, and it is combined with the standard conversation generation loss to enhance grounded conversation~\cite{dinan2019wizard,kim2020sequential}. Similar tasks include keyword extraction loss~\cite{li2020keywords}, template re-ranking loss~\cite{cao2018retrieve,wang2019biset}, 
link prediction loss on knowledge graph~\cite{ji2020language},
path reasoning loss~\cite{liu2019knowledge},
mode loss~\cite{zhou2018commonsense,wu2020topicka}, bag-of-word (BOW) loss~\cite{xu2020neural,lian2019learning}, etc.
The second category of methods directly derive the text generation targets from the knowledge, and use those (typically noisy) targets as supervisions in the standard text generation task. The approach is called \textit{weakly-supervised} learning. Weakly-supervised learning enforces the relevancy between the knowledge and the target sequence.
For example, in the problem of aspect based summarization, the work \cite{tan2020summarizing} automatically creates target summaries based on external knowledge bases, which are used to train the summarization model in a supervised manner.
% For examples, Tan et al. trained an unsupervised text summarization model by creating a target summary based on an external knowledge base~\cite{tan2020summarizing}; Feng et al. trained an unsupervised content manipulation model by selecting similar records and learning the semantic relationship between records and reference~\cite{feng2020learning}.

\vspace{-0.05in}
\paragraph{\textbf{Knowledge as condition}}
The second way of devising knowledge-related tasks is to augment the text generation task by conditioning the generation on the knowledge. That is, the goal is to learn a function $p_{\theta}(Y|X, K)$, where $X$ is the input sequence, $Y$ is the target text and $K$ is the knowledge. Generally, the knowledge $K$ is first given externally (e.g., style, emotion) or retrieved from external resources (e.g., facts from knowledge base, a document from Wikipedia) or extracted from the given input text (e.g., keywords, topic words). Second, a conditional text generation model is used to incorporate knowledge and generate target output sequence. In practice, knowledge is often remedied by soft enforcing algorithms such as attention mechanism~\cite{bahdanau2015neural} and copy/pointing mechanism~\cite{gu2016incorporating,see2017get}.
Regarding knowledge as condition is widely used in knowledge-enhanced text generation. For examples, work has been done in making personalized dialogue response by taking account of persona~\cite{zhang2018personalizing} and emotion~\cite{zhou2018emotional}, controlling various aspects
of the response such as politeness~\cite{niu2018polite}, grounding the responses in external source of knowledge~\cite{zhou2018commonsense,dinan2019wizard,ghazvininejad2018knowledge} and generating topic-coherent sequence~\cite{tang2019target,xu2020neural}.
Besides, using variational autoencoder (VAE) to enforce the generation process conditioned on knowledge is one popular approach to unsupervised NLG.
By manipulating latent space for certain attributes, such as topic~\cite{wang2019topic} and style~\cite{hu2017toward}, the output sequence can be generated with desired attributes without supervising with parallel data.
% Soft-constrained text generation requires the generated sentences to be semantically related to the given constraints, without strictly enforcing the presence of those constraints (e.g., topic words) in the generated content~\cite{qin2019conversing,tang2019target}. On the contrary, hard-constrained text generation refers to the mandatory inclusion of certain keywords in the output sentences. 

\vspace{-0.05in}
\subsubsection{\textbf{Learning with knowledge constraints}}

Instead of creating training objectives in standalone tasks that encapsulate knowledge, another paradigm of knowledge-enhanced learning is to treat the knowledge as the \emph{constraints} to regularize the text generation training objective. 
% \hzt{Move this part to ``Knowledge as the condition''}
% Constraint-driven learning places explicit constraints on independent attribute controls and combines these with differentiable approximation to produce discrete text~\cite{hokamp2017lexically}. 
% In the literature, constraint text generation methods include (i) soft-constrained text generation~\cite{yang2019enhancing,tang2019target} and (ii) hard-constrained text generation~\cite{welleck2019non}, depending on whether inclusion of specified keywords in the output is mandatory. 

% Unlike soft-constrained models which are straightforward to design, the problem of hard-constrained text generation requires the design of complex dedicated neural network architectures such as ~\cite{hokamp2017lexically,welleck2019non,zhang2020pointer}.
 
% \paragraph{Posterior regularization} 
The posterior regularization (PR) framework was proposed to restrict the space of the model posterior on unlabeled data as a way to guide the model towards desired behavior~\cite{ganchev2010posterior,zhu2014bayesian}. PR has been used as a principled framework to impose knowledge constraints on probabilistic models (including deep networks) in general~\cite{hu2018deep,zhang2017prior}.
PR augments any regular training objective $\mathcal{L}(\theta)$ (e.g., negative log-likelihood, as in Eq.\eqref{eq:Seq2Seq-loss}) with a constraint term to encode relevant knowledge. Formally, 
denote the constraint function as $f(X,Y) \in \mathbb{R}$ such that a higher $f(X,Y)$ value indicates a better generated sequence $Y$ that incorporates the knowledge. 
%so a straightforward way to impose the constraint on the model is to maximize $\mathbb{E}p_{\theta}[g(Y)]$. Such method is efficient only when $p_{\theta}$ is a generative adversarial network (GAN) like implicit generative model or an explicit distribution that can be efficiently reparameterized~\cite{hu2018deep}. For other models such as the large set of non-reparameterizable explicit distributions, the gradient $\nabla_\theta\mathbb{E}p_{\theta}[g(Y)]$ is usually computed with the log-derivative trick that may suffer from high variance. 
PR introduces an auxiliary distribution $q(Y|X)$, and imposes the constraint on $q$ by encouraging a large expected $f(X, Y)$ value: $\mathbb{E}q [f(X,Y)]$. Meanwhile, the model $p_{\theta}$ is encouraged to stay close to $q$ through a KL divergence term. The learning problem is thus a constrained optimization:
\begin{align}
    \max_{\theta, q} &\mathcal{L}(\theta) - \mathrm{KL}(q(Y|X)||p_\theta(Y|X)) + \xi \\
    %\text{where}\ &\mathcal{L}_{PR}(\theta, q) = - \mathrm{KL}(q(Y|X)||p_\theta(Y|X)) + \xi\\
    &s.t.~~ \mathbb{E}q [f(X,Y)] > \xi,
\end{align}
where $\xi$ is the slack variable. The PR framework is also related to other constraint-driven learning methods~\cite{chang2007guiding,mann2007simple}. We refer readers to \cite{ganchev2010posterior} for more discussions.

\vspace{-0.05in}
\subsubsection{\textbf{Inference with knowledge constraints}}

Pre-trained language models leverage large amounts of unannotated data with a simple log-likelihood training objective. Controlling language generation by particular knowledge in a pre-trained model is difficult if we do not modify the model architecture to allow for external input knowledge or fine-tuning with specific data~\cite{dathathri2020plug}. Plug and play language model (PPLM) opened up a new way to control language generation with particular knowledge during inference. At every generation step during inference, the PPLM shifts the history matrix in the direction of the sum of two gradients: one toward higher log-likelihood of the attribute
$a$ under the conditional attribute model $p(a|Y)$ and the other toward higher log-likelihood of the unmodified pre-trained generation model $p(Y|X)$ (e.g., GPT). Specifically, the attribute model $p(a|Y)$ makes gradient based updates to $\Delta\textbf{S}_t$ as follows:
\begin{equation}
\Delta \textbf{S}_t \leftarrow \Delta \textbf{S}_t +  \frac{\nabla_{\Delta \textbf{S}_t} \log p(a|\textbf{S}_t + \Delta\textbf{S}_t)}{||\nabla_{\Delta \textbf{S}_t} \log p(a|\textbf{S}_t + \Delta\textbf{S}_t)||^{\gamma} },
\end{equation}
where $\gamma$ is the scaling coefficient for the normalization term; $\Delta \textbf{S}_t$ is update of history matrix $\textbf{S}_t$ (see Eq.(\ref{eq:transformer})) and initialized as zero. The update step is repeated multiple times. Subsequently, a forward pass through the generation model is performed to obtain the updated $\widetilde{\textbf{S}}_{t+1}$ as $\widetilde{\textbf{S}}_{t+1} = \textsc{Decoder}((\textbf{S}_t + \Delta \textbf{S}_t), \textbf{e}(y_t), \textbf{H})$. The perturbed $\widetilde{\textbf{S}}_{t+1}$ is then used to generate a new logit vector. PPLMs is efficient and flexible to combine differentiable attribute models to steer text generation~\cite{qin2020backpropagation}.
% Recent work has followed its idea in other tasks.

\vspace{-0.05in}
\section{NLG enhanced by Internal Knowledge}
\subsection{NLG Enhanced by Topic}
Topic, which can be considered as a representative or compressed form of text, has been often used to maintain the semantic coherence and guide the NLG process.
Topic modeling is a powerful tool for finding the high-level content of a document collection in the form of latent topics~\cite{blei2003latent}.
A classical topic model, Latent Dirichlet allocation (LDA), has been widely used for inferring a low dimensional representation that captures latent semantics of words and documents~\cite{blei2003latent}. In LDA, each topic is defined as a distribution over words and each document as a mixture distribution over topics. LDA generates words in the documents from topic distribution of document and word distribution of topic.
% However, LDA suffers from the limitations of  imprecise prior knowledge and restricted unigram topic representation.
% A variety of generative models were designed to infer latent topics~\cite{chong2009simultaneous,mcauliffe2008supervised}.
Recent advances of neural techniques open a new way of learning low dimensional representations of words from the tasks of word prediction and context prediction, making neural topic models become a popular choice of finding latent topics from text~\cite{cao2015novel,guo2020recurrent}.

\begin{figure}[t]
  \begin{center}
    \includegraphics[width=0.95\textwidth]{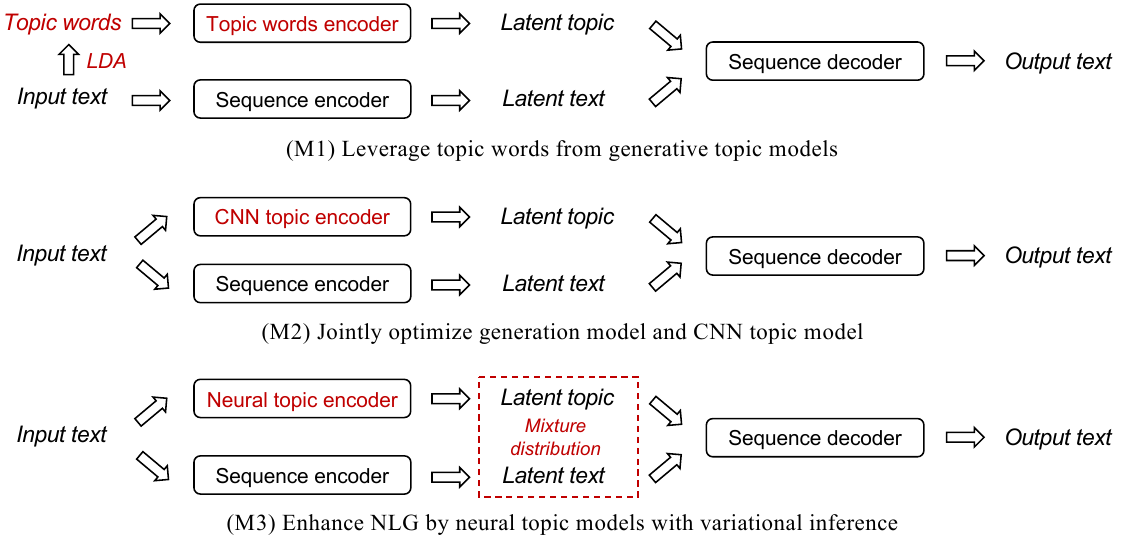}
  \end{center}
  \vspace{-0.15in}
  \caption{Three typical methodologies for incorporating topics into NLG. Detailed designs are not included.}
  \label{fig:topic}
  \vspace{-0.05in}
\end{figure}

Next, we introduce popular NLG applications enhanced by topics:
\begin{itemize}
    \item \textbf{Dialogue system.} A vanilla Seq2Seq often  generates trivial or non-committal sentences of frequent words or phrases in the corpus~\cite{xing2017topic}. For example,
    a chatbot may say \textit{``I do not know''}, \textit{``I see''} too often. Though these off-topic responses are safe to reply to many queries, they are boring with very little information. Such responses may quickly lead the conversation to an end, severely hurting user experience. Thus, on-topic response generation is highly needed.
    \item \textbf{Machine translation.} Though the input and output languages are different (e.g., translating English to Chinese), the contents are the same, and globally, under the same topic. Therefore, topic can serve as an auxiliary guidance to preserve the semantics information of input text in one language into the output text in the other language. 
    \item \textbf{Paraphrase.} Topic information helps understand the potential meaning and determine the semantic range to a certain extent. Naturally, paraphrases concern the same topic, which can serve as an auxiliary guidance to promote the preservation of source semantic.
\end{itemize}

As shown in Figure \ref{fig:topic}, we summarize topic-enhanced NLG methods into three methodologies: (M1) leverage topic words from generative topic models; (M2) jointly optimize generation model and CNN topic model; (M3) enhance NLG by neural topic models with variational inference.

\vspace{-0.05in}
\subsubsection{\textbf{M1: Leverage Topic Words from Generative Topic Models}}
\label{sec:topic-m1}

Topics help understand the semantic meaning of sentences and determine the semantic spectrum to a certain extent. 
% For example, in human-human conversations, people might first select topically related concepts (e.g., sport, food) in their minds, then organize content and select words to respond.
To enhanced text generation, an effective solution is to first discover topics using generative topic models (e.g., LDA), and then incorporate the topics representations into neural generation models, as illustrated in Figure \ref{fig:topic}(a). In existing work, there are two mainstream methods to represent topics obtained from generative topic models. The first way is to use the generated topic distributions for each word (i.e., word distributions over topics) in the input sequence~\cite{zhang2016topic,narayan2018don}. The second way is to assign a specific topic to the input sequence, then picks the top-$k$ words with the highest probabilities under the topic, and use word embeddings (e.g., GloVe) to represent topic words~\cite{xing2017topic,liu2019generating}. Explicitly making use of topic words can bring stronger guidance than topic distributions, but the guidance may deviate from the target output sequence when some generated topic words are irrelevant.
Zhang et al. proposed the first work of using a topic-informed Seq2Seq model by concatenating the topic distributions with encoder and decoder hidden states~\cite{zhang2016topic}.
Xing et al. designed a topic-aware Seq2Seq model in order to use topic words as prior knowledge to help dialogue generation~\cite{xing2017topic}. 
% Specifically, topic words are obtained from a pre-trained LDA model.
% the word distribution of topics in the input sequence are obtained from a pre-trained LDA model. To obtain topic words, Topic-Seq2Seq uses the LDA model to assign a topic $t$ to the input sequence $X$, then picks the top $k$ words with the highest probabilities under $t$. 
% Here, we use $\{\textbf{p}_1, \cdots , \textbf{p}_k\}$ to denote the embeddings of chosen topic words~\cite{xing2017topic}.
% At the decoding phase, each word is generated according to both the input sequence and topic words through a joint attention mechanism (details in Table \ref{tab:mul-source-attn}). It summarizes context vectors from input sequence attention and topic attention simultaneously. 
% The probability of generating a target word is a combination of the probabilities of two modes: generate-mode and topic-mode (details in Table \ref{tab:mode}). 

\begin{table*}[t]
\centering
\caption{Natural language generation methods that incorporate topic knowledge in text generation. Since most of the methods are tested on different tasks and datasets, we only compare the performance between ``w/o topic'' setting and ``with topic'' setting. For evaluation metrics, PPL is short for perplexity (lower is better); B-4 is short for BLEU-4 (higher is better); R-L is short for ROUGE-L (higher is better).}
\label{tab:topicinseq2seq}
\vspace{-0.1in}
{\scalebox{0.82}{\begin{tabular}{|l|l|c|c|l|l|l|l|l|}
\hline
\multirow{2}*{Task} & \multirow{2}*{Method} & \multirow{2}*{Ref.} & \multirow{2}*{Cat.} & \multicolumn{2}{c|}{Framework components} & \multicolumn{3}{c|}{Effect of topic modeling} \\
\cline{5-9}
& & & & Seq. Enc/Dec & Topic model & Dataset & w/o topic & with topic \\
\hline \hline
\multirow{2}*{\makecell[l]{Dialogue \\ system}} & Tp-S2S & \cite{xing2017topic} & M1 & RNN Seq2Seq & LDA topics & Baidu Tieba & (PPL) 147.0 & (PPL) 134.6\\
% \cline{2-8}
& PEE & \cite{xu2020neural} & M3 & RNN Seq2Seq & Neural topics & PersonaChat& (B-4) 2.98 & (B-4) 3.56 \\ % Topic-Net
\hline \hline
\multirow{2}*{\makecell[l]{Machine \\ translation}} & Tp-NMT & \cite{zhang2016topic} & M1 & RNN Seq2Seq & LDA topics & NIST & (B-4) 34.76 & (B-4) 35.91 \\
% \cline{2-8}
& BLT-NMT & \cite{wei2019translating} & M2 & RNN Seq2Seq & CNN topics & NIST & (B-4) 38.97 & (B-4) 40.10 \\
\hline \hline
\multirow{3}*{\makecell[l]{Summari\\ -zation}} & Tp-CS2S & \cite{narayan2018don} & M1 & CNN Seq2Seq & LDA topics & XSum & (R-L) 25.23 & (R-L) 25.75 \\
& TGVAE & \cite{wang2019topic} & M3 & RNN with VAE & Neural topics & Gigawords & (R-L) 32.13 & (R-L) 33.02 \\
& VHTM & \cite{fu2020document} & M3 & RNN with VAE & Neural topics & CNN/DM & (R-L) 36.73 & (R-L) 37.18 \\
\hline \hline
\multirow{2}*{Paraphrase} & TGLM & \cite{gao2019topic} & M2 & RNN Seq2Seq & CNNs topics & Yahoo! Ans & (PPL) 99.13 & (PPL) 88.69 \\
& PTA & \cite{liu2019generating} & M1 & RNN Seq2Seq & LDA topics & Quora & (B-4) 28.76 & (B-4) 31.75 \\
\hline
\end{tabular}}}
\vspace{-0.1in}
\end{table*}

\vspace{-0.05in}
\subsubsection{\textbf{M2: Jointly Optimize Generation Model and CNN Topic Model}}
\label{sec:topic-m2}

% LDA models are usually unsupervised with an assumption that the word distributions of topics are Dirichlet distributions. However, LDA models may not be able to find proper topics that the target task (e.g., NLG) requires.
% Also, 
The LDA models were separated from the training process of neural generation model and were not able to adapt to the diversity of dependencies between input and output sequences. 
Therefore, the idea of addressing this issue is to use neural topic models. 
Convolutional neural networks (CNN) were used to
% LDA may fail to provide high-quality topics, since it takes topics as a dirchlet distribution, hence, the topics might be weakly correlated. In addition, a pre-trained LDA topic model cannot be further updated through the learning process, which apparently does not hold for all of input sequences fetched from diverse sources. Convolutation neural network could be used to
learn latent topic representations through iterative convolution and pooling operations.
% \cite{kim2014convolutional}. 
There are growing interests of using the CNNs to map latent topics implicitly into topic vectors that can be used to enhance text generation tasks~\cite{wei2019translating,gao2019topic}. Empirical analyses showed that convolution-based topic extractors could outperform LDA-based topic models for multiple applications (e.g., dialogue system, text summarization, machine translation). However, theoretical analysis was missing to ensure the quality of the topics captured by the convolutions. And their interpretability is not as satisfactory as the LDA-based topic models.
% it still suffers certain limitations as follows. First, it lacks theoretical proof to show convolutional neural network can accurately extract the topic information from sequences/documents. Second, 
% it lacks interpretability comparing with LDA-based topic models because it cannot capture topic assignments explicitly.

\vspace{-0.05in}
\subsubsection{\textbf{M3: Enhance NLG by Neural Topic Models with Variational Inference}}
\label{sec:topic-m3}

Neural topic models can be trained efficiently by backpropagation~\cite{cao2015novel}. In neural topic models, Dirichlet distributions can be employed as the prior to generate the parameters of the multinomial distribution $\theta_d$ for each document~\cite{miao2017discovering}. The generative process of LDA is represented as: (1) $\theta_d \sim \mathrm{Dirichlet} (\alpha)$; (2) $t_i \sim \mathrm{Multinomial} (\theta_d)$; (3) $w_i \sim \mathrm{Multinomial} (\beta_{t_i})$,
% \begin{align}
%     \theta_d & \sim \mathrm{Dirichlet} (\alpha); ~
%     t_i \sim \mathrm{Multinomial} (\theta_d); ~
%     w_i \sim \mathrm{Multinomial} (\beta_{t_i}),
% \end{align}
where $d$ denotes the bag-of-words representation of a document, $t_i$ represents the topic assignment for word $w_i$, and $\beta_{t_i}$ represents the topic distribution over words given topic assignment $t_i$. 
% The marginal likelihood for document $d$ is:
% \begin{equation}
%     p(d|\alpha, \beta)= \int_{\theta} p(\theta|\alpha) \big{(} \prod_{i=1}^{n} \sum_{t_i} p(w_i|\beta_{t_i}) p(t_i|\theta)  \big{)} d\theta = \int_\theta p(\theta|\alpha) p(d|\beta, \theta) d\theta.
%     \label{eq:dir}
% \end{equation}
However, a directed generative model comes up against the problem of establishing low variance gradient estimators. Miao et al. parameterized the multinomial distributions with neural networks and jointly learned the model parameters via variational inference~\cite{miao2017discovering}. They created neural structures for constructing topic distributions conditioned on a draw from a multivariate Gaussian distribution, represented as 
% Then, the generative process could be modeled as:
% \begin{equation}
$\theta_d\sim\mathrm{G} (\mu_0, \sigma_0^2)$,
% \end{equation}
where $\mathrm{G} (\mu_0, \sigma_0^2)$ is composed of a neural network conditioned on an isotropic Gaussian $\mathrm{N} (\mu_0, \sigma_0^2)$. 
% Now, the marginal likelihood for document $d$ is:
% \begin{equation}
%     p(d|\beta, \mu_0, \sigma_0^2)= \int_{\theta} p(\theta| \mu_0, \sigma_0^2) \big{(} \prod_{i=1}^{n} \sum_{t_i} p(w_i|\beta_{t_i}) p(t_i|\theta)  \big{)} d\theta = \int_\theta p(\theta|\mu_0, \sigma_0^2) p(d|\beta, \theta) d\theta.
% \end{equation}
% Compared with Eq.(\ref{eq:dir}), the latent variable $\theta$ is parameterized by a neural network conditioned on a draw from a Gaussian distribution. 
% To carry out neural variational inference, Miao et al. constructed an inference network $q(\theta|\mu(d), \sigma(d))$ to approximate the posterior $p(\theta|d)$, where $\mu(d)$ and $\sigma^2(d)$ are functions of $d$. The network was implemented by a multi-layer perceptron (MLP)~\cite{miao2017discovering}. 
Taking a Gaussian prior distribution makes re-parameterization feasible to build an unbiased and low-variance gradient estimator for the variational distribution~\cite{diederik2014auto}. Without conjugacy prior, the updates of the parameters are derived directly and easily from the variational lower bound. Formally, a variational lower bound
% (aka., evidence lower bound (ELBO)) 
for the document log-likelihood is:
\begin{equation}
    \setlength\abovedisplayskip{2pt} 
    \setlength\belowdisplayskip{2pt}
    \mathcal{J}_{topic} = \mathbb{E}_{q(\theta|d)} [\log p(d|\beta, \theta)] - \mathrm{KL} (q(\theta|d)||p(\theta |\mu_0, \sigma^2_0)),
\end{equation}
where $q(\theta|d)$ is the variational distribution approximating the true posterior $p(\theta|d)$. Its lower bound is estimate by sampling $\theta$ from $q(\theta|d) = \mathrm{G}(\theta|\mu(d), \sigma^2(d))$. 

In order to combine neural topic model and neural generation model, the idea is to use the Variational Auto-Encoder (VAE)~\cite{diederik2014auto}. It adopts autoregressive networks (e.g., LSTM) both as the encoder and decoder. VAE can learn latent codes $z$ of texts by reconstructing texts with its decoder. It assumes that the generation process is controlled by codes in a continuous latent space. This kind of VAE implementation considers sequential information of texts that can model the linguistic structure of texts. Wang et al. proposed topic guided variational autoencoder (TGVAE), to draw latent code $z$ from a topic-dependent Gaussian Mixture Prior in order to incorporate the topical knowledge into latent variables~\cite{wang2019topic}. The topic-dependent Gaussian Mixture Model (GMM) is defined as:
% \begin{equation}
$p(z|\beta, t) = \sum^{T}_{i=1} t_i \mathrm{N} (\mu(\beta_i), \sigma^2(\beta_i))$,
% \end{equation}
where $T$ is the number of topics, $\mu(d)$ and $\sigma^2(d)$ are functions implemented by MLP. TGVAE uses bag-of-words as input and embeds an input document into a topic vector. The topic vector is then used to reconstruct the bag-of-words input, and the learned topic distribution over words is used to model a topic-dependent prior to generate an output sequence $Y$ from conditioned on an input sequence $X$. 
% Specifically, the joint marginal likelihood can be written as:
% \begin{equation}
%     p(Y,d|X) = \int_\theta \int_z p(\theta) p(d|\beta, \theta) p(z|\beta, \theta) p (Y|X, z) d\theta dz.
% \end{equation}
Therefore, to maximize the log-likelihood log $p(Y,d|X)$, a variational objective function is constructed as:
\begin{equation}
    \setlength\abovedisplayskip{2pt} 
    \setlength\belowdisplayskip{2pt}
    % \mathcal{J}_{topic} & = \mathbb{E}_{\theta \sim q(\theta|d)} [\log p(d|\beta, \theta)] - KL (q(\theta|d)||p(\theta |\mu_0, \sigma^2_0)),   \\ 
    \mathcal{J}_{seq2seq} = \mathbb{E}_{q(z|X)} [\log p(Y|X, z)] - \mathbb{E}_{q(\theta|d)} [\mathrm{KL} (q(z|X)||p(z|\beta, \theta))],
\end{equation}
% where $q(\theta|d)$ and $q(z|X)$ are the variational distributions for $\theta$ and $z$, respectively. 
where $q(z|X)$ is variational distributions for $z$. The combined object function is given by:
\begin{equation}
    \setlength\abovedisplayskip{2pt} 
    \setlength\belowdisplayskip{2pt}
    \mathcal{J} = \mathcal{J}_{topic} +  \mathcal{J}_{seq2seq}.
\end{equation}

\subsubsection{\textbf{Discussion and Analysis of Different Methods}} 

\vspace{-0.05in}

For \textbf{M1}, topic models (e.g., LDA) has a strict probabilistic explanation since the semantic representations of both words and documents are combined into a unified framework. Besides, topic models can be easily used and integrated into generation frameworks. For example, topic words can be represented as word embeddings; topic embeddings can be integrated into the decoding phase through topic attention. However,
% since LDA models assumes the word distributions of topics are Dirichlet distributions, so they may fail to find proper topics that the target task (i.e., NLG) requires. Besides,
LDA models are separated from the training process of generation, so they cannot adapt to the diversity of dependencies between input and output sequences.

For \textbf{M2}, it is an end-to-end neural framework that simultaneously learns latent topic representations and generates output sequences. Convolutional neural networks (CNN) are often used to generate the latent topics through iterative convolution and pooling operations. However, theoretical analysis is missing to ensure the quality of the topics captured by the convolutions. And their interpretability is not as good as the LDA-based topic models.

For \textbf{M3}, neural topic models combine the advantages of neural networks and probabilistic topic models. They enable back propagation for joint optimization, contributing to more coherent topics, and can be scaled to large data sets. Generally, neural topic models can provide better topic coherence than LDAs~\cite{cao2015novel,wang2019topic,xu2020neural}. However, neural variational approaches share a same drawback that topic distribution is assumed to be an isotropic Gaussian, which makes them incapable of modeling topic correlations. Existing neural topic models assume that the documents should be i.i.d. to adopt VAE, while they are commonly correlated. The correlations are critical for topic modeling.
\vspace{-0.05in}
\subsection{NLG Enhanced by Keywords}
Keyword (aka., key phrase, key term) is often referred as a sequence of one or more words, providing a compact representation of the content of a document. The mainstream methods of keyword acquisition for documents can be divided into two categories~\cite{siddiqi2015keyword}: keyword assignment and keyword extraction. Keyword assignment means that keywords are chosen from a controlled vocabulary of terms or predefined taxonomy.
Keyword extraction selects the most representative words explicitly presented in the document, which is independent from any vocabulary. Keyword extraction techniques (e.g., TF-IDF, TextRank, PMI) have been widely used over decades. Many NLG tasks can benefit from incorporating such a condensed form of essential content in a document to maintain the semantic coherence and guide the generation process.

Next, we introduce popular NLG applications enhanced by keywords:

\begin{itemize}
    \item \textbf{Dialogue system.} Keywords help enlighten and drive the generated responses to be informative and avoid generating universally relevant replies which carry little semantics. Besides, 
    % personalized dialogue, as an emerging topic, plays an important role for improving user experience in human-computer interaction. 
    % Being empathetic is a necessary step for the dialogue agent to be perceived as a social character by users. 
    recent work introduced personalized information into the generation of dialogue to help deliver better dialogue response such as emotion ~\cite{li2018syntactically,zhou2018emotional,song2019generating}, and persona~\cite{zhang2018personalizing,zheng2020pre}.
    \item \textbf{Summarization.} 
    % It is popular for displaying a document summary in a coherent form that is easily readable and grammatically correct. However, it 
    Vanilla Seq2Seq models often suffer when the generation process is hard to control and often misses salient information~\cite{li2018guiding}. Making use of keywords as explicit guidance can provide significant clues of the main points about the document~\cite{li2018guiding,li2020keywords}. It is closer to the way that humans write summaries: make sentences to contain the keywords, and then perform necessary modifications to ensure the fluency and grammatically correctness.
    \item \textbf{Question generation.} It aims to generate questions from a given answer and its relevant context. Given an answer and its associated context, it is possible to raise multiple questions with different focuses on the context and various means of expression.
\end{itemize}

\begin{table}[t]
\caption{Natural language generation methods that incorporate keyword in text generation.}
\vspace{-0.1in}
\label{tab:keyword}
\begin{subtable}[t]{0.98\textwidth}
\centering
\caption{(M1) Descriptions and quantitative comparisons between three methods for emotional dialogue systems.}
\label{tab:emotion}
\vspace{-0.05in}
{\scalebox{0.82}{\begin{tabular}{|l|l|c|l|c|c|c|}
\hline
\multirow{2}*{Task} & \multirow{2}*{Method} & \multirow{2}*{Ref.} & \multirow{2}*{Assignment method} & \multicolumn{3}{c|}{Experiments on NLPCC dataset} \\
\cline{5-7}
& & & & BLEU & D-1/D-2 & Emotion w/s \\
\hline \hline
\multirow{4}*{\makecell[l]{Dialogue \\ system}} & Seq2Seq & \cite{bahdanau2015neural} & Seq2Seq attention \textit{without} using keywords & 1.50 & 0.38/1.20 & 33.5/37.1 \\
& E-SCBA & \cite{li2018syntactically} & MLP classifier to 7 emotions (categories) & 1.69 & 0.54/4.84 & 72.0/51.2 \\
& EmoChat & \cite{zhou2018emotional} & E-SCBA + two memory modules for decoding & 1.68 & 0.90/7.35 & 76.5/58.0 \\ 
& EmoDS & \cite{song2019generating} & MLP classifier after decoding (discriminator) & 1.73 & 1.13/8.67 & 81.0/68.7 \\
\hline
\end{tabular}}}
\end{subtable}
\begin{subtable}[t]{0.98\textwidth}
\centering
\vspace{0.1in}
\caption{(M2) As most methods are tested on different tasks and datasets, we only compare the performance between ``w/o keyword'' setting and ``with keyword'' setting. Besides, HM is short for human evaluation.}
\label{tab:hm}
\vspace{-0.05in}
{\scalebox{0.82}{\begin{tabular}{|l|l|c|l|l|l|l|l|}
\hline
\multirow{2}*{Task} & \multirow{2}*{Method} & \multirow{2}*{Ref.} & Extraction & Keyword & \multicolumn{3}{c|}{Effect of keyword} \\
\cline{6-8}
& & & method & labels & Dataset & w/o keyword & with keyword \\
% \hline \hline
% \multirow{2}*{\makecell[l]{Dialogue \\ system}} & \multirow{2}*{Seq2BF} & \multirow{2}*{\cite{mou2016sequence}} & \multirow{2}*{PMI} & \multirow{2}*{Unsupervised} & \multirow{2}*{Baidu Tieba} & \multirow{2}*{(HM) 0.58} & \multirow{2}*{(HM) 0.67} \\ 
% & & & & & & & \\ 
\hline \hline
\multirow{4}*{\makecell[l]{Summari- \\ zation}} & \multirow{2}*{\makecell[l]{KIGN}} & \multirow{2}*{\makecell[l]{\cite{li2018guiding}}} & \multirow{2}*{\makecell[l]{TextRank}} & \multirow{2}*{\makecell[l]{Unsupervised}} & CNN/DM & (R-2) 15.66 & (R-2) 17.12 \\ 
& & & & & Gigaword & (R-2) 23.61 & (R-2) 23.93 \\ 
\cline{2-8} 
& ComGen & \cite{li2019coherent} & PMI and TFIDF & Unsupervised & Tencent & (HM) 5.77 & (HM) 7.19 \\ 
% & G-S2A & \cite{han2019inferring} & KExt, KGraphE & Unsupervised, no labels \\ 
& KGAS & \cite{li2020keywords} & BiLSTM-Softmax & $\text{w}(X) \cap \text{w}(Y)$ & Gigaword & (R-2) 23.61 & (R-2) 25.06 \\
\hline \hline
\multirow{2}*{\makecell[l]{Question \\ generation}} & Selector & \cite{cho2019mixture} & BiLSTM-Softmax & $\text{w}(X) \cap \text{w}(Y)$ & SQuAD & (B-4) 14.72 & (B-4) 15.87 \\
& Prior & \cite{wang2020diversify} & BiLSTM-Softmax & $\text{w}(X) \cap \text{w}(Y)$ & SQuAD & (B-4) 14.72 & (B-4) 15.34 \\
\hline
\end{tabular}}}
\end{subtable}
\vspace{-0.05in}
\end{table}

Researchers have developed a great line of keyword-enhanced NLG methods. These methods can be categorized into two methodologies: (M1) Incorporate keyword assignment into text generation; (M2) Incorporate keyword extraction into text generation.

\vspace{-0.05in}
\subsubsection{\textbf{M1: Incorporate Keyword Assignment into Text Generation}}
\label{sec:keywork-m1}

When assigning a keyword to an input document, the set of possible keywords is bounded by a pre-defined vocabulary~\cite{siddiqi2015keyword}. 
% One advantage is that the quality of keywords is guaranteed, because irrelevant keywords would not be included in the vocabulary. 
The keyword assignment is typically implemented by a classifier that maps the input document to a word in the pre-defined vocabulary~\cite{choudhary2017domain,li2018syntactically,zhou2018emotional,song2019generating}. 
% Another advantage is that even if two semantically similar documents do not have common words, they can still be assigned with the same keyword. 
Unfortunately, some NLG scenarios do not hold an appropriate pre-defined vocabulary, so keyword assignment cannot be widely used to enhance NLG tasks.
One applicable scenario is to use a pre-determined domain specific vocabulary to maintain relevance between the input and the output sequence~\cite{choudhary2017domain}. Another scenario is to generate dialogue with specific attributes such as persona~\cite{song2019exploiting,xu2020neural}, emotion~\cite{li2018syntactically,zhou2018emotional,song2019generating}. 

\vspace{-0.05in}
\paragraph{\textbf{M1.1: Adding assigned keyword into the decoder}}

A straightforward method of keyword assignment is to assign the words from pre-defined vocabulary and use them as the keywords~\cite{song2019exploiting,xu2020neural}.
Sometimes, the input sequence does not have an explicit keyword, but we can find one from the pre-defined vocabulary. For example, a dialogue utterance \emph{``If you had stopped him that day, things would have been different.''} expresses sadness but it does not have the word ``sad.''
To address this issue, Li et al. propose a method to predict an emotion category by fitting the sum of hidden states from encoder into a classifier~\cite{li2018syntactically}.
Then, the response will be generated with the guidance of the emotion category. 
In order to dynamically track how much the emotion is expressed in the generated sequence, Zhou et al. propose a memory module to capture the emotion dynamics during decoding~\cite{zhou2018emotional}.
Each category is initialized with an emotion state vector before the decoding phase starts. At each step, the emotion state decays by a certain amount. Once the decoding process is completed, the emotion state decays to zero, indicating that the emotion is completely expressed. 

\vspace{-0.05in}
\paragraph{\textbf{M1.2: Assigning keyword for generated sequence}}
As mentioned in~\cite{song2019generating}, explicitly incorporating emotional keywords suffers from expressing a certain emotion overwhelmingly. Instead, Song et al. propose to increase the intensity of the emotional experiences not by using emotional words explicitly, but by implicitly combining neutral words in distinct ways on emotion~\cite{song2019generating}. Specifically, they use an emotion classifier to build a sentence-level emotion discriminator, which helps to recognize the responses that express a certain emotion but not explicitly contain too many literal emotional words. The discriminator is connected to the end of the decoder.
% (i.e., the predicted output sequence $\hat{Y}$).  Overall, the generation loss of assigning keyword for generated sequence is: $\mathcal{L}_{KA-NLL}(\theta) = - \log(p(k|\hat{Y})) - \sum^m_{t=1} \log \left(p(y_t|y_{<t}, X) \right).$

\vspace{-0.05in}
\subsubsection{\textbf{M2: Incorporate Keyword Extraction into Text Generation}}
\label{sec:keywork-m2}

Keyword extraction selects salient words from input documents~\cite{siddiqi2015keyword}.
% So, there is no need to create or maintain vocabularies.
% The primary advantage is important keywords that occur in the document can be selected.
Recent work has used statistical keyword extraction techniques (e.g., PMI~\cite{li2019coherent}, TextRank~\cite{li2018guiding}), and neural-based keyword extraction techniques (e.g., BiLSTM~\cite{li2020keywords}).
% We denote the process of keyword extraction as $K=C(X)$, where $K$ is a set of extracted keywords and $C(\cdot)$ is a function of keyword extraction. 
The process of incorporating extracted keywords into generation is much like the process discussed in Section \ref{sec:keywork-m1}. It takes keywords as an additional input into decoder. Recent work improves encoding phase by adding another sequence encoder to represent keywords~\cite{li2018guiding,li2020keywords}.
Then, the contextualized keywords representation is fed into the decoder together with input sequence representation.
To advance the keyword extraction, Li et al. propose to use multi-task learning for training a keyword extractor network and generating summaries~\cite{cho2019mixture,li2020keywords}. Because both summarization and keyword extraction aim to select important information from input document, these two tasks can benefit from sharing parameters to improve the capacity of capturing the gist of the input text. 
In practice, they take overlapping words between the input document and the ground-truth summary as keywords, and adopt a BiLSTM-Softmax as keyword extractor. Similar idea has also been used in question generation tasks~\cite{cho2019mixture,wang2020diversify}. They use overlapping words between the input answer context and the ground-truth question as keywords.
% Overall, the generation loss is written as:
% $\mathcal{L}_{KE-NLL}(\theta) = - \sum_{k \in K} \log(p(k|X)) - \sum^m_{t=1} \log \left(p(y_t|y_{<t}, X, K) \right)$.

\subsubsection{\textbf{Discussion and Analysis of Different Methods}}

\vspace{-0.05in}
\paragraph{\textbf{Pros and cons.}}
For \textbf{M1}, the primary advantage of keyword assignment is that the quality of keywords is guaranteed, because irrelevant keywords are not included in the pre-defined vocabulary. Another advantage is that even if two semantically similar documents do not have common words, they can still be assigned with the same keyword. However, there are mainly two drawbacks. On one hand, it is expensive to create and maintain dictionaries in new domains. So, the dictionaries might not be available. On the other hand, potential keywords occurring in the document would be unfortunately ignored if they were not in the vocabulary. Therefore, keyword assignment is suitable for the task that requires specific categories of keywords to guide the generated sentences with these key information. For example, dialogue systems generate responses with specific attitudes.

For \textbf{M2}, keyword extraction selects the most representative words explicitly presented in the document, which is independent from any vocabulary.
It is easy to use but has two drawbacks.
% So, keyword extraction is easy to use. The drawbacks of using keyword extraction lie in two aspects.
First, it cannot guarantee consistency because similar documents may still be represented by different keywords if they do not share the same set of words. Second, when an input document does not have a proper representative word, and unfortunately, the keyword extractor selects an irrelevant word from the document as a keyword, this wrong guidance will mislead the generation.
Therefore, keyword extraction is suitable for the task that the output sequence needs to keep important information in the input sequence such as document summarization and paraphrase.

\vspace{-0.05in}
\paragraph{\textbf{Quantitative analysis.}} Table \ref{tab:keyword} summarizes tasks and datasets used in keyword-enhanced NLG work. Comparing with keyword-enhanced methods (E-SCBA~\cite{li2018syntactically}) and the basic Seq2Seq attention model, keyword-enhanced methods can greatly improve both generation quality (evaluated by BLEU) and emotional expression (evaluated by emotion-w and emotion-s) on the NLPCC dataset. Besides, as shown in Table \ref{tab:keyword}(a), EmoDS~\cite{song2019generating} achieved the best performance among three M1 methods, which indicates taking keyword assignment as a discriminant task can make better improvement than assigning keyword before the sentence decoding. For M2 methods, since most methods were evaluated on different tasks, we can only compare the performance between ``without using keyword'' and ``using keyword''. As shown in Table \ref{tab:keyword}(b), leveraging extracted keywords from input sequence into Seq2Seq model can improve the generation quality on summarization and question generation tasks. Comparing with KGAS~\cite{li2020keywords} and KIGN~\cite{li2018guiding}, we can observe using BiLSTM-Softmax to extract keyword (a supervised manner by using overlapping words between $X$ and $Y$ as labels) can make better performance than using TextRank (an unsupervised manner).

% First, all the datasets in the table are public, and we include their links in Table \ref{tab:code}. CommonGen~\cite{lin2020commongen}, ComVE~\cite{wang2020semeval} and $\alpha$-NLG~\cite{bhagavatula2020abductive} have a public leaderboard for competition. 
% Second, for KG sources, we observe that eight (57.1\%) papers use ConceptNet as external resource, while six (42.9\%) papers constructed their own KGs from domain-specific corpus. For example, Koncel et al. created a scientific knowledge graph by applying the SciIE tool (science domain information extraction)~\cite{koncel2019text}. 
% Besides, Zhao et al. compared the performance of models between using ConceptNet and using a self-built KG, and found the model with self-built KG could work better on story generation and review generation tasks~\cite{zhao2020graph}.
% Third, we observed that KG-enhanced NLG methods made the largest improvement on generative commonsense reasoning tasks, in which the average improvement is +2.55\% in terms of $\Delta$BLEU, while the average improvement on all different tasks is +1.32\%.

\vspace{-0.05in}
\subsection{NLG Enhanced by Linguistic Features}

Feature enriched encoder means that the encoder not only reads the input sequence, but also incorporates auxiliary hand-crafted features~\cite{zhou2017neural,sennrich2016linguistic,yu2021sentence}. Linguistic features are the most common hand-crafted features, such as part-of-speech (POS) tags, dependency parsing, and semantic parsing.

\vspace{-0.1in}
\subsubsection{\textbf{POS tags and NER tags}}
Part-of-speech tagging (POS) assigns token tags to indicate the token's grammatical categories and part of speech such as \textit{noun (N)}, \textit{verb (V)}, \textit{adjective (A)}. Named-entity recognition (NER) classifies named entities mentioned in unstructured text into pre-defined categories such as \textit{person (P)}, \textit{location (L)}, \textit{organization (O)}. 
% Commonly used tools include NLTK~\cite{loper2002nltk} and 
CoreNLP is the most common used tool~\cite{manning2014stanford}. In spite of homonymy and word formation processes, the same surface word form may be shared between several word types. Incorporating NER tags and POS tags can detect named entities and understand input sequence better, hence, further improve NLG~\cite{zhou2017neural,nallapati2016abstractive,dong2021inject}.

\vspace{-0.05in}
\subsubsection{\textbf{Syntactic dependency graph}}
\label{sec:syn-graph}

Syntactic dependency graph is a directed acyclic graph representing syntactic relations between words~\cite{bastings2017graph}. For example, in the sentence ``The monkey eats a banana'', ``monkey'' is the subject of the predicate ``eats'', and ``banana'' is the object.
% Besides utilizing the dependency labels of words, the dependency tree structures can also be utilized.
Enhancing sequence representations by utilizing dependency information captures source long-distance dependency constraints and parent-child relation for different words~\cite{chen2018syntax,bastings2017graph,aharoni2017towards}. 
In NLG tasks, dependency information is often modeled in three different ways as follows: (i) linearized representation: linearize dependency graph and then use sequence model to obtain syntax-aware representation ~\cite{aharoni2017towards}; (ii) path-based representation: calculate attention weights based on the linear distance between a word and the aligned center position, i.e., the greater distance a word to the center position on the dependency graph is, the smaller contribution of the word to the context vector is~\cite{chen2018syntax}; and (iii) graph-based representation: use GNNs to aggregate information from dependency relations~\cite{bastings2017graph}. 

\vspace{-0.05in}
\subsubsection{\textbf{Semantic dependency graph}}
\label{sec:sem-graph}

Semantic dependency graph represents \textit{predicate-argument} relations between content words in a sentence and have various semantic representation schemes (e.g., DM) based on different annotation systems.
Nodes in a semantic dependency graph are extracted by semantic role labeling (SRL) or dependency parsing, and connected by different intra-semantic and inter-semantic relations~\cite{pan2020semantic}.
Since semantic dependency graph introduces a higher level of information abstraction that captures commonalities between different realizations of the same underlying predicate-argument structures, it has been widely used to improve text generation~\cite{pan2020semantic,jin2020semsum,liao2018abstract}. Jin et al. propose a semantic dependency guided summarization model~\cite{jin2020semsum}. They incorporate the semantic dependency graph and the input text by stacking encoders to guide summary generation process. The stacked encoders consist of a sequence encoder and a graph encoder, in which the sentence encoder first reads the input text through stacked multi-head self-attention, and then the graph encoder captures semantic relationships and incorporates the semantic graph structure into the contextual-level representation. 
% Some recent work leverages abstract meaning representation (AMR) as a structured semantic representation to improving text generation performance~\cite{liao2018abstract}. Compared with semantic roles, AMR is able to directly capture entity relations and abstract away inflections and function words. 

\vspace{-0.05in}
\subsection{NLG Enhanced by Open Knowledge Graphs}
\label{sec:openkg}

For those KGs (e.g., ConceptNet) constructed based on data beyond the input text, we refer them as \textit{external KGs}. On the contrary, an \textit{internal KG} is defined as a KG constructed solely based on the input text. In this section, we will discuss incorporating internal KG to help NLG~\cite{huang2020knowledge,fan2019using}.

Internal KG plays an important role in understanding the input sequence especially when it is of great length. By constructing an internal KG intermediary, redundant information can be merged or discarded, producing a substantially compressed form to represent the input document~\cite{fan2019using}. Besides, representations on KGs can produce a structured summary and highlight the proximity of relevant concepts, when complex events related with the same entity may span multiple sentences~\cite{huang2020knowledge}.
% Since an internal KG is constructed solely based on the input text, many unseen entities and relations can be incorporated into the KG. In this way, the problem that the entity in the input text cannot be mapped to the corresponding entity in an existing external KG (e.g., ConceptNet~\cite{speer2017conceptnet}) is avoided. 
One of the mainstream methods of constructing an internal KG is using open information extraction (OpenIE).
Unlike traditional information extraction (IE) methods, OpenIE is not limited to a small set of target entities and relations known
in advance, but rather extracts all types of entities and relations found in input text~\cite{niklaus2018survey}. In this way, OpenIE facilitates the domain independent discovery of relations extracted from text and scales to large heterogeneous corpora.
% Commonly used OpenIE systems in NLG include Stanford OpenIE~\cite{angeli2015leveraging} and AW-OIE~\cite{stanovsky2018supervised}.

After obtaining an internal KG, the next step is to learn the representation of the internal KG and integrate it into the generation model.  For example, Zhu et al. use a graph attention network (GAT) to obtain the representation of each node, and fuse that into a transformer-based encoder-decoder architecture via attention~\cite{zhu2020boosting}. Their method generates abstractive summaries with higher factual correctness. Huang et al. extend by first encoding each paragraph as a sub-KG using GAT, and then connecting all sub-KGs with a Bi-LSTM~\cite{huang2020knowledge}. This process models topic transitions and recurrences, which enables the identification of notable content, thus benefiting summarization. 

\vspace{-0.05in}
\section{NLG enhanced by External Knowledge}

\subsection{NLG Enhanced by Knowledge Base}
One of the biggest challenges in NLG is to discover the dependencies of elements within a sequence and/or across input and output sequences. The dependencies are actually various types of \emph{knowledge} such as commonsense, factual events, and semantic relationship. Knowledge base (KB) is a popular technology that collects, stores, and manages large-scale information for knowledge-based systems like search engines. It has a great number of triples composed of subjects, predicates, and objects. People also call them ``facts'' or ``factual triplets''. Recently, researchers have been designing methods to use KB as external knowledge for learning the dependencies easier, faster, and better. 

Next, we introduce popular NLG applications enhanced by knowledge base:
\begin{itemize}
    \item \textbf{Question answering.} It is often difficult to generate proper answers only based on a given question. This is because, depending on what the question is looking for, a good answer may have different forms. It may completes the question precisely with the missing information. It may elaborate details of some part of the question. It may need reasoning and inference based on some facts and/or commonsense. So, only incorporating input question into neural generation models often fails the task due to the lack of commonsense/factual knowledge \cite{bi2019incorporating}. Related structured information of commonsense and facts can be retrieved from KBs.
    \item \textbf{Dialogue system.} The needs of KB in generating conversations or dialogues are relevant with QA but differ from two aspects. First, a conversation or dialogue could be open discussions when started by an open topic like ``\emph{Do you have any recommendations?}'' Second, responding an utterance in a certain step needs to recall previous contexts to determine involved entities. KB will play an important role to recognize dependencies in the long-range contexts. 
    % \item \textbf{Content manipulation.} The task of content manipulation is to describe KB data in a more desired way rather than just putting subject, predicate, and object together as a rigid sentence~\cite{feng2020learning}. 
    % The gold output of content manipulation is the text consisting of the content of KB data in the same style as the reference sentence. For example, a sports journalist may use particular  expressions and language styles to describe specific kinds of games~\cite{feng2020learning}.
\end{itemize}

\begin{figure}[t]
  \begin{center}
    \includegraphics[width=1.0\textwidth]{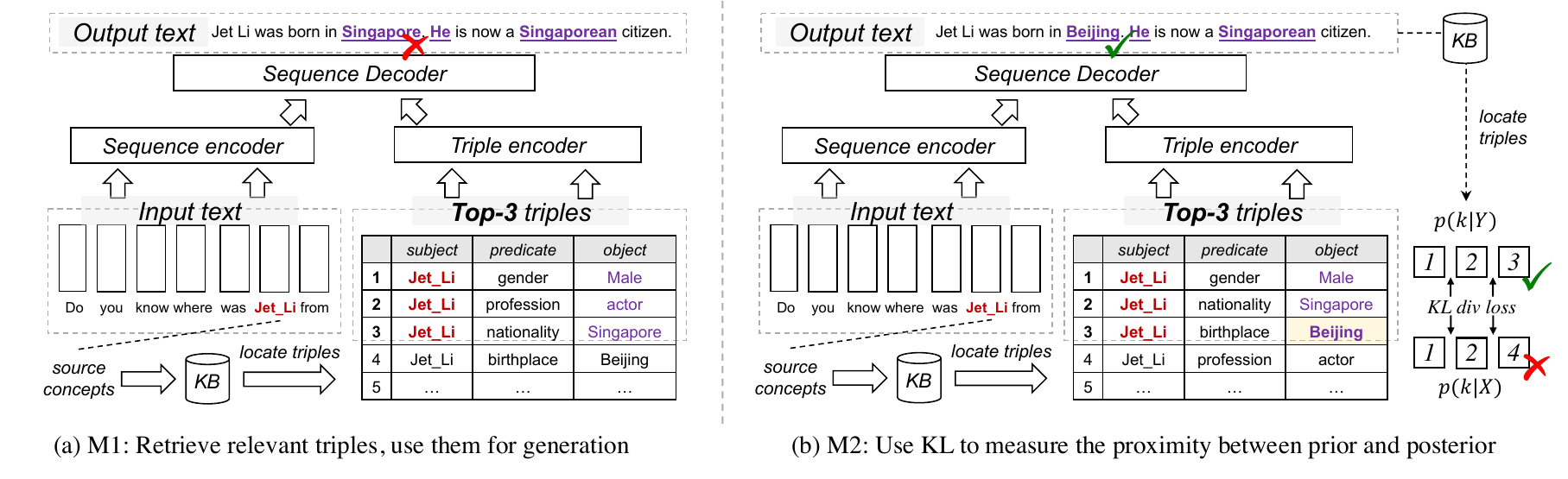}
  \end{center}
  \vspace{-0.1in}
  \caption{The left figure demonstrates retrieving relevant triples, then using them for generation; the right figure demonstrate using KL to measure the proximity between prior and posterior distribution.}
  \label{fig:overall_kb_e}
\vspace{-0.15in}
\end{figure}

To handle different kinds of relationships between KB and input/output sequences, these methods can be categorized into two methodologies which is shown in Figure \ref{fig:overall_kb_e}: (M1) design supervised tasks around KB for joint optimization; (M2) enhance incorporation by selecting KB or facts.

\vspace{-0.05in}
\subsubsection{\textbf{M1: Design Supervised Tasks around KB for Joint Optimization}}

Knowledge bases (KBs) that acquire, store, and represent factual knowledge can be used to enhance
text generation. However, designing effective incorporation to achieve a desired enhancement is challenging because a vanilla Seq2Seq often fails to represent discrete isolated concepts though they perform well to learn smooth shared patterns (e.g., language diversity).  
% In other words, when a model learns to perform well in finding information in KB, this extra training can help it generate the output sequence. 
To fully utilize the knowledge bases, {the idea is} to jointly train neural models on multiple tasks. For example, the target task is answer sequence generation, and additional tasks include question understanding and fact retrieval in the KB. Knowledge can be shared across a unified encoder-decoder framework design. Typically, question understanding and fact retrieval are relevant and useful tasks, because a question could be parsed to match (e.g., string matching, entity linking,
named entity recognition) its subject and predicate with the components of a fact triple in KB, and the answer is the object of the triple. 
% For example, if the question is \emph{``Where was Barack Obama born in the U.S.?''}, the phrase \emph{``Barack Obama''} can be matched to a fact triple, (\emph{Barack Obama}, \emph{born}, \emph{Hawaii}), in the KB. Parsing the question and retrieving relevant facts can exclude unrelated information and prevent unrelated knowledge from hindering answer generation.
KBCopy was the first work to generate responses using factual knowledge bases~\cite{eric2017copy}. During the generation, KBCopy is able to copy words from the KBs. However, the directly copying relevant words from KBs is extremely challenging.
% GenQA was the first work to generate answers using factual knowledge bases~\cite{yin2016neural}. During the generation, GenQA is able to retrieve words from the KBs. 
% However, it could not adopt relevant words from the question (i.e., input sequence). It could not handle complex questions that were associated with multiple facts, either. He et al. recognized these two issues and proposed CoreQA. 
CoreQA used both copying and retrieving mechanisms to generate answer sequences with an end-to-end fashion~\cite{he2017generating}. Specifically, it had a retrieval module to understand the question and find related facts from the KB. 
Then, the question and all retrieved facts are transformed into latent representations by two separate encoders. During the decoding phase, the integrated representations are fed into the decoder by performing a joint attention on both input sequence and retrieved facts.
Figure \ref{fig:overall_kb_e}(a) demonstrates a general pipeline that first retrieves relevant triples from KBs, then leverages the top-ranked triples into the generation process.

\vspace{-0.05in}
\subsubsection{\textbf{M2: Enhance Incorporation by Selecting KB or Facts in KB}}

Ideally, the relevance of the facts is satisfactory with the input and output sequence dependencies, however, it is not always true in real cases. Lian et al. addressed the issue of selecting relevant facts from KBs based on retrieval models (e.g. semantic similarity) might not effectively achieve appropriate knowledge selection~\cite{lian2019learning}. The reason is that different kinds of selected knowledge facts can be used to generate diverse responses for the same input utterance. Given a specific utterance and response pair, the posterior distribution over knowledge base from both the utterance and the response may provide extra guidance on knowledge selection. The challenge lies in the discrepancy between the prior and posterior distributions. Specifically, the model learns to select effective knowledge only based on the prior distribution, so it is hard to obtain the correct posterior distribution during inference.

To tackle this issue, the work of Lian et al.~\cite{lian2019learning} and Wu et al.~\cite{wu2020topicka} (shown in Figure \ref{fig:overall_kb_e}(b)) approximated the posterior distribution using the prior distribution in order to select appropriate knowledge even without posterior information. They introduced an auxiliary loss, called Kullback-Leibler divergence loss ({KLDivLoss}), to measure the proximity between the prior distribution and the posterior distribution. The {KLDivLoss} is defined as follows:
\begin{equation}
    \setlength\abovedisplayskip{2pt} 
    \setlength\belowdisplayskip{2pt}
    \mathcal{L}_{\emph{KLDiv}}(\theta) = \sum^{N}_{i=1} p(k = k_i|X, Y) \log\frac{p(k = k_i|X, Y)}{p(k = k_i|X)},
\end{equation}
where $N$ is the number of retrieved facts. When minimizing {KLDivLoss}, the posterior distribution $p(k|X, Y)$ can be regarded as labels to apply the prior distribution $p(k|X)$ for approximating $p(k|X,Y)$. Finally, the total loss is written as the sum of the {KLDivLoss} and {NLL} (generation) loss.

\vspace{-0.05in}
\subsubsection{\textbf{Discussion and Analysis of Different Methods}} 
The relevance between triples in KBs and input sequences plays a central role in discovering knowledge for sequence generation.
Methods in \textbf{M1} typically follows the process that parses input sequence, retrieves relevant facts, and subsequently, a knowledge-aware output can be generated based on the input sequence and previously retrieved facts.
Even though the improvement by modeling KB with memory network~\cite{madotto2018mem2seq}, existing KG-enhanced methods still suffer from effectively selecting precise triples.

Methods of \textbf{M2} improve the selection of facts, in which the ground-truth responses used as the posterior context knowledge to supervise the training of the prior fact probability distribution.
Wu et al. used exact match and recall to measure whether the retrieved triples is used to generate the target outputs~\cite{wu2020diverse}. Table \ref{tab:kb-quant} shows the entity recall scores of M1-based methods and M2-based methods reported in \cite{wu2020diverse,wu2020topicka}. We observe that compared to M1-based methods, M2-based methods can greatly improve the accuracy of triple retrieval, as well as the generation quality.

There are still remaining challenges in KB-enhanced methods.
One is that retrieved facts may contain noisy information, making the generation unstable~\cite{kim2020sequential}.
This problem is extremely harmful in NLG tasks, e.g., KB-based question answering and task-oriented dialogue system, since the information in KB is usually the expected entities in the response.

\begin{table*}[t]
\caption{M2-based methods can retrieve more precise triples, and further improve the generation performance.}
\vspace{-0.15in}
\centering
\setlength{\tabcolsep}{2.5mm}\scalebox{0.85}{\begin{tabular}{|l|l|c|c|c|c|c||c|c|c|c|}
\hline
\multirow{3}*{Method} & \multirow{3}*{Cat.} & \multirow{3}*{Ref.} & \multicolumn{4}{c||}{Chinese Weibo (large)~\cite{wu2020topicka}} & \multicolumn{4}{c|}{Chinese Weibo (small)~\cite{wu2020diverse}} \\ 
\cline{4-11}
& & & \multicolumn{2}{c|}{Entity score} & \multicolumn{2}{c||}{Generation score} & \multicolumn{2}{c|}{Entity score} & \multicolumn{2}{c|}{Generation score} \\ 
\cline{4-11}
& & & Match & Recall & BLEU-2 & Dist-2 & Match & Recall & BLEU-2 & Dist-2 \\
\hline \hline
GenDS & M1 & \cite{zhu2017flexible} & 0.97 & 0.37 & 3.42 & 4.27 & 0.75 & 0.26 & 2.09 & 1.66 \\
CCM & M1 & \cite{zhou2018commonsense} & 1.09 & 0.37 & 4.75 & 4.87 & 0.99 & 0.28 & 3.26 & 2.59 \\
\hline
ConKADI & M2 & \cite{wu2020diverse} & - & - & - & - & \textbf{1.48} & \textbf{0.38} & \textbf{5.06} & \textbf{23.93} \\
TaFact & M2 & \cite{wu2020topicka} & \textbf{1.81} & \textbf{0.47} & \textbf{5.07} & \textbf{23.56} & - & - & - & - \\
\hline
\end{tabular}}
\vspace{-0.05in}
\label{tab:kb-quant}
\end{table*}

\subsection{NLG Enhanced by Knowledge Graph}
\label{sec:know-graph}

Knowledge graph (KG), as a type of structured human knowledge, has attracted great attention from both academia and industry. A KG is a structured representation of facts (a.k.a. knowledge triplets) consisting of entities\footnote{For brevity, we use “entities” to denote both entities (e.g., prince) and concepts (e.g., musician) throughout the paper.}, relations, and semantic descriptions~\cite{ji2020survey}. The terms of “knowledge base” and “knowledge graph” can be interchangeably used, but they do not have to be synonymous. The knowledge graph is organized as a graph, so the connections between entities are first-class citizens in it. {In the KG, people can easily traverse links to discover how entities are interconnected to express certain knowledge.} Recent advances in artificial intelligence research have demonstrated the effectiveness of using KGs in various applications like recommendation systems~\cite{wang2019knowledge}.

Next, we introduce popular NLG applications that have been enhanced by knowledge graph:

\begin{itemize}
    \item \textbf{Commonsense reasoning.} It aims to empower machines to capture the human commonsense from KG during generation.
    The methods exploit both structural and semantic information of the commonsense KG and perform reasoning over multi-hop relational paths, in order to augment the limited information with chains of evidence for commonsense reasoning. Popular tasks in commonsense reasoning generation include abductive reasoning (e.g., the $\alpha$NLG task)~\cite{bhagavatula2020abductive,ji2020language}, counterfactual reasoning~\cite{ji2020language,ji2020generating}, and entity description generation~\cite{cheng2020entdesc}. 
    \item \noindent\textbf{Dialogue system.} It frequently makes use of KG for the semantics in linked entities and relations~\cite{zhou2018commonsense,niu2019knowledge,tuan2019dykgchat,zhang2020grounded}. A dialogue may shift focus from one entity to another, breaking one discourse into several segments, which can be represented as a linked path connecting the entities and their relations.
    \item \noindent\textbf{Creative writing.} This task can be found in both scientific and story-telling domains. Scientific writing aims to explain natural processes and phenomena step by step, so each step can be reflected as a link on KG and the whole explanation is a path~\cite{koncel2019text,wang2019paperrobot}. In story generation, the implicit knowledge in KG can facilitate the understanding of storyline and better predict what will happen in the next plot~\cite{guan2019story,guan2020knowledge,liu2021kg}.
\end{itemize}

Compared with separate, independent knowledge triplets, knowledge graph provides comprehensive and rich entity features and relations for models to overcome the influence of the data distribution and enhance its robustness.
Therefore, node embedding and relational path have played important roles in various text generation tasks. The corresponding techniques are knowledge graph embedding (KGE)~\cite{wang2017knowledge} and path-based knowledge graph reasoning~\cite{chen2020review}. Furthermore, it has been possible to encode multi-hop and high-order relations in KGs using the emerging graph neural network (GNN)~\cite{wu2020comprehensive} and graph-to-sequence (Graph2Seq) frameworks~\cite{beck2018graph}.
% Formally, a KG can be defined as below.

\vspace{-0.05in}
\begin{definition}[Knowledge graph (KG)]
A knowledge graph (KG) is a directed and multi-relational graph composed of entities and relations which are regarded as nodes and different types of edges. Formally, a KG is defined as $\mathcal{G} = (\mathcal{U}, \mathcal{E}, \mathcal{R})$, where $\mathcal{U}$ is the set of entity nodes and $\mathcal{E} \subseteq \mathcal{U} \times \mathcal{R} \times \mathcal{U}$ is the set of typed edges between nodes in $\mathcal{U}$ with a certain relation in the relation schema $\mathcal{R}$.
\label{def:kg}
\end{definition}

Then given the input/output sequences in the text generation task, a subgraph of the KG which is associated with the sequences can be defined as below.
% Since there is no widely-accepted formal definition of KG-related concepts~\cite{ji2020survey}, we give the definition of (1) KG, (2) sequence-associated K-hop subgraph.
% and (3) k-hop path that will be used in the following sections.

\vspace{-0.05in}
\begin{definition}[Sequence-associated K-hop subgraph]
A sequence-associated K-hop subgraph is defined as $\mathcal{G}_{sub} = (\mathcal{U}_{sub}, \mathcal{E}_{sub}, \mathcal{R})$, where $\mathcal{U}_{sub}$ is the union of the set of entity nodes mapped through an \emph{entity linking} function $\psi: \mathcal{U} \times \mathcal{X} \rightarrow \mathcal{U}_{sub}$ \textit{and} their neighbors within \textit{K}-hops.
% from a word $v$ in an input sequence $X$ to their linked entities in $\mathcal{U}$.
Similarly, $\mathcal{E}_{sub} \subseteq \mathcal{U}_{sub} \times \mathcal{R} \times \mathcal{U}_{sub}$ is the set of typed edges between nodes in $\mathcal{U}_{sub}$. 
\label{def:sks}
\end{definition}
Sequence-associated subgraph provides a graphical form of the task data (i.e., sequences) and thus enables the integration of KGs and the sequences into graph algorithms.

Many methods have been proposed to learn the relationship between KG semantics and input/output sequences. They can be categorized into four methodologies as shown in Figure \ref{fig:overall_kg}: (M1) incorporate knowledge graph embeddings into language generation; (M2) transfer knowledge into language model with triplet information; (M3) perform reasoning over knowledge graph via path finding strategies; and (M4) improve the graph embeddings with graph neural networks.

\vspace{-0.03in}
\subsubsection{\textbf{M1: Incorporate Knowledge Graph Embeddings into Language Generation}}

Knowledge graph embedding (KGE) techniques learn node embedding from a KG~\cite{wang2017knowledge}.
% Embedding based methods means to pre-process a KG with knowledge graph embedding (KGE) technique~\cite{wang2017knowledge}.
KGE aims to capture the semantic relatedness between entity nodes from their connectivity information (i.e., different types of relations) in the KG. The primary idea is to represent entities and relations in a low-dimensional vector space $\mathbb{R}^d$, where $d \ll |\mathcal{U} \cup \mathcal{R}|$, to reduce data dimensionality while preserving the inherent structure of the KG.
TransE~\cite{bordes2013translating} 
% and its extensions such as TransH~\cite{wang2014knowledge} and TransR~\cite{lin2015learning} are 
is the most widely used KGE technique. In TransE, given a KG edge $(u_i, r, u_j)$, the relation is seen as a translation vector $\mathbf{r}$ so that the embedded entities $\mathbf{u}_i$ and $\mathbf{u}_j$ can be connected with low translation error, namely $\mathbf{u}_i + \mathbf{r} \approx \mathbf{u}_j$. For example, we have $\overrightarrow{Tokyo} + \overrightarrow{IsCapticalOf} \approx \overrightarrow{Japan} $ for the knowledge edge $(\textit{Tokyo},~ \textit{IsCapticalOf}, ~\textit{Japan})$. 
% TransE is known for its simplicity and effectiveness~\cite{moussallem2019thoth,wang2019paperrobot}. 
As shown in Figure \ref{fig:overall_kg}(a), a common strategy of incorporating KGE into NLG is to concatenate the original word representations ($\textbf{x}$) with the corresponding entity representations ($\textbf{u}$) from KGE~\cite{zhou2018commonsense,zhang2020grounded}.

\vspace{-0.03in}
\subsubsection{\textbf{M2: Transfer Knowledge into Language Model with Knowledge Triplet Information}}

\begin{figure}[t]
  \begin{center}
    \includegraphics[width=1.0\textwidth]{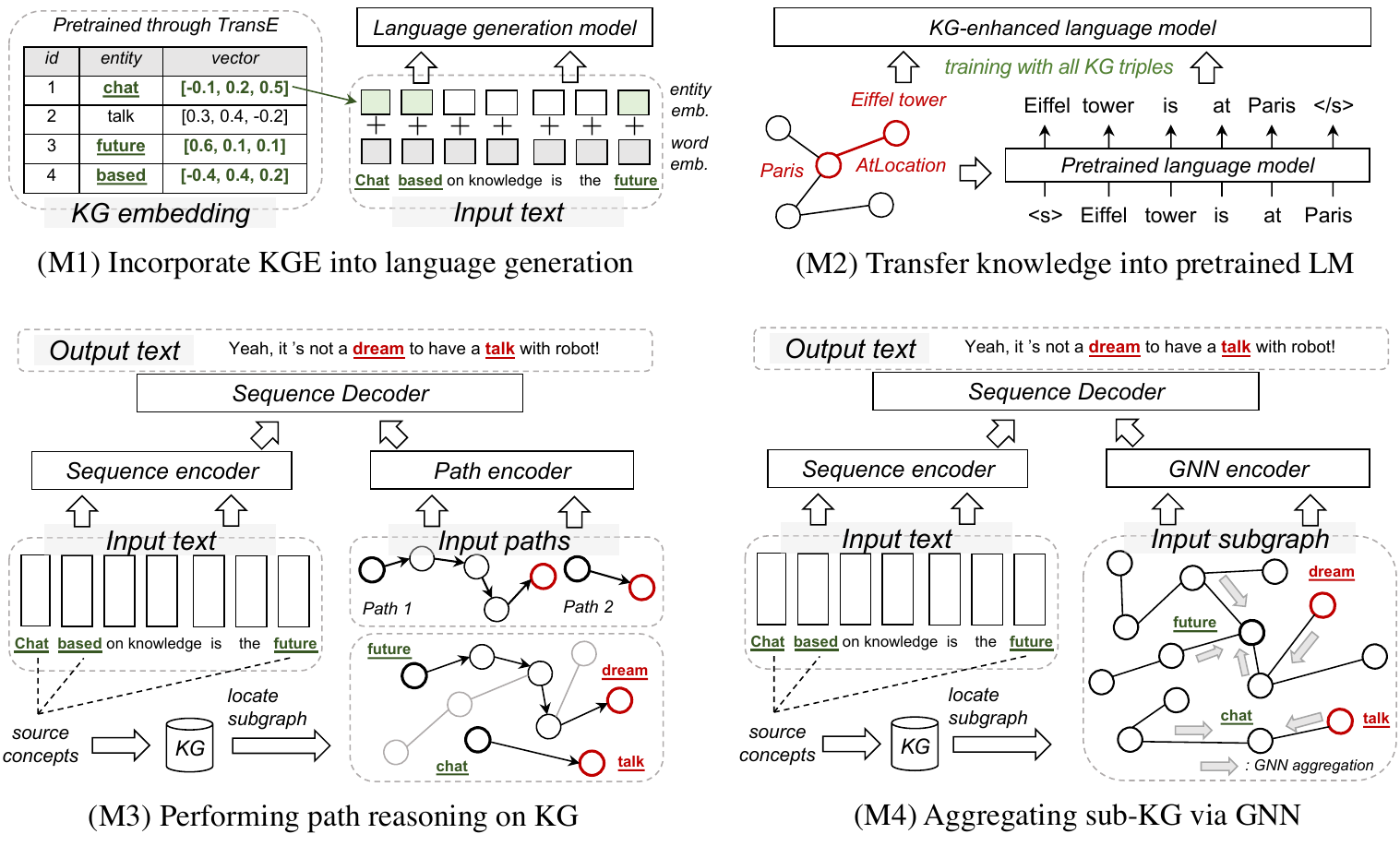}
  \end{center}
  \vspace{-0.15in}
  \caption{Four typical methodologies for incorporating KG semantics into text generation.}
  \label{fig:overall_kg}
\vspace{-0.15in}
\end{figure}

The vector spaces of entity embeddings (from KGE) and word embeddings (from pre-trained language models) are usually inconsistent~\cite{liu2021kg}.
% The embedding vectors of entity nodes (from KGE) and vectors of words in text (from pre-trained language models) are typically obtained in separate ways, making their vector-space inconsistent~\cite{liu2021kg}.
Beyond a simple concatenation,
% Instead of concatenating two channels of word representation and corresponding entity representation,
recent methods have explored to fine-tune the language models directly on knowledge graph triplets.
Guan et al. transformed the commonsense triplets (in ConceptNet and ATOMIC) into readable sentences using templates, as illustrated in Figure \ref{fig:overall_kg}(b). And then the language model (e.g., GPT-2) is fine-tuned on the transformed sentences to learn the commonsense knowledge to improve text generation.

\vspace{-0.03in}
\subsubsection{\textbf{M3: Perform Reasoning over Knowledge Graph via Path Finding Strategies}}

KGE learns node representations from one-hop relations through a certain semantic relatedness (e.g. TransE). However, Xiong et al. argued that an intelligent machine is supposed to be able to conduct explicit reasoning over relational paths to make multiple inter-related decisions rather than merely embedding entities in the KGs~\cite{xiong2017deeppath}.
Take the QA task an example. The machine performs reasoning over KGs to handle complex queries that do not have an obvious answer, infer potential answer-related entities, and generate the corresponding answer. So, the challenge lies in identifying a subset of desired entities and mentioning them properly in a response~\cite{moon2019opendialkg}. Because the connected entities usually follow natural conceptual threads, they help generate reasonable and logical answers to keep conversations engaging and meaningful. As shown in Figure \ref{fig:overall_kg}(c), path-based methods explore various patterns of connections among entity nodes such as meta-paths and meta-graphs. Then they learn from walkable paths on KGs to provide auxiliary guidance for the generation process. The path finding based methods can be mainly divided into two categories: (1) path ranking based methods and (2) reinforcement learning (RL) based path finding methods. 

\vspace{-0.03in}
\paragraph{\textbf{M3.1: Path routing and ranking}}
Path ranking algorithm (PRA) emerges as a promising method for learning and inferring paths on large KGs~\cite{lao2011random}. PRA uses random walks to perform multiple bounded depth-first search processes to find relational paths. Coupled with elastic-net based learning~\cite{zou2005regularization}, PRA picks plausible paths and prunes non-ideal, albeit factually correct KG paths. For example, Tuan et al. proposed a neural conversation model with PRA on dynamic knowledge graphs~\cite{tuan2019dykgchat}. In the decoding phase, it selected an output from two networks, a general GRU decoder network and a PRA based multi-hop reasoning network, at each time step. 
% However, one weak point of PRA is that it operates in a fully discrete space and thus the complexity is high of finding related entities and relations over a KG. 
Bauer et al. ranked and filtered paths to ensure both the information quality and variety via a 3-step scoring strategy: initial node scoring, cumulative node scoring, and path selection~\cite{bauer2018commonsense}. Ji et al. heuristically pruned the noisy edges between entity nodes and proposed a path routing algorithm to propagate the edge probability along
multi-hop paths to the entity nodes~\cite{ji2020generating}.

\vspace{-0.03in}
\paragraph{\textbf{M3.2: Reinforcement learning based path finding}}
% Different from PRA, reinforcement learning based methods perform reasoning in a continuous space by incorporating various criteria in reward functions, making the path finding process flexible.
Reinforcement learning (RL) based methods make an agent to perform reasoning to find a path in a continuous space. These methods incorporate various criteria in their reward functions of path finding, making the path finding process flexible.
Xiong et al. proposed DeepPath, the first work that employed Markov decision process (MDP) and used RL based approaches to find paths in KGs~\cite{xiong2017deeppath}. 
% However, the state of MDP requires the target entity to be known in advance, so the path finding strategy depends on the answer entity. Thus, it is often not applicable in many real-world QA and dialogue scenarios. 
% Recent RL based knowledge graph reasoning methods have demonstrated appealing performance on non-generative QA tasks like answer retrieval~\cite{das2018go}. 
% and reading comprehension~\cite{qiu2019machine}. 
Leveraging RL based path finding for NLG tasks typically consists of two stages~\cite{niu2019knowledge,liu2019knowledge}. First, they take a sequence as input, retrieve a starting node $u_0$ on $\mathcal{G}$, then perform multi-hop graph reasoning, and finally arrive at a target node $u_k$ that incorporates the knowledge for output sequence generation. Second, they represent the sequence $X$ and selected path $\Phi_k(u_0,u_k)$ through two separate encoders. They decode a sequence with multi-source attentions on the input sequence and selected path. 
Path-based knowledge graph reasoning converts the graph structure of a KG into a linear path structure that can be easily represented by sequence encoders (e.g, RNN)~\cite{niu2019knowledge,tuan2019dykgchat,fan2019using}. For example, Niu et al. encoded selected path and input sequence with two separate RNNs and generated sequence with a general attention-based RNN decoder~\cite{niu2019knowledge}.
% Besides, Niu et al. addressed that previous methods could not effectively exploit the textual information of entity nodes, when an agent was looking for the next proper node~\cite{niu2019knowledge}. To fully leverage the textual information, the agent scored each possible node from both global and local views.
% In the local view, the model used a bilinear model to calculate the similarity score based on non-contextual representation of a node. In the global view, the model adopted a machine reading comprehension to calculate the similarity score based on the textual information of the node.
To enhance the RL process, Xu et al. proposed six reward functions for training an agent in the reinforcement learning process.
For example, the functions looked for accurate arrival at the target node as well as the shortest path between the start and target node, i.e., minimize the length of the selected path $\Phi_k(u_0,u_k)$~\cite{xu2020conv}.
% For example, they not only consider whether the agent arrives at the target node, but also measures whether an agent chooses the shortest path between the start node to the target node, i.e., to minimize the selected path $\Phi_k(u_0,u_k)$
% an adversarial meta-learning algorithm to facilitate dialogue generation with dynamic KGs, when the temporal relations evolved as a single time scale process~\cite{xu2020conversational}.
% \textcolor{red}{what's the connection of this paragraph with RL? \textbf{Add two important phrases, the first one is to enhance the node transition by using two scoring function, the second one is 6 new reward functions.}}

\vspace{-0.03in}
\subsubsection{\textbf{M4: Improve the Graph Embeddings with Graph Neural Networks}}

% Instead of treating knowledge triples (entities) separately and independently, leveraging knowledge graph into Seq2Seq generation could better represent the semantics of a graph via linked entities and relations. 
% Path-based methods have shown great abilities for reasoning knowledge over a certain KG to bridge knowledge gap between question and answer, or capture conceptual flow in conversation/dialogue systems. However, many NLG tasks might not need to follow a reasoning process but to better understand global context under a particular generation process~\cite{guan2019story,koncel2019text}.
The contexts surrounding relevant entities on KGs play an important role in understanding the entities and generating proper text about their interactions~\cite{guan2019story,koncel2019text}.
For example, in scientific writing, it is important to consider the neighboring nodes of relevant concepts on a taxonomy and/or the global context of a scientific knowledge graph~\cite{koncel2019text}.
% scientific writing requires structured representation to facilitate the connection of relevant entities, and the preservation of global context (e.g. entity interactions)~\cite{koncel2019text}. 
% Complex events related with the same entity may span multiple sentences, making it challenging for existing sequential models to capture.
However, neither KGE nor relational path could fully represent such information. Graph-based representations aim at aggregating the context/neighboring information on graph data;
% A graph representation produces a structured summary and highlights the proximity of relevant concepts.
and recent advances of GNN models demonstrate a promising advancement in graph-based representation learning~\cite{wu2020comprehensive}. In order to improve text generation, graph-to-sequence (Graph2Seq) models encode the structural information of the KG in a neural encoder-decoder architecture~\cite{beck2018graph}. Since then, GNNs have been playing an important role in improving the NLG models.
They have been applied to both \emph{encoding} and \emph{decoding} phases.

\vspace{-0.03in}
\paragraph{\textbf{Learning KG-aware input text representation with GNNs (Encoding).}}
For encoding phase, a general process of leveraging GNNs for incorporating KG is to augment semantics of a word in the input text by combining with the vector of the corresponding entity node vector to the word on the KG~\cite{zhou2018commonsense,guan2019story,zhang2020grounded,huang2020knowledge,zeng2021enhancing}. A pre-defined entity linking function $\psi: \mathcal{U} \times \mathcal{X} \rightarrow \mathcal{U}_{sub}$ maps words in the input sequence to entity nodes on the KG. Given an input sequence, all the linked entities and their neighbors within $K$-hops compose a \textit{sequence-associated K-hop subgraph} $\mathcal{G}_{sub}$ (formally defined in Definition \ref{def:sks}).
% subset of nodes $\mathcal{U}_{sub} \subseteq \mathcal{U}$. 
For each entity node in $\mathcal{G}_{sub}$,
% $u =\psi(v, X) \in \mathcal{U}_{sub}$,
it uses the KG structure as well as entity and edge features (e.g., semantic description if available) to learn a representation vector $\textbf{u}$. 
Specifically, a GNN model follows a neighborhood aggregation approach that iteratively updates the representation of a node by aggregating information from its neighboring nodes and edges. After $k$ iterations of aggregation, the node representation captures the structural information within its $k$-hop neighborhood. Formally, the $k$-th layer of a node $u \in \mathcal{U}_{sub}$ is:
\begin{equation}
\textbf{u}^{(k)} = \textsc{Combine}_k (\textbf{u}^{(k-1)},  \textsc{Aggregate}_k(\big{\{} (\textbf{u}^{(k-1)}_{i}, \textbf{e}_{ij}^{(k-1)}, \textbf{u}^{(k-1)}_{j}): \forall (u_i, e_{ij}, u_j) \in \mathcal{N}(u)\big{\}})).
% \label{eq:gnns}
\end{equation}
The sub-graph representation $\textbf{h}_{subG}$ is learned thorough a $\textsc{Readout}(\cdot)$ function from all entity node representations (i.e., $
\textbf{h}_{subG} = \textsc{Readout}(\big{\{}\textbf{u}^{(k)}, u \in \mathcal{U}_{sub}\big{\}})$.
Zhou et al. was the first to design such a knowledge graph interpreter to enrich the context representations with neighbouring concepts on ConceptNet using graph attention network (GAT)~\cite{zhou2018commonsense}.

\vspace{-0.05in}
\paragraph{\textbf{Dynamically attending KG representation (Decoding).}}
The sequence decoder uses attention mechanism to find useful semantics from the representation of KG as well as the hidden state of the input text, where the KG's representation is usually generated by GNNs.
% The obtained KG representation via GNN can be further used to enhance the decoding phase via attention mechanism at each step.
Specially, the hidden state is augmented by subgraph representation $\textbf{h}_{subG}$, i.e., $\textbf{s}_0 = \textbf{h}_n \oplus \textbf{h}_{subG} $~\cite{beck2018graph}. Then, the decoder attentively reads the retrieved subgraph to obtain a graph-aware context vector. Then it uses the vector to update the decoding state~\cite{zhou2018commonsense,guan2019story,zhang2020grounded,ji2020language,liu2021kg}. It adaptively chooses a generic word or an entity from the retrieved subgraph to generate output words. Because graph-level attention alone might overlook fine-grained knowledge edge information, some recent methods adopted the hierarchical graph attention mechanism~\cite{zhou2018commonsense,guan2019story,liu2021kg}. It attentively read the retrieved subgraph $\mathcal{G}_{sub}$ and then attentively read all knowledge edges $\mathcal{E}_{sub}$ involved in $\mathcal{G}_{sub}$. Ji et al. added a relevance score that reflected the relevancy of the knowledge edge according to the decoding state~\cite{ji2020language}.

\begin{table*}[t]
\caption{Tasks, datasets and KG sources used in different KG-enhanced papers. We also compared the performance of different models before and after incorporating KG into the generation process, in which ``w/o KG'' performance comes from the best baseline method; ``with KG'' comes from the KG-enhanced method.}
\vspace{-0.15in}
\begin{center}
\scalebox{0.815}{\begin{tabular}{|l|l|c|c|l|r|r|r|c|r|}
\hline
\multirow{2}*{Tasks} & \multirow{2}*{Methods} & \multirow{2}*{Ref.} & \multirow{2}*{Cat.} & \multicolumn{2}{c|}{Dataset Information} & \multicolumn{3}{c|}{Effect of KG} & KG\\
\cline{5-9}
& & & & Name & \#Instance & w/o KG & with KG & $\Delta$BLEU & source \\
\hline \hline
\multirow{4}*{\makecell[l]{Common- \\sense \\ reasoning}} & KG-BART & \cite{liu2021kg} & M4 & CommonGen & 77,449 & 28.60 & 30.90 & +2.30 & ConceptNet \\ 
\cline{2-10}
& CE-PR & \cite{ji2020generating} & M3 & ComVE & 30,000 & 15.70 & 17.10 & +1.60 & ConceptNet \\ 
\cline{2-10}
& GRF & \cite{ji2020language} & M4 & $\alpha$NLG-ART & 60,709 & 9.62 & 11.62 & +2.00 & ConceptNet \\
\cline{2-10}
& MGCN & \cite{cheng2020entdesc} & M3 & EntDesc & 110,814 & 24.90 & 30.00 & +4.30 & Self-built KG \\ 
\hline \hline
\multirow{5}*{\makecell[l]{Story \\ generation}}  & IE+MSA & \cite{guan2019story} & M4 & ROCStories & \multirow{2}*{98,162} & 8.25 & 9.36 & +1.11 & ConceptNet \\ 
\cline{2-4}\cline{7-10}
& GRF & \cite{ji2020language} & M4 & (split-1) & & 10.40 & 11.00 & +0.60 & ConceptNet \\ 
\cline{2-10}
& \multirow{2}*{KEPM} & \multirow{2}*{\cite{guan2020knowledge}} & \multirow{2}*{M2} & ROCStories & \multirow{2}*{98,162} & \multirow{2}*{14.10} & \multirow{2}*{14.30} & \multirow{2}*{+0.20} & ConceptNet\\ 
& &  & & (split-2) & & & & & \& ATOMIC  \\ 
\cline{2-10}
& MRG & \cite{zhao2020graph} & M3 & VisualStory & 50,000 & 3.18 & 3.23 & +0.05 & ConceptNet \\ 
\hline \hline
\multirow{2}*{\makecell[l]{Scientific \\ writing}} & GraphWriter & \cite{koncel2019text} & M4 & AGENDA & 40,000 & 12.20 & 14.30 & +1.90 & Self-built KG \\ 
\cline{2-10}
& PaperRobot & \cite{wang2019paperrobot} & M4 & PaperWriting & 27,001 & 9.20 & 13.00 & +3.80 & Self-built KG \\
\hline \hline
\multirow{3}*{\makecell[l]{Dialogue \\ system}} & ConceptFlow &
\cite{zhang2020grounded} & M4 & Reddit-10M & 3,384K & 1.62 & 2.46 & +0.84 & ConceptNet \\ 
\cline{2-10}
& AKGCM & \cite{liu2019knowledge} & M3 & EMNLP dialog & 43,192 & 32.45 & 30.84 & \underline{-1.61} & Self-built KG \\ 
\cline{2-10}
& AKGCM & \cite{liu2019knowledge} & M3 & ICLR dialog & 21,569 & 6.74 & 6.94 & +0.20 & Self-built KG \\
\hline \hline
\multirow{2}*{\makecell[l]{Question \\ answering}} & \multirow{2}*{MHPGM} & \multirow{2}*{\cite{bauer2018commonsense}} & \multirow{2}*{M3} & \multirow{2}*{NarrativeQA} & \multirow{2}*{46,765} & \multirow{2}*{19.79} & \multirow{2}*{21.07} & \multirow{2}*{+1.28} & \multirow{2}*{Self-built KG} \\
&&&&&&&&& \\
 \hline
\end{tabular}}
\end{center}
\vspace{-0.15in}
\label{tab:data-kg}
\end{table*}

\begin{table*}[t]
\caption{Qualitative comparison between different KG-enhanced methods.}
\vspace{-0.15in}
\begin{center}
\scalebox{0.86}{\begin{tabular}{|l|c|c|c|c|c|l|l|l|}
\hline
{\multirow{2}*{Methods}} & \multirow{2}*{Ref.} & \multicolumn{4}{c|}{Method category} & Multi-hop info. & Multi-hop path & Auxiliary (knowledge \\
\cline{3-6}
& & M1 & M2 & M3 & M4 & aggregation & reasoning & related) task(s) \\
\hline \hline
THOTH & \cite{moussallem2019thoth} & $\checkmark$ & & & & $\times$ & $\times$ & $\times$ \\
\hline
CCM & \cite{zhou2018commonsense} & & & & $\checkmark$ & $\times$, one-hop & $\times$ & $\times$ \\
\hline
KEPM & \cite{guan2020knowledge} & & $\checkmark$ & & & $\times$ & $\times$ & $\times$ \\
\hline
AKGCM & \cite{liu2019knowledge} & & & $\checkmark$ & & $\times$ & $\checkmark$, Markov decision & $\checkmark$, Path selection \\
% \hline
% DyKgChat & \cite{tuan2019dykgchat} & & $\checkmark$ & & & $\times$ & $\checkmark$, Markov decision & $\times$ \\
\hline
IE+MSA & \cite{guan2019story} & & & & $\checkmark$ & $\checkmark$, by GNN & $\times$ & $\times$ \\
\hline
ConceptFlow & \cite{zhang2020grounded} & & & & $\checkmark$ & $\checkmark$, by GNN & $\times$ & $\times$ \\
\hline
CE-PR & \cite{ji2020generating} & & & $\checkmark$ & & $\times$ & $\checkmark$, Path routing & $\checkmark$, Concept selection \\
\hline
GRF & \cite{ji2020language} & & & & $\checkmark$ & $\checkmark$, by GNN & $\checkmark$, Path scoring & $\checkmark$, Link prediction \\
\hline
\end{tabular}}
\end{center}
\vspace{-0.15in}
\label{tab:qual-kg}
\end{table*}

\subsubsection{\textbf{Discussion and Analysis of the Methodologies and Methods}}
% \revise{First, we generally discuss the strong points and weak points of the aforementioned methods that use knowledge graphs to enhance text generation. Second, we discuss on what methodologies are more suitable for what applications. And lastly, we provide qualitative and quantitative analysis of the methods.}

\vspace{-0.05in}
\paragraph{\textbf{Pros and cons.}}
% Knowledge graph embedding (\textbf{M1}) was the earliest attempt to project knowledge graph data into continuous representations and use them to improve text generation. 
Knowledge graph embedding (\textbf{M1}) was the earliest attempt to embed components of a KG including entities and relations into continuous vector spaces and use them to improve text generation. 
Those entity and relation embeddings can simply be used to enrich input text representations (e.g., concatenating embeddings), bridging connections between entity words linked from input text in latent space. 
Because the graph projection and text generation are performed as two separate steps, the embedding vectors from knowledge graph and the hidden states from input text were in two different vector spaces. The model would have to learn to bridge the gap, which might make a negative impact on the performance of text generation.

% For M1, knowledge graph embedding (KGE) is able to embed components of a KG including entities and relations into continuous vector spaces, so as to simplify the manipulation while preserving the inherent structure of the KG. 
% Those entity and relation embeddings can simply be used to enrich input text representations (e.g., concatenating embeddings), bridging connections between entity words linked from input text in latent space. 
% However, the embedding vectors of entity nodes (from KGE) and vectors of words in text (from pre-trained language models) are typically obtained in separate ways, making their vector-space inconsistent.

Fine tuning pre-trained language models on the KG triplets (\textbf{M2}) can eliminate the gap between the two vector spaces.
% For M2, fine tuning the pre-trained language models on KG triples solves the problem of inconsistency in vector-space between entity nodes and words in text. 
% In this way, there is no need to bridge the representation gap between entity embeddings and text-contextual embeddings.
Nevertheless, M1 and M2 share two drawbacks.
% have two similar drawbacks.
First, they only preserve information of direct (one-hop) relations in a KG, such as pair-wise proximity in M1 and KG triplet in M2, but ignore the indirect (multi-hop) relations of concepts. The indirect relations may provide plausible evidence of complex reasoning for some text generation tasks.
Second, from the time KGs were encoded in M1 or M2 methods, the generation models would no longer be able to access the KGs but their continuous representations. Then the models could not support reasoning like commonsense KG reasoning for downstream tasks.
Due to these two reasons, M1 and M2 were often used to create basic KG representations upon which the KG path reasoning (M3) and GNNs (M4) could further enrich the hidden states~\cite{zhou2018commonsense,zhang2020grounded}.

% First, they are only directly dependent on one-hop relation path and restricted with rigorous objective of either a certain scoring function or a language model. So, it ignores the abundant structural relational relevance of the concepts in the knowledge graph that may provide multiple plausible evidence for complex reasoning. 
% Second, since training KGE or using KG triples to fine-tune pre-trained language models are separate from fine-tuning the generation model on downstream tasks, it is hard to enable the model to utilize the encoded knowledge in downstream tasks which requires reasoning over underlying commonsense knowledge. 
% Therefore, M1 and M2 are often used as the first step to enrich the textual representation, and further enhanced by KG path reasoning (M3) and GNNs (M4)~\cite{zhou2018commonsense,zhang2020grounded}.
% The choice between performing reasoning over KG via path finding or aggregating multi-hop information via GNNs mainly depends on the purpose and application of using knowledge graphs. 

The path finding methods of KG reasoning (\textbf{M3}) perform multi-hop walks on the KGs beyond one-hop relations. It enables reasoning that is needed in many text generation scenarios such as commonsense reasoning and conversational question answering.
At the same time, it provides better interpretability for the entire generation process, because the path selected by the KG reasoning algorithm will be explicitly used for generation.
However, the selected paths might not be able to capture the full contexts of the reasoning process due to the limit of number.
Besides, reinforcement-learning based path finding uses heuristic rewards to drive the policy search, making the model sensitive to noises and adversarial examples.
% For M3, performing reasoning over KG via path finding provides flexible multi-hop walks on graphs, not restricted to one hop relations. More importantly, path-finding on KG is a process of knowledge reasoning, which can be well integrated with the scenarios that requires complex reasoning.
% Commonsense reasoning and conversational question answering system are the most important applications.
% However, path finding strategies cannot well understand the global context under a particular generation process since only a few paths are selected. 

The algorithms of GNN and Graph2Seq (\textbf{M4}) can effectively aggregate semantic and structural information from multi-hop neighborhoods on KGs, compared to M3 that considers multi-hop paths.
% For M4, as recent advances of GNNs and Graph2Seq potentiate to bridge up the gap between graph representation learning and language generation, GNNs serve as an important role of integrating rich semantic and structural knowledge into text generation.
% Compared to M3, it directly enriches the textual representation (i.e., word embedding) of input sequence with structured representation (i.e., node embedding) from knowledge graph.
Therefore, the wide range of relevant information can be directly embedded into the encoder/decoder hidden states. Meanwhile, M4 enables back propagation for jointly optimizing text encoder and graph encoder. 
Furthermore, the attention mechanism that has been applied in GNN and Graph2Seq (e.g., graph attention) can explain the model's output at some extent, though the multi-hop paths from M3 has better interpretability.

% though performing multi-hop aggregation through GNN cannot provide as powerful interpretation capabilities as M3, it can still focus on different multi-hop relational paths by attentively aggregating information from neighboring nodes and edges (e.g., using graph attention).

% \textcolor{red}{The two sentences from here are confusing. Compared to M1/M2/M3, what would be the strong "unique" points of M4? "directly"? "structured knowledge"? "related to entity words"? "global context"? "particular generation process"? "perform reasoning over multi-hop relational paths"? "dynamically aggregating"?} 

% \textcolor{red}{Where do this "However" come from? Are all four methodologies have this weak point? Is there a reference that defined "low entity coverage"? Has any paper discussed about this challenge? If no, that'd be weird. It's not clear what this paragraph is talking about and why it should be here (what's the connection with the above paragraphs).} 

M3 and M4 are able to use multi-hop relational information, compared to M1 and M2. However, they have two weak points. First, they have higher complexity than M1 and M2. In M3, the action space of path finding algorithms can be very large due to the large size and sparsity of the knowledge graph. In M4, the decoder has to attentively read both input sequence and knowledge graph. Second, the subgraphs retrieved by M3 and M4 might provide low coverage of useful concepts for generating the output.
% since reasoning over the entire knowledge graph is intractable, M3 and M4 have to retrieve the subgraph based on the concepts contained in the input sequence (defined as sequence-associated subgraph in Definition \ref{def:sks}). However, the retrieved subgraph might have low coverage of useful concepts for generation.
For example, people use ConceptNet, a widely used commonsense KG, to retrieve the subgraph on three generative commonsense reasoning tasks. The task datasets are ComVE~\cite{ji2020language}, $\alpha$-NLG~\cite{bhagavatula2020abductive}, and ROCSories~\cite{guan2019story}. We found 25.1\% / 24.2\% / 21.1\% of concepts in the output could be found on ConceptNet, but only 11.4\% / 8.1\% / 5.7\% of concepts in the output can be found on the retrieved 2-hop sequence-associated subgraph, respectively. It means that a large portion of relevant concepts on the KG are not utilized in the generation process.

% New relations among entities can be derived through knowledge reasoning and can feed back to enrich the knowledge graphs, and then support the advanced applications

% However, explicitly modeling KG may suffer from the low entity coverage problem. We made statistics on the coverage of entities in ConceptNet~\cite{speer2017conceptnet} on three commonsense reasoning datasets. First, only around 21.1\%-25.1\% of words in the input sequence and output sequence can be found in the ConceptNet. Second, we extract sequence-associated 2-hop subgraphs (as defined in definition \ref{def:sab}) of each input sequence. We pruned relations between an entity and stop words. There are around 200 entities and more than 1,000 triples in a 2-hop subgraphs, however, only around 0.6 entities and 3.1 triples can be found in output sequence. The low entity coverage problem causes extreme imbalance between positive and negative entities, making it difficult for the model to effectively leverage correct entities in the generation process.

\vspace{-0.05in}
\paragraph{\textbf{Quantitative analysis.}}
Table \ref{tab:data-kg} summarizes tasks, datasets, and KG sources used in existing KG-enhanced works. Three important things should be mentioned. First, all the datasets in the table are public, and we include their links in Table \ref{tab:code}. CommonGen~\cite{lin2020commongen}, ComVE~\cite{wang2020semeval} and $\alpha$-NLG~\cite{bhagavatula2020abductive} have a public leaderboard for competition. 
Second, for KG sources, we observe that eight (57.1\%) papers use ConceptNet as external resource, while six (42.9\%) papers constructed their own KGs from domain-specific corpus. For example, Koncel et al. created a scientific knowledge graph by applying the SciIE tool (science domain information extraction)~\cite{koncel2019text}. 
Besides, Zhao et al. compared the performance of models between using ConceptNet and using a self-built KG, and found the model with self-built KG could work better on story generation and review generation tasks~\cite{zhao2020graph}.
Third, we observed that KG-enhanced NLG methods made the largest improvement on generative commonsense reasoning tasks, in which the average improvement is +2.55\% in terms of $\Delta$BLEU, while the average improvement on all different tasks is +1.32\%.

\vspace{-0.05in}
\paragraph{\textbf{Qualitative analysis.}}
Table \ref{tab:qual-kg} compares different KG-enhanced methods from three dimensions: multi-hop information aggregation, multi-hop path reasoning, and auxiliary knowledge graph related tasks. M3 is commonly used for multi-hop path reasoning and M4 is used for multi-hop information aggregation, except that CCM~\cite{zhou2018commonsense} only aggregates one-hop neighbors. Besides, the auxiliary KG-related tasks are often used to further help the model learn knowledge from the KG. For example, ablation studies in \cite{liu2019knowledge,ji2020generating,ji2020language} show that the tasks of path selection, concept selection and link prediction can further boost the generation performance. GRF~\cite{ji2020language} learns these three abilities at the same time. It achieves the state-of-art performance on three generation tasks.
% as shown in Table \ref{tab:quant-kg}.

% In Table \ref{tab:quant-kg}, we make quantitative comparison between different KG-enhanced methods on four NLG tasks. All of the four tasks have (i) public resource; and (ii) been used by at least two papers. 
% \textcolor{red}{Observations and Conclusions from the tables must be clearly given in the text. Otherwise, readers might not be able to capture your idea or even not able to know where to start to read in the table, though they could find the tables.}

\begin{figure}[t]
  \begin{center}
    \includegraphics[width=1.0\textwidth]{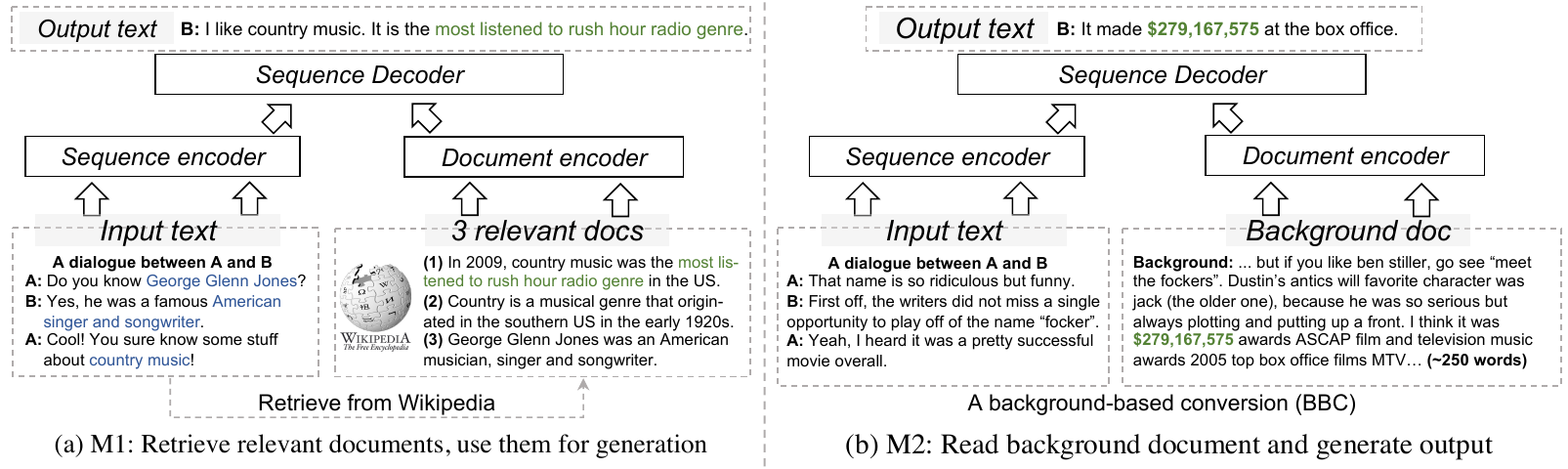}
  \end{center}
  \vspace{-0.15in}
  \caption{The left figure demonstrates retrieving relevant documents, then using them for generation; the right figure demonstrate reading background document to conduct conversions.}
  \label{fig:overall_kt}
\vspace{-0.1in}
\end{figure}

\subsection{NLG enhanced by Grounded Text}

Knowledge grounded text refers to textual information that can provide additional knowledge relevant to the input sequence. The textual information may not be found in training corpora or structured databases, but can be obtained from massive textual data from online resources. These online resources include encyclopedia (e.g., Wikipedia), social media (e.g., Twitter), shopping 
websites (e.g., Amazon reviews). Knowledge grounded text plays an important role in understanding the input sequence and its surrounding contexts. For example, Wikipedia articles may offer textual explanations or background information for the input text. Amazon reviews may contain necessary descriptions and reviews needed to answer a product-related question. Tweets may contain people's comments and summaries towards an event.
Therefore, knowledge grounded text is often taken as an important external knowledge source to help with a variety of NLG applications.

Next, we introduce popular NLG applications enhanced by knowledge grounded text:

\begin{itemize}
    \item \textbf{Dialogue system.} Building a fully data-driven dialogue system is difficult since most of the universal knowledge is not presented in the training corpora~\cite{ghazvininejad2018knowledge}. The lack of universal knowledge considerably limits the appeal of fully data-driven generation methods, as they are bounded to respond evasively or defectively and seldom include meaningfully factual contents. To infuse the response with factual information, an intelligent machine is expected to obtain necessary background information to produce appropriate response.
    \item \noindent\textbf{Summarization.} Seq2Seq models that purely depend on the input text tend to ``lose control'' sometimes. For example, 3\% of summaries contain less than three words, and 4\% of summaries repeat a word for more than 99 times as mentioned in~\cite{cao2018retrieve}. Furthermore, Seq2Seq models usually focus on copying source words in their exact order, which is often sub-optimal in abstractive summarization. Therefore, leveraging summaries of documents similar as the input document as templates can provide reference for the summarization process~\cite{cao2018retrieve,wang2019biset}.
    \item \noindent\textbf{Question answering (QA).} It is often difficult to generate proper answers only based on the given question. For example, without knowing any information of an Amazon product, it is hard to deliver satisfactory answer to the user questions such as \emph{``Does the laptop have a long battery life?''} or \emph{``Is this refrigerator frost-free?''} So, the product description and customer reviews can be used as a reference for answering product-related questions~\cite{chen2019driven,bi2020generating}.
    % \item \noindent\textbf{Argument generation.} It is helpful in a variety of decision-making situations such as business, law, and politics. Inspired by the observation that when humans construct arguments, they often collect references from external sources, e.g., Wikipedia or research papers, and then write their own arguments by synthesizing talking points from the references. Thus, retrieving external evidence is also an important step for automatic argument generation.
\end{itemize}

To handle different kinds of relationships between grounded text and input/output sequences, these methods can be categorized into two methodologies as shown in Figure \ref{fig:overall_kt}: (M1) guiding generation with retrieved information; (M2) modeling background knowledge into response generation.

\vspace{-0.05in}
\subsubsection{\textbf{M1: Guiding Generation with Retrieved Information}}
Because knowledge grounded text is not presented in the training corpora, an idea is to retrieve relevant textual information (e.g., a review, a relevant document, a summary template) from \emph{external sources} based on the input text and to incorporate the retrieved grounded text into the generation process. This process is similar to designing knowledge acquisition and incorporation of KBs and KGs in text generation tasks. The difference is that ground text is unstructured and noisy. So, researchers design knowledge selection and incorporation methods to address the challenges.
Based on the number of stages, we further divide related methods into two categories: retrieve-then-generate (also known as retrieval-augmented generation, short as RAG, in many existing papers~\cite{lewis2020retrieval,petroni2021kilt,krishna2021hurdles}) methods (2-stage methods) and retrieve, rerank and rewrite methods (3-stage methods).

\vspace{-0.05in}
\paragraph{\textbf{M1.1: Retrieval-augmented generation (RAG)}}
RAG follows a two-stage process: retrieval and generation. 
Specially, as shown in Figure \ref{fig:overall_kt}(a), a retriever $p(Z|X)$ first returns (usually top-K truncated) distributions over text passages given a query $X$, and then a generator $p(y_i|X, Z, y_{1:i-1})$ generates a current token based on a context of the previous tokens $y_{1:i-1}$, the original input $X$ and a retrieved passage $Z$.
Methods for retrieving fact or review snippets are various, including matching from a collection of raw text entries indexed by named entities~\cite{ghazvininejad2018knowledge}; scoring relevant documents within a large collection by statistical approaches such as BM25~\cite{dinan2019wizard}, or neural-based retrieval approaches such as dense paragraph retrieval (DPR)~\cite{lewis2020retrieval}.
For training the retriever and generator, most of existing work has jointly optimized these two components, without any direct supervision on what document should be retrieve~\cite{lewis2020retrieval,krishna2021hurdles}.
However, by asking human experts to label what document should be retrieved and adding the retrieval loss (resulting in a multi-task learning setting), the generation performance can be greatly improved~\cite{dinan2019wizard,kim2020sequential}, though the labelling process is an extremely time-consuming and labor-intensive task.

\begin{table*}[t]
\caption{Tasks, datasets and evidence sources used in retrieve-then-generate (M1) papers. We also include their document(d)/sentence(s) retrieval space and the number of retrieved document(d)/sentence(s).}
\vspace{-0.15in}
\begin{center}
\scalebox{0.83}{\begin{tabular}{|l|l|l|c|l|r|r|r|}
\hline
Evidence & \multirow{2}*{Tasks} & \multirow{2}*{Methods} & \multirow{2}*{Ref.} & \multicolumn{2}{c|}{Dataset Information} & Retrieval & \# Retri- \\
\cline{5-6}
sources & & & & Name & \#Instance & space (d/s) & eved d/s \\
\hline \hline
\multirow{7}*{\makecell[l]{Wikipedia}} & \multirow{2}*{\makecell[l]{Dialogue \\ system}} & MemNet & \cite{dinan2019wizard} & \multirow{2}*{\makecell[l]{Wizard of \\ Wikipedia (WoW)}} & \multirow{2}*{22,311} & \multirow{2}*{5.4M/93M} & 7 \\
 &  & SKT & \cite{kim2020sequential} & & & & 7 \\ 
\cline{2-8}
& \multirow{3}*{\makecell[l]{Question \\ answering}} & RAG & \cite{lewis2020retrieval} & MS-MARCO & 267,287 & 21M/- & 10 \\ 
\cline{3-8}
& & BART+DPR & \cite{petroni2021kilt} & \multirow{2}*{ELI5} & \multirow{2}*{274,741} & 3.2M/- & - \\
& & RT+C-REALM & \cite{krishna2021hurdles} & & & 3.2M/- & 7 \\
\cline{2-8}
& \multirow{2}*{\makecell[l]{Argument \\ generation}} & H\&W & \cite{hua2018neural} & \multirow{2}*{ChangeMyView} & \multirow{2}*{287,152} & 5M/- & 10 \\ 
& & CANDELA & \cite{hua2019argument} & & & 5M/- & 10 \\
\hline \hline
\multirow{2}*{\makecell[l]{Online platform \\ (e.g., Amazon)}} & \multirow{2}*{\makecell[l]{Dialogue (for \\ business)}} & AT2T & \cite{kim2020retrieval} & Amazon books & 937,032 & -/131K & 10 \\ 
& & KGNCM & \cite{ghazvininejad2018knowledge} & Foursquare & 1M & -/1.1M & 10 \\ 
\hline \hline
\multirow{2}*{\makecell[l]{Gigawords}} & \multirow{2}*{\makecell[l]{Summari- \\ zation}} & R$^3$Sum & \cite{cao2018retrieve} & \multirow{2}*{Gigawords} & \multirow{2}*{3.8M} & -/3.8M & 30 \\ 
& & BiSET & \cite{wang2019biset} & & & -/3.8M & 30 \\ 
\hline
\end{tabular}}
\end{center}
\vspace{-0.1in}
\label{tab:data-kt}
\end{table*}

\begin{table*}[t]
\caption{Qualitative comparison between different grounded text enhanced methods.}
\vspace{-0.15in}
\begin{center}
\scalebox{0.895}{\begin{tabular}{|l|c|c|c|c|l|l|l|}
\hline
{\multirow{2}*{Methods}} & \multirow{2}*{Ref.} & \multicolumn{3}{c|}{Method category} & \multirow{2}*{Retrieval supervision} & Retriever  & Number \\
\cline{3-5}
& & M1.1 & M1.2 & M2 & & pre-training & of stages \\
\hline \hline
MemNet & \cite{dinan2019wizard} & $\checkmark$ & & & $\checkmark$, Human annotated labels & $\times$ & 2 \\
% retrieve-generate \\
\hline
SKT & \cite{kim2020sequential} & $\checkmark$ & & & $\checkmark$, Human annotated labels & $\times$ & 2 \\
\hline
R$^3$Sum & \cite{cao2018retrieve} & & $\checkmark$ & & $\checkmark$, Pseudo labels & $\times$ & 3, with rerank \\
\hline
BiSET & \cite{wang2019biset} & & $\checkmark$ & & $\checkmark$, Pseudo labels & $\times$ & 3, with rerank \\
\hline
RefNet & \cite{meng2020refnet} & & & $\checkmark$ & $\times$ & $\times$ & 1, no retrieval \\
\hline
GLKS & \cite{ren2020thinking} & & & $\checkmark$ & $\times$ & $\times$ & 1, no retrieval \\
\hline
RAG & \cite{lewis2020retrieval} & $\checkmark$ & & & $\times$ & $\checkmark$, DPR & 2 \\
\hline
Kilt & \cite{petroni2021kilt} & $\checkmark$ & & & $\times$ & $\checkmark$, DPR & 2 \\
\hline
RT+C-REALM & \cite{krishna2021hurdles} & $\checkmark$ & & & $\times$ & $\checkmark$, REALM & 2 \\
\hline
\end{tabular}}
\end{center}
\vspace{-0.1in}
\label{tab:qual-kt}
\end{table*}

Ghazvininejad et al. proposed a knowledge grounded neural conversation model (KGNCM), which is the first work to retrieve review snippets from Foursquare and Twitter. Then it incorporates the snippets into dialogue response generation~\cite{ghazvininejad2018knowledge}. It uses an end-to-end memory network~\cite{sukhbaatar2015end} to generate responses based on the selected review snippets.
% In QA, Chen et al. used online customer reviews to answer product-related questions~\cite{chen2019driven} . 
% In argument generation, Hua et al. proposed an end-to-end framework that first retrieves relevant articles from Wikipedia with topic signatures from statement as queries. Then, the statement and the evidence are concatenated and encoded. During decoding, the keyphrase decoder first generates talking points as phrases, followed by the argument decoder which constructs the argument by attending both input and keyphrases~\cite{hua2018neural,hua2019argument}. 
Lewis et al. introduced a general retrieval-augmented generation (RAG) framework by leveraging a pre-trained neural retriever and generator. It can be easily fine-tuned on downstream tasks, and it has demonstrated state-of-the-art performance on various knowledge intensive NLG tasks~\cite{lewis2020retrieval}. Recently, the fusion-in-decoder methods (i.e., the decoder performs attention over the concatenation of the resulting representations of all retrieved passages~\cite{li2020leveraging,yu2021kg}) could even outperform RAG as reported in KILT benchmark~\cite{petroni2021kilt}.

\vspace{-0.05in}
\paragraph{\textbf{M1.2: Retrieve, rerank and rewrite ($R^{3}$)}}
Different from RAG, a $R^{3}$-based method is expected to retrieve a most precise reference document that can be directly used for rewriting/editing.
$R^{3}$-based method has proved successful in a number of NLG tasks such as machine translation~\cite{gu2018search}, and summarization~\cite{cao2018retrieve,wang2019biset}.
In summarization, Seq2Seq models that purely depend on the input document to generate summaries tend to deteriorate with the accumulation of word generation, e.g., they generate irrelevant and repeated words frequently~\cite{cao2018retrieve,wang2019biset}. Template-based summarization assume the golden summaries of the similar sentences (i.e., templates) can provide a reference point to guide the input sentence summarization process~\cite{cao2018retrieve,wang2019biset}. These templates are often called \textit{soft templates} in order to distinguish from the traditional rule-based templates. Soft template-based summarization typically follows a three-step design: retrieve, rerank, and rewrite. The step of retrieval aims to return a few candidate templates from a summary collection. The reranking identifies the best template from the retrieved candidates. And the rewriting leverages both the source document and template to generate more faithful and informative summaries. 

\vspace{-0.05in}
\paragraph{Difference between RAG and $R^3$} Compared with $R^3$-based methods, RAG-based have several differences, including less of emphasis on lightly editing a retrieved item, but on aggregating content from several pieces of retrieved content, as well as learning latent retrieval, and retrieving evidence documents rather than related training pairs.

\vspace{-0.05in}
\subsection{M2: Modeling Background Knowledge into Response Generation}

Background document, with more global and comprehensive knowledge, has been often used for generating informative responses and ensuring a conversation to not deviate from its topic. Keeping a conversation grounded on a background document is referred as background based conversation (BBC)~\cite{moghe2018towards,bi2020generating}. Background knowledge plays an important role in human-human conversations. For example, when talking about a movie, people often recall important points (e.g., a scene or review about the movie) and appropriately mention them in the conversation context. Therefore, an intelligent NLG model is expected to find an appropriate background snippet and generate response based on the snippet.
As shown in Figure \ref{fig:overall_kt}(b), the task of BBC is often compared with machine reading comprehension (MRC), in which a span is extracted from the background document as a response to a question~\cite{rajpurkar2016squad}. However, since BBC needs to generate natural and fluent responses, the challenge lies in not only locating the right semantic units in the background, but also referring to the right background information at the right time in the right place during the decoding phase.

As MRC models tie together multiple text segments to provide a unified and factual answer, many BBC models use the same idea to connect different pieces of information and find the appropriate background knowledge based on which the next response is to be generated~\cite{qin2019conversing,meng2020refnet}. For instance, Qin et al. proposed an end-to-end conversation model that jointly learned response generation together with on-demand machine reading~\cite{qin2019conversing}.
The MRC models can effectively encode the input utterance by treating it as a question in a typical QA task (e.g., SQuAD \cite{rajpurkar2016squad}) and encode the background document as the context. Then, they took the utterance-aware background representation as input into decoding phase.

\subsubsection{\textbf{Discussion and Analysis of Different Methods}}

% First, we generally discuss the strong points and weak points of the aforementioned methods that use grounded text to enhance text generation. Then, we provide qualitative analysis of the methods.

\vspace{-0.05in}
\paragraph{\textbf{Pros and cons.}}

For M1, guiding generation with retrieved information explicitly exposes the role of world knowledge by asking the model to decide what knowledge to retrieve and use during language generation. 
Since retrieval-augmented generation (RAG) captures knowledge in a interpretable and modular way, it is often used for knowledge-intensive tasks such as long-form QA and argument generation.
However, a knowledge retriever is expected to retrieve documents from a large-scale corpus, e.g., the entire Wikipedia, which causes significant computational challenge.
Besides, one input often requires retrieved text whose amount is much larger than the input itself (as indicated in Table~\ref{tab:data-kt}), leading to serious information overwhelming for the generation model. 

For M2, background based conversations (BBCs) avoid generating generic responses in a dialogue system and are able to generate more informative responses by exploring related background information. However, existing methods still cannot solve inherent problems effectively, such as tending to break a complete semantic unit and generate shorter responses~\cite{meng2020refnet}.

\vspace{-0.05in}
\paragraph{\textbf{Qualitative analysis.}}

Table \ref{tab:data-kt} summarizes tasks, datasets and evidence sources used in existing grounded text enhanced work. Three important things should be mentioned. First, all the datasets in the table are public, and we include their links in Table \ref{tab:code}.
Second, Wikipedia is the most commonly used evidence source since it is the largest free online encyclopedia. Besides, some online platforms contain plenty of product-related textural information, e.g., product reviews on Amazon, which are often used to build up task/goal oriented dialogue systems for business purpose.
Third, the retrieval space of candidate documents are usually larger than 1 million and only 7-10 documents are selected. So, the process of retrieving relevant documents is challenging.

Table \ref{tab:qual-kt} compares different grounded text enhanced methods from three dimensions: retrieval supervision, pre-training of the retriever, and number of stages. First, as mentioned above, retrieving relevant documents from a large candidate set is a challenging task. To improve the retrieval accuracy, four (57.1\%) papers added the retrieval supervision either by human annotated labels or pseudo labels, resulting in a multi-task learning setting. Besides, three (42.9\%) papers used pre-trained language models to produce document representation for better retrieval. Though existing work has greatly improved the retrieval accuracy, the performance is still far from satisfactory in many text generation tasks~\cite{lewis2020retrieval,krishna2021hurdles}.
How to learn mutually enhancement between retrieval and generation is still a promising direction in the grounded text enhanced text generation systems.

% In Table \ref{tab:quant-kg}, we make quantitative comparison between different KG-enhanced methods on four NLG tasks. All of the four tasks have (i) public resource; and (ii) been used by at least two papers. 
% \textcolor{red}{Observations and Conclusions from the tables must be clearly given in the text. Otherwise, readers might not be able to capture your idea or even not able to know where to start to read in the table, though they could find the tables.}

\vspace{-0.05in}
\section{Benchmark, Toolkit and Leaderboard Performance}
The development of general evaluation benchmarks for text generation helps to promote the development of research in related fields. Existing text generation benchmarks did not specially focus on choosing the tasks and datasets that have been widely used for knowledge-enhanced text generation. Therefore, we re-screened from the existing four text generation benchmarks, i.e., GLGE~\cite{liu2021glge}, GEM~\cite{gehrmann2021gem}, KilT~\cite{petroni2021kilt}, GENIE~\cite{khashabi2021genie}, and determined 9 benchmark datasets for evaluating knowledge-enhanced NLG methods. Here is our criteria for selection:
\begin{itemize}
    \item We only consider benchmark datasets that have open-access downloading link.
    \item We focus on diverse text generation tasks, involving various applications.
    \item We select at most three benchmark datasets for each text generation task.
    \item We include a mix of internal and external knowledge focused datasets.
    \item We prefer multi-reference datasets for robust automatic evaluation.
\end{itemize}

Based on the benchmark selection criteria, we finalize 9 knowledge-centric tasks that covers various NLG tasks, including commonsense reasoning, text summarization, question generation, generative question answering, and dialogue. The data statistics is shown in Table \ref{tab:benchmark}. Descriptions and dataset links are listed as follows:
\begin{itemize}
    \item \textbf{Wizard of Wikipedia (WOW):} It is an open-domain dialogue dataset, where two speakers conduct an open-ended conversion that is directly grounded with knowledge retrieved from Wikipedia. (Data link: \url{https://parl.ai/projects/wizard\_of\_wikipedia/})
    \item \textbf{CommonGen:} It is a generative commonsense reasoning dataset. Given a set of common concepts, the task is to generate a coherent sentence describing an everyday scenario using these concepts. (Data link: \url{https://inklab.usc.edu/CommonGen/})
    \item \textbf{$\alpha$NLG-ART:}  It is a generative commonsense reasoning dataset. Given the incomplete observations about the world, the task it to generate a valid hypothesis about the likely explanations to partially observable past and future. (Data link: \url{http://abductivecommonsense.xyz/})
    \item \textbf{ComVE:}  It is a generative commonsense reasoning dataset. The task is to generate an explanation given a counterfactual statement for sense-making. (Data link: \url{https://github.com/wangcunxiang/SemEval2020-Task4-Commonsense-Validation-and-Explanation}
    \item \textbf{ELI5:} It is a dataset for long-form question answering. The task is to produce explanatory multi-sentence answers for diverse questions. Web search results are used as evidence documents to answer questions. (Data link: \url{https://facebookresearch.github.io/ELI5/})
    \item \textbf{SQuAD:} It is a dataset for answer-aware question generation. The task is to generate a question asks towards the given answer span based on a given text passage or document. (Data link: \url{https://github.com/magic282/NQG})
    \item \textbf{CNN/DailyMail (CNN/DM):} It is a dataset for summarization. Given a news aticles, the goal is to produce a summary that represents the most important or relevant information within the original content. (Data link: \url{https://www.tensorflow.org/datasets/catalog/cnn_dailymail})
    \item \textbf{Gigaword:} It is a dataset for summarization. Similar with CNN/DM, the goal is to generate a headline for a news article. (Data link: \url{https://www.tensorflow.org/datasets/catalog/gigaword})
    \item \textbf{PersonaChat:} It is an open-domain dialogue dataset.
    It presents the task of making chit-chat more engaging by conditioning on profile information. (Data link: \url{https://github.com/facebookresearch/ParlAI/tree/master/projects/personachat})
    % \item \textbf{Holl-E:} It is a movie-related dialogue dataset. The goal is to conduct a meaningful conversation about a movie, one uses their background knowledge about the plot, reviews, comments and facts about the movie. (Data link: \url{https://github.com/nikitacs16/Holl-E})
\end{itemize}

\begin{table*}[t]
\caption{We choose 9 knowledge-enhanced NLG benchmark datasets. These datasets have been included in four existing general NLG benchmarks (i.e., GLGE~\cite{liu2021glge}, GEM~\cite{gehrmann2021gem}, Kilt~\cite{petroni2021kilt}, GENIE~\cite{khashabi2021genie}) or in SemEval tasks.}
\vspace{-0.15in}
\begin{center}
\scalebox{0.85}{\begin{tabular}{|l|c|l|r|r|r|c|c|l|}
\hline
\multirow{2}*{Tasks} & \multirow{2}*{Ref.} & \multicolumn{4}{c|}{Dataset Information} & Leader & In which NLG & Papers including \\
\cline{3-6}
& & Name & \#Train & \#Dev. & \#Test & board & benchmark & this dataset \\
\hline \hline
\multirow{3}*{\makecell[l]{Dialogue \\ system}} & \multirow{2}*{\makecell[l]{\cite{dinan2019wizard}}} & \multirow{2}*{\makecell[l]{Wizard of \\ Wikipedia}} & \multirow{2}*{18,430} & \multirow{2}*{1,948} & \multirow{2}*{1,933}  & \multirow{2}*{$\checkmark$\footnotemark[1]} & \multirow{2}*{Kilt} & \multirow{2}*{\cite{dinan2019wizard,kim2020sequential,lian2019learning}} \\
&&&&&&&& \\
\cline{2-9}
& \cite{zhang2018personalizing} & PersonaChat & 122,499 & 14,602 & 14,056 & $\times$ & GLGE & \cite{lian2019learning,dinan2019wizard} \\
% \cline{2-9}
% & \cite{moghe2018towards} & Holl-E & 7,228 & 930 & 913 & $\checkmark$\footnotemark[2] & - & \cite{meng2020refnet,kim2020sequential,chen2020bridging} \\
\hline \hline
\multirow{2}*{\makecell[l]{Question \\ answering}} & \multirow{2}*{\cite{fan2019eli5}} & \multirow{2}*{ELI5} & \multirow{2}*{272,634} & \multirow{2}*{1,507} & \multirow{2}*{600} & \multirow{2}*{$\checkmark$\footnotemark[3]} & \multirow{2}*{Kilt} & \multirow{2}*{\cite{petroni2021kilt,krishna2021hurdles}} \\
&&&&&&&& \\
\hline \hline
\multirow{2}*{\makecell[l]{Question \\ generation}} & \multirow{2}*{\cite{rajpurkar2016squad}} & \multirow{2}*{SQuAD} & \multirow{2}*{75,722} & \multirow{2}*{10,570} & \multirow{2}*{11,877} & \multirow{2}*{$\times$} & \multirow{2}*{GLGE} & \multirow{2}*{\cite{chen2020reinforcement,cho2019mixture,wang2020diversify}} \\
&&&&&&&& \\
\hline \hline
\multirow{3}*{\makecell[l]{Commonsense \\ reasoning}} & \cite{lin2020commongen} & CommonGen & 67,389 & 4,018 & 6,042 & $\checkmark$\footnotemark[4] & GEM & \cite{liu2021kg,fan2020enhanced,wang2021retrieval} \\
\cline{2-9}
& \cite{bhagavatula2020abductive} & $\alpha$NLG-ART & 50,481 & 7,252 & 2,976 & $\checkmark$\footnotemark[5] & GENIE & \cite{bhagavatula2020abductive,ji2020language} \\
\cline{2-9}
& \cite{wang2020semeval} & ComVE & 25,596 & 1,428 & 2,976 & $\checkmark$\footnotemark[6] & SemEval & \cite{ji2020language,ji2020generating} \\
\hline \hline
\multirow{2}*{\makecell[l]{Summarization}} & \cite{see2017get} & CNN/DM & 287,226 & 13,368 & 11,490 & $\checkmark$\footnotemark[7] & GLGE & \cite{fu2020document,zhu2020boosting,gehrmann2018bottom} \\ 
\cline{2-9}
& \cite{see2017get} & Gigaword & 3.8M & 189K & 1,951 & $\checkmark$\footnotemark[8] & GLGE & \cite{li2018guiding,jin2020semsum,cao2018retrieve} \\ 
\hline
\end{tabular}}
\end{center}
\vspace{-0.1in}
\label{tab:benchmark}
\end{table*}

\footnotetext[1]{\url{https://parl.ai/projects/wizard\_of\_wikipedia}}
\footnotetext[2]{\url{https://nikitacs16.github.io/holl-e-website/}}
\footnotetext[3]{\url{https://facebookresearch.github.io/ELI5/}}
\footnotetext[4]{\url{https://inklab.usc.edu/CommonGen/leaderboard.html}}
\footnotetext[5]{\url{https://leaderboard.allenai.org/genie-anlg/submissions/public}}
\footnotetext[6]{\url{https://competitions.codalab.org/competitions/21080\#results}}
\footnotetext[7]{\url{https://paperswithcode.com/sota/document-summarization-on-cnn-daily-mail}}
\footnotetext[8]{\url{https://paperswithcode.com/sota/text-summarization-on-gigaword}}

\vspace{-0.05in}
\section{Discussion on Future Directions}
Many efforts have been conducted to tackle the problem of knowledge-enhanced text generation and its related applications. To advance the field, there remains several open problems and future directions. Designing more effective ways to represent knowledge and integrate them into the generation process is still the most important trend in knowledge-enhanced NLG systems.
% For examples, to improve knowledge relevance, Cao et al. designed knowledge related multi-task learning~\cite{cao2018retrieve}; Lian et al.  separate the posterior distribution from the prior distribution~\cite{lian2019learning}. To improve knowledge richness, Fu et al. combined both structured (knowledge base) and unstructured knowledge (grounded text)~\cite{fu2018natural}. 
% For example, multi-task learning can make mutual enhancement between knowledge representation and text generation~\cite{cao2018retrieve,li2020keywords}.
From a broader perspective, we provide three directions that make focusing such efforts worthwhile now: (i) incorporating knowledge into visual-language generation tasks, (ii) learning knowledge from broader sources, especially pre-trained language models, (iii) learning knowledge from limited resources, (iv) learning knowledge in a continuous way.

\vspace{-0.05in}
\subsection{Incorporate Knowledge into Visual-Language Generation Tasks}

Beyond text-to-text generation tasks, recent years have witnessed a growing interest in visual-language (VL) generation tasks, such as describing visual scenes~\cite{hossain2019comprehensive}, and answering visual-related questions~\cite{mao2019neuro}.
% , a task that is remarkably easy for humans yet remains difficult for machines~\cite{hossain2019comprehensive}. 
% Image/video captioning problem which requires model to analyze the visual content of an image/video, and then generate a caption gains particular interest.
% Of particularly interest in image/video captioning problem requires analyzing the visual content of an image/video, and generating a caption.
Although success has been achieved in recent years on VL generation tasks, there is still room for improvement due to the fact that image-based factual descriptions are often not enough to generate high-quality captions or answers~\cite{zhou2019improving}. External knowledge can be added in order to generate attractive image/video captions. We observed some pioneer work has attempted to utilize external knowledge to enhance the image/video captioning tasks. For example, Tran et al. proposed to detect a diverse set of visual concepts and generate captions by using an external knowledge base (i.e., Freebase), in recognizing a broad range of entities such as celebrities and landmarks~\cite{tran2016rich}. 
% \citet{wu-etal-2019-generating} tries to jointly generate question-related captions to provide additional information for VQA.
% \citet{ziaeefard-lecue-2020-towards,shevchenko-etal-2021-reasoning} inject extenal knowledge base such as  ConceptNet and Wikidata into visual-text transformers to improve VQA performance.
Zhou et al. used a commonsense knowledge graph (i.e., ConceptNet), to infer a set of terms directly or indirectly related to the words that describe the objects found in the scene by the object recognition module~\cite{zhou2019improving}.
% \citet{cho2021vlt5} proposes a unified framework for visual-language generation tasks.
In addition,
\citet{mao2019neuro} proposed a neuro-symbolic learner for improving visual-language generation tasks (e.g., visual question answering).

However, existing approaches for knowledge-enhanced visual-language generation tasks still have a lot of space for exploration. Some promising directions for future work include using other knowledge sources, such as retrieving image/text to help solve open-domain visual question answering and image/video captioning tasks; bringing structured knowledge for providing justifications for the captions that they produce, tailoring captions to different audiences and contexts, etc.

\vspace{-0.05in}
\subsection{Learning Knowledge from Broader Sources}

More research efforts should be spent on learning to discover knowledge more broadly and combine multiple forms of knowledge from different sources to improve the generation process. More knowledge sources can be but not limited to network structure, dictionary and table. For examples, Yu et al.~\cite{yu2020identifying} and An et al.~\cite{an2021enhancing} augmented the task of scientific papers intention detection and summarization by introducing the citation graph; Yu et al. augmented the rare word representations by retrieving their descriptions from Wiktionary and feed them as additional input to a pre-trained language model~\cite{yu2021dict}.
Besides, structured knowledge and unstructured knowledge can play a complementary role in enhancing text generation. To improve knowledge richness, Fu et al. combined both structured (knowledge base) and unstructured knowledge (grounded text)~\cite{fu2018natural}.

\vspace{-0.05in}
\paragraph{Leveraging Knowledge from Pre-trained Language Models}

Pre-trained language models can learn a substantial amount of in-depth knowledge from data without any access to an external memory, as a parameterized implicit knowledge base~\cite{lewis2020retrieval,raffel2020exploring}.
However, as mentioned in~\cite{guan2020knowledge}, directly fine-tuning pre-trained language generation models on the story generation task still suffers from 
% many same problems without pre-training, such as
insufficient knowledge by representing the input text thorough a pre-trained encoder,
leading to repetition, logic conflicts, and lack of long-range coherence in the generated output sequence.
Therefore, discovering knowledge from pre-trained language models can be more flexible, such as knowledge distillation, data augmentation, and using pre-trained models as external knowledge~\cite{petroni2019language}. More efficient methods of obtaining knowledge from pre-trained language models are expected.

\vspace{-0.05in}
\subsection{Learning Knowledge from Limited Resources}

Most of current NLG research conduct on extensively labelled data to favor model training. However, this is in contrast to many real-world application scenarios, where only a few shots of examples are available for new domains. Limited data resources lead to \textit{limited knowledge} that can be learnt in new domains. For examples, learning topical information of a dialogue occurring under a new domain is difficult since the topic may be rarely discussed before; 
% leveraging OpenIE to construct an internal knowledge graph in a new domain is difficult since many unseen entities and relations cannot be extracted. 
constructing a syntactic dependency graph of a sequence in a low-resource language is hard since many linguistic features are of great uniqueness.
Besides, external knowledge bases are often incomplete and insufficient to cover full entities and relationships due to the human costs of collecting domain-specific knowledge triples. Therefore, quick domain adaptation is an essential task in text generation tasks. One potential route towards addressing these issues is meta-learning, which in the context of NLG means a generation model develops a broad set of skills and pattern recognition abilities at training time, and quickly adapt to a new task given very few examples without retraining the model from scratch. Recently, there has been raising interests in both academia and industry to investigate meta-learning in different NLG tasks.
% such as dialogue system~\cite{song2020learning} and machine translation~\cite{gu2018meta}. 
Thus, it is a promising research direction to build efficient meta-learning algorithms that only need a few task-specific fine-tuning to learn the new task quickly. And for knowledge-enhanced text generation, it is of crucial importance to adapt the model quickly on new domains with limited new knowledge (e.g., only a few knowledge triples).

\vspace{-0.05in}
\subsection{Learning Knowledge in a Continuous Way}

A machine learning is expected to learn continuously, accumulate the knowledge learned in previous tasks, and use it to assist future learning. This research direction is referred as lifelong learning~\cite{chen2018lifelong}. In the process, the intelligent machine becomes more and more knowledgeable and effective at learning new knowledge. To make an analogy, humans continuously acquire new knowledge and constantly update the knowledge system in the brain. However, existing knowledge-enhanced text generation systems usually do not keep updating knowledge in real time (e.g., knowledge graph expansion). 
A meaningful exploration of was discussed in~\cite{mazumder2018towards}. They built a general knowledge learning engine for chatbots to enable them to continuously and interactively learn new knowledge during conversations.
Therefore, it is a promising research direction to 
continuously update knowledge obtained from various information sources, empowering intelligent machines with incoming knowledge and improving the performance on new text generation tasks.

\vspace{-0.05in}
\section{Conclusions}

In this survey, we present a comprehensive review  of current representative research efforts and trends on knowledge-enhanced text generation, and expect it can facilitate future research. To summarize, this survey aims to answer two questions that commonly appears in knowledge-enhanced text generation: \textit{how to acquire knowledge} and \textit{how to incorporate knowledge to facilitate text generation}. Base on knowledge acquisition, the main content of our survey is divided into three sections according to different sources of knowledge enhancement. Based on knowledge incorporation, we first present general methods of incorporating knowledge into text generation and further discuss a number of specific ideas and technical solutions that incorporate the knowledge to enhance the text generation systems in each section. Besides, we review a variety of  text generation applications in each section to help practitioners learn to choose and employ the methods.

\vspace{-0.05in}
\section*{Acknowledgements}
We thank all anonymous reviewers for valuable comments. We also appreciate the suggestions from readers of the pre-print version.
We thank Dr. Michael Zeng (Microsoft) and Dr. Nazneen Rajani (Saleforce) for their constructive comments and suggestions.
Wenhao Yu and Dr. Meng Jiang's research is supported by National Science Foundation grants IIS-1849816, CCF-1901059, and IIS-2119531.
Qingyun Wang and Dr. Heng Ji's research is based upon work supported by Agriculture and Food Research Initiative (AFRI) grant no. 2020-67021-32799/project accession no.1024178 from the USDA National Institute of Food and Agriculture, U.S. DARPA SemaFor Program No. HR001120C0123, DARPA AIDA Program No. FA8750-18-2-0014, and DARPA KAIROS Program No. FA8750-19-2-1004. The views and conclusions contained herein are those of the authors and should not be interpreted as necessarily representing the official policies, either expressed or implied, of DARPA, or the U.S. Government. The U.S. Government is authorized to reproduce and distribute reprints for governmental purposes notwithstanding any copy-right annotation therein.

\bibliographystyle{ACM-Reference-Format}
\bibliography{reference_short}

\clearpage
\appendix
\section{Appendix}
\begin{figure}[hb]
  \vspace{-0.1in}
  \begin{center}
    \includegraphics[width=0.48\textwidth]{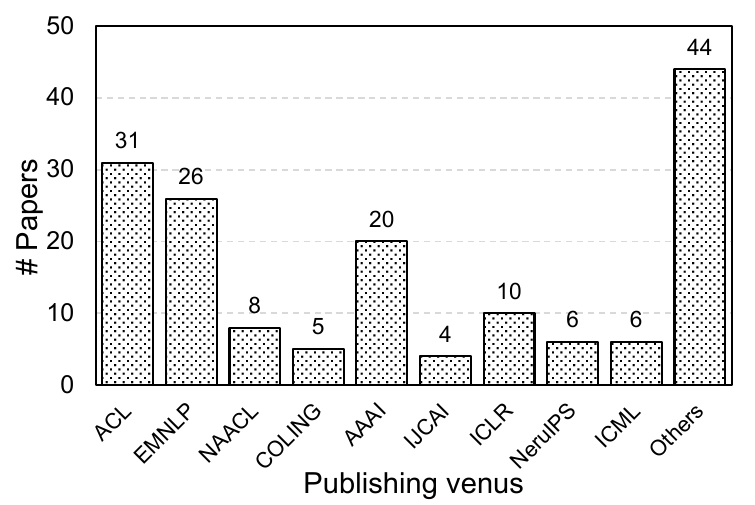}
    \includegraphics[width=0.48\textwidth]{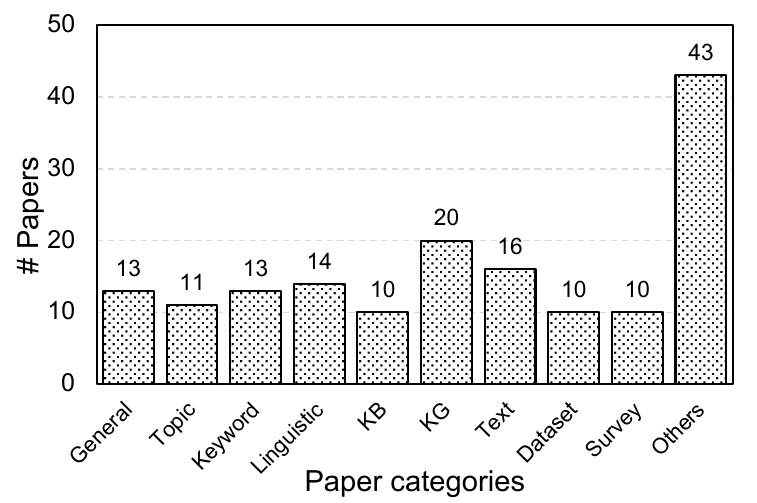}
  \end{center}
  \vspace{-0.1in}
  \caption{Paper statistics of selected publications in this survey.}
  \vspace{-0.2in}
  \label{fig:paper-stat}
\end{figure}

\vspace{0.05in}
\noindent \textbf{Figure \ref{fig:paper-stat}} demonstrates the statistics of selected publications in this survey. The left figure shows the paper publishing venues. Most papers were published in top machine learning, artificial intelligence, and natural language processing conferences, such as ACL, EMNLP, AAAI, ICLR, NeurIPS. Besides, many selected papers were published in high-impact journals, such as TNNLS, JMLR, TACL. The right figure shows the paper categories. Among 160 selected papers, 87 papers (``general methods (General)'', ``topic'', ``keyword'', ``knowledge base (KG)'', ``knowledge graph (KG)'', ``grounded text (Text)'') are directly relevant to the different kinds of knowledge-enhanced text generation methods; 10 papers are relevant to benchmark datasets; 10 papers are related survey papers. Besides, other 43 papers are about basic (pre-trained) generation methods (e.g., Seq2Seq, CopyNet, BART, T5), or necessary background (e.g., TransE, OpenIE, GNN, LDA), or future direction. 

\begin{figure}[hb]
  \begin{center}
    \includegraphics[width=0.7\textwidth]{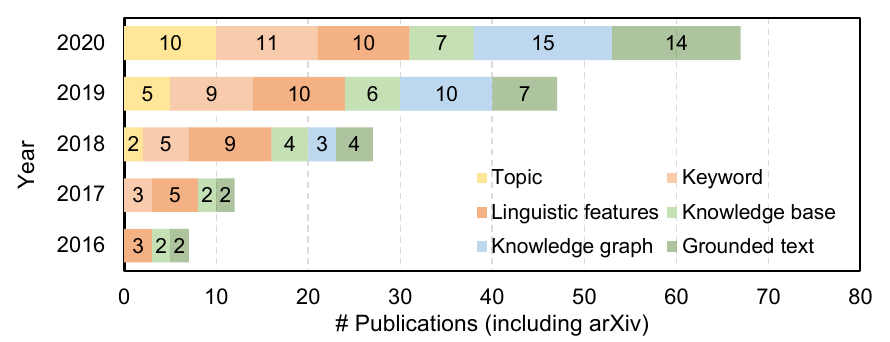}
  \end{center}
  \vspace{-0.2in}
  \caption{Knowledge-enhanced text generation has been gaining emerging interests in the recent five years.}
  \label{fig:paper-year}
\vspace{-0.1in}
\end{figure}

\vspace{0.05in}
\noindent \textbf{Figure \ref{fig:paper-year}} summarized different papers according to years, knowledge sources, and methods. 

\vspace{0.05in}
\noindent \textbf{Table \ref{tab:leaderboard}} lists the leaderboard performance on ten knowledge-enhanced generation benchmarks. 

\vspace{0.05in}
\noindent\textbf{Table \ref{tab:code}} lists code links and programming language of representative open-source knowledge-enhanced text generation systems that have been introduced in this survey. \\

\subsection{Evaluation Metrics}

\vspace{0.05in}
\noindent \textbf{BLEU-$m$ (short as B-$m$):} BLEU is a weighted geometric mean of $n$-gram precision scores.

\vspace{0.05in}
\noindent \textbf{ROUGE-$m$ (short as R-$m$):} ROUGE measures the overlap of n-grams between the reference and hypothesis; ROUGE-L measures the longest matched words using longest common sub-sequence.

\vspace{0.05in}
\noindent \textbf{Distinct-$k$ (short as D-k):} Distinct measures the total number of unique $k$-grams normalized by the total number of generated $k$-gram tokens to avoid favoring long sentences.

\begin{table}
\caption{Leaderboard performance on ten knowledge-enhanced generation benchmarks.}
\vspace{-0.1in}
\begin{subtable}[t]{1.0\textwidth}
\caption{Leaderboard performance on two summarization benchmark datasets with different knowledge-enhanced NLG methods. Evaluation metrics are standard n-gram based metrics: ROUGE-2 and ROUGE-L.}
\vspace{-0.1in}
\begin{center}
\scalebox{0.83}{\begin{tabular}{|l|c|l|l|cc|cc|l|}
\hline
{\multirow{2}*{Methods}} & \multirow{2}*{Ref.} & Knowledge & Method & \multicolumn{2}{c|}{\textbf{CNN/DM}} & 
\multicolumn{2}{c|}{\textbf{Gigaword}} & \\
 & & source & category & R-2 & R-L & R-2 & R-L & \\
\hline \hline
\rowcolor{gray!12}\multicolumn{9}{|c|}{Baseline methods (w/o KG)} \\ \hline
Seq2Seq & \cite{sutskever2014sequence} & & & 11.81 & 28.83 & 11.32 & 26.42 & with attention mechanism \\
PG & \cite{see2017get} & & & 15.66 & 33.42 & 17.63 & 33.66 & w/o coverage mechanism \\
\hline \hline
\rowcolor{gray!12}\multicolumn{9}{|c|}{Knowledge enhanced methods} \\ \hline
VHTM & \cite{fu2020document} & Topic & M3 & 18.05 & 37.18 & - & - & -  \\
SELECTOR & \cite{cho2019mixture} & Keyword & M2 & 18.31 & - & - & - & * Improve generation diversity \\
FASUM & \cite{zhu2020boosting} & OpenKG & - & 17.84 & 37.40 & - & - & * Improve factual correctness \\
KIGN & \cite{li2018guiding} & Keyword & M1 & 17.12 & 35.68 & 17.93 & 34.44 & -  \\
BottomUp & \cite{gehrmann2018bottom} & Keyword & M2 & 18.68 & 38.34 & 17.61 & 33.54 & - \\
TGVAE & \cite{wang2019topic} & Topic & M3 & - & - & 17.27 & 33.02 & - \\
HierDualPG & \cite{li2020keywords} & Keyword & M2 & - & - & 18.06 & 34.39 & - \\
R$^3$Sum & \cite{cao2018retrieve} & Text & M1 & - & - & 19.03 & 34.46 & -  \\
BiSET & \cite{wang2019biset} & Text & M1 & - & - & \textbf{19.78} & \textbf{36.87} & - \\ 
SemSUM & \cite{jin2020semsum} & DepGraph & - & - & - & 19.75 & 36.09 & - \\ 
ASGARD & \cite{huang2020knowledge} & OpenKG & - & \textbf{20.37} & \textbf{40.48} & - & - & - \\
\hline
\end{tabular}}
\end{center}
\label{tab:quant-kg}
\end{subtable}
\begin{subtable}[t]{1.0\textwidth}
\begin{minipage}{0.48\linewidth}
\centering
\vspace{0.05in}
\caption{Leaderboard performance on $\alpha$NLG-ART dataset. Both B-4 and R-L are commonly used.}
\vspace{-0.05in}
\scalebox{0.9}{\begin{tabular}{|l|c|l|c|c|} 
\hline
Method & Ref. & Source & B-4 & R-L \\
\hline \hline
\rowcolor{gray!12}\multicolumn{5}{|c|}{Baseline methods} \\
\hline  
Seq2Seq & \cite{sutskever2014sequence} & - & 2.37 & 22.30 \\
\hline  
GPT-2 &  \cite{radford2019language} & - & 9.80 & 32.90 \\
\hline \hline
\rowcolor{gray!12}\multicolumn{5}{|c|}{Knowledge-enhanced methods} \\
\hline 
GPT-COMeT & \cite{bhagavatula2020abductive} & KG & 9.62 & 32.88 \\
\hline 
GRF & \cite{liu2021kg} & KG & \textbf{11.62} & \textbf{34.62} \\
\hline 
\end{tabular}}
\end{minipage}
\hspace{0.15in}
\begin{minipage}{0.47\linewidth}
\centering
\vspace{0.05in}
\caption{Leaderboard performance on ComVE dataset. Both B-4 and R-L are commonly used.}
\vspace{-0.05in}
\scalebox{0.9}{\begin{tabular}{|l|c|l|c|c|} 
\hline
Method & Ref. & Source & B-4 & R-L \\
\hline \hline
\rowcolor{gray!12}\multicolumn{5}{|c|}{Baseline methods} \\
\hline  
Seq2Seq & \cite{sutskever2014sequence} & - & 6.10 & 25.80 \\
\hline  
GPT-2 &  \cite{radford2019language} & - & 15.70 & 36.50 \\
\hline \hline
\rowcolor{gray!12}\multicolumn{5}{|c|}{Knowledge-enhanced methods} \\
\hline 
CE-PR & \cite{ji2020generating} & KG & 17.10 & 37.90 \\
\hline 
GRF & \cite{ji2020language} & KG & \textbf{17.19} & \textbf{38.10} \\
\hline 
\end{tabular}}
\end{minipage}
\end{subtable}
\begin{subtable}[t]{1.0\textwidth}
\begin{minipage}{0.48\linewidth}
\centering
\vspace{0.05in}
\caption{Leaderboard performance on CommonGen dataset. SPICE is the primary evaluation metric.}
\vspace{-0.05in}
\scalebox{0.9}{\begin{tabular}{|l|c|l|c|c|} 
\hline
Method & Ref. & Source & B-4 & SPICE \\
\hline \hline
\rowcolor{gray!12}\multicolumn{5}{|c|}{Baseline methods} \\
\hline  
BART & \cite{lewis-etal-2020-bart} & - & 31.83 & 27.99 \\
\hline  
T5 &  \cite{raffel2020exploring} & - & 31.96 & 28.86 \\
\hline \hline
\rowcolor{gray!12}\multicolumn{5}{|c|}{Knowledge-enhanced methods} \\
\hline 
EKI-BART & \cite{fan2020enhanced} & Text & 35.95 & 29.59 \\
\hline 
KG-BART & \cite{liu2021kg} & KG & 33.87 & 29.63 \\
\hline 
RE-T5 & \cite{wang2021retrieval} & Text & \textbf{40.87} & \textbf{31.08} \\
\hline 
\end{tabular}}
\end{minipage}
\hspace{0.15in}
\begin{minipage}{0.47\linewidth}
\centering
\vspace{0.05in}
\caption{Leaderboard performance on Holl-E (mix-short setting) dataset. R-L are the primary metric.}
\vspace{-0.05in}
\scalebox{0.9}{\begin{tabular}{|l|c|l|c|c|} 
\hline
Method & Ref. & Source & R-L & B-4 \\
\hline \hline
\rowcolor{gray!12}\multicolumn{5}{|c|}{Baseline methods} \\
\hline  
Seq2Seq & \cite{sutskever2014sequence} & - & 21.48 & 5.26 \\
\hline  
BiDAF &  \cite{seo2017bidirectional} & - & 35.09 & 27.44 \\
\hline \hline
\rowcolor{gray!12}\multicolumn{5}{|c|}{Knowledge-enhanced methods} \\
\hline 
AKGCM & \cite{liu2019knowledge} & KG & 34.72 & \textbf{30.84} \\
\hline 
RefNet & \cite{meng2020refnet} & Text & 36.17 & 29.38 \\
\hline 
GLKS & \cite{ren2020thinking} & Text & \textbf{39.63} & - \\
\hline
\end{tabular}}
\end{minipage}
\end{subtable}
\label{tab:leaderboard}
\end{table}

\begin{table}
\ContinuedFloat
\vspace{-0.1in}
\begin{subtable}[t]{1.0\linewidth}
\begin{minipage}{0.48\linewidth}
\centering
\vspace{0.05in}
\caption{Leaderboard performance on Wizard of Wikipedia with seen (S) and unseen (UnS) test set. }
\vspace{-0.05in}
\scalebox{0.85}{\begin{tabular}{|l|c|c|c|} 
\hline
Method & Ref. & R-1/R-2 (S) & R-1/R-2 (UnS) \\
\hline \hline
\rowcolor{gray!12}\multicolumn{4}{|c|}{Baseline methods} \\
\hline  
Transformer & \cite{vaswani2017attention} & 17.8/ --- & 14.0/ --- \\
\hline \hline
\rowcolor{gray!12}\multicolumn{4}{|c|}{Knowledge-enhanced methods} \\
\hline 
MemNet & \cite{dinan2019wizard} & 16.9/ --- & 14.4/ --- \\
\hline 
PostKS & \cite{lian2019learning} & 18.1/5.3 & 13.5/2.0 \\
\hline 
SKT & \cite{kim2020sequential} & 19.3/6.8 & 16.1/4.2 \\
\hline 
PIPM+KDBTS & \cite{chen2020bridging} & \textbf{19.9}/\textbf{7.3} & \textbf{17.6}/\textbf{5.4} \\
\hline 
\end{tabular}}
\end{minipage}
\hspace{0.15in}
\begin{minipage}{0.47\linewidth}
\centering
\vspace{0.05in}
\caption{State-of-the-art performance on SQuAD.}
\vspace{-0.05in}
\scalebox{0.87}{\begin{tabular}{|l|c|l|c|} 
\hline
Method & Ref. & Source & B-4 \\
\hline \hline
\rowcolor{gray!12}\multicolumn{4}{|c|}{Baseline methods} \\
\hline  
Seq2Seq & \cite{sutskever2014sequence} & - & 3.01 \\
\hline  
Transformer &  \cite{vaswani2017attention} & - & 3.09 \\
\hline \hline
\rowcolor{gray!12}\multicolumn{4}{|c|}{Knowledge-enhanced methods} \\
\hline 
NQG++ & \cite{zhou2017neural} & LF & 13.27 \\
\hline 
SELECTOR & \cite{cho2019mixture} & LF+Keyword & 15.87 \\
\hline 
G2S+BERT & \cite{chen2020reinforcement} & LF+DepGraph & 17.49 \\
\hline 
G2S+BERT+RL & \cite{chen2020reinforcement} & LF+DepGraph & \textbf{18.30} \\
\hline
\end{tabular}}
\end{minipage}
\end{subtable}
\begin{subtable}[t]{1.0\linewidth}
\begin{minipage}{0.48\linewidth}
\centering
\vspace{0.05in}
\caption{Leaderboard performance on ELI5 dataset. The Kilt R-L (KRL) is the primary evaluation metric.}
\vspace{-0.05in}
\scalebox{0.9}{\begin{tabular}{|l|c|l|c|c|} 
\hline
Method & Ref. & Source & KRL & R-L \\
\hline \hline
\rowcolor{gray!12}\multicolumn{5}{|c|}{Baseline methods} \\
\hline  
T5 & \cite{raffel2020exploring} & - & 0.0 & 19.1 \\
\hline  
BART & \cite{lewis-etal-2020-bart} & - & 0.0 & 20.1 \\
\hline \hline
\rowcolor{gray!12}\multicolumn{5}{|c|}{Knowledge-enhanced methods} \\
\hline 
RAG & \cite{lewis2020retrieval} & Text & 1.7 & 17.4 \\
\hline 
BART+DPR & \cite{petroni2021kilt} & Text & 1.9 & 17.4 \\
\hline 
RT+\textsc{c}-REALM & \cite{kim2020sequential} & Text & \textbf{2.4} & \textbf{23.2} \\
\hline 
\end{tabular}}
\end{minipage}
\hspace{0.15in}
\begin{minipage}{0.47\linewidth}
\centering
\vspace{0.05in}
\caption{Some state-of-the-art performance on PersonaChat dataset (no leaderboard on this dataset).}
\vspace{-0.05in}
\scalebox{0.9}{\begin{tabular}{|l|c|c|c|} 
\hline
Method & Ref. & B-1/B-2 & D-1/D-2\\
\hline \hline
\rowcolor{gray!12}\multicolumn{4}{|c|}{Baseline methods} \\
\hline  
Seq2Seq & \cite{sutskever2014sequence} & 18.2/9.3 & 2.6/7.4 \\
\hline \hline
\rowcolor{gray!12}\multicolumn{4}{|c|}{Knowledge-enhanced methods} \\
\hline 
MemNet(soft) & \cite{dinan2019wizard} & 17.7/9.1 & 3.5/9.6 \\
\hline 
MemNet(hard) & \cite{dinan2019wizard} & 18.6/9.7 & 3.7/9.9 \\
\hline 
% PostKS(concat) & \cite{lian2019learning} & 18.2/9.6 & \textbf{4.8}/12.6 \\
% \hline 
% PostKS(fusion) & \cite{lian2019learning} & \textbf{19.0}/\textbf{9.8} & 4.6/\textbf{13.4} \\
PostKS & \cite{lian2019learning} & 19.0/9.8 & \textbf{4.6}/\textbf{13.4} \\
\hline 
PEE & \cite{xu2020neural} & \textbf{23.2}/\textbf{11.5} & - / - \\
\hline 
\end{tabular}}
\end{minipage}
\end{subtable}
\end{table}

\begin{table*}[t]
\caption{A list of representative open-source knowledge-enhanced text generation systems.}
\vspace{-0.15in}
\begin{center}
\scalebox{0.72}{\begin{tabular}{|l|c|l|c|l|}
\hline
\multirow{2}*{Task} & \multirow{2}*{Ref.} &  \multirow{2}*{Paper title and open source code/toolkit} & Programming & Venue \\
& & & language & \& Year \\
\hline \hline
\rowcolor{gray!12}\multicolumn{5}{|c|}{Topic-enhanced methods} \\
\hline \hline
\multirow{4}*{\makecell[l]{Summarization}} & \multirow{2}*{\cite{narayan2018don}} & Topic-Aware Convolutional Neural Networks for Extreme Summarization & \multirow{2}*{PyTorch} & EMNLP \\
& & ------ Code: \url{https://github.com/EdinburghNLP/XSum} &  & 2018 \\
\cline{2-5}
& \multirow{2}*{\cite{wu2019global}} & Friendly Topic Assistant for Transformer Based Abstractive Summarization & \multirow{2}*{-} & EMNLP \\
& & ------ Code: \url{https://github.com/BoChenGroup/TA} & & 2020 \\
\hline \hline
\rowcolor{gray!12}\multicolumn{5}{|c|}{Keyword-enhanced methods} \\
\hline \hline
\multirow{4}*{\makecell[l]{Dialogue \\ system}} & \multirow{2}*{\cite{narayan2018don}} & A Content-Introducing Approach to Generative Short-Text Conversation & \multirow{2}*{Tensorflow} & COLING \\
& & ------ Code: \url{https://github.com/MaZhiyuanBUAA/Seq2BFforDialogueGeneration} &  & 2016 \\
\cline{2-5}
& \multirow{2}*{\cite{wu2019global}} & Emotional Chatting Machine: Emotional Conversation Generation with & \multirow{2}*{PyTorch} & AAAI \\
& & Internal and External Memory ------ Code: \url{https://github.com/loadder/ECM-tf} & & 2018 \\
\hline \hline
\multirow{2}*{\makecell[l]{Summarization}} & \multirow{2}*{\cite{narayan2018don}} & Coherent Comment Generation with a Graph-to-Sequence Model & \multirow{2}*{PyTorch} & ACL \\
& & ------ Code: \url{https://github.com/lancopku/Graph-to-seq-comment-generation} &  & 2018 \\
\hline \hline
\rowcolor{gray!12}\multicolumn{5}{|c|}{KB-enhanced methods} \\
\hline \hline
\multirow{10}*{\makecell[l]{Dialogue \\ system}} & \multirow{2}*{\cite{madotto2018mem2seq}} & Mem2Seq: Effectively Incorporating Knowledge Bases into End-to-End & \multirow{2}*{PyTorch} & ACL \\
& & Dialog Systems ------ Code: \url{https://github.com/HLTCHKUST/Mem2Seq} &  & 2019 \\
\cline{2-5}
& \multirow{2}*{\cite{wu2019global}} & Global-to-local Memory Pointer Networks for Task-Oriented Dialogue & \multirow{2}*{PyTorch} & ICLR \\
& & ------ Code: \url{https://github.com/jasonwu0731/GLMP} & & 2019 \\
\cline{2-5}
& \multirow{2}*{\cite{wang2020improving}} & Improving Knowledge-aware Dialogue Generation via Knowledge Base & \multirow{2}*{PyTorch} & AAAI \\
& & Question Answering ------ Code: \url{https://github.com/siat-nlp/TransDG} & & 2020 \\
\cline{2-5}
& \multirow{2}*{\cite{wu2020diverse}} & Diverse and Informative Dialogue Generation with Context-Specific Knowledge & \multirow{2}*{Tensorflow} & ACL \\
& & Awareness ------ Code: \url{https://github.com/pku-sixing/ACL2020-ConKADI} & & 2020 \\
\cline{2-5}
& \multirow{2}*{\cite{wu2020topicka}} & TopicKA: Generating Commonsense Knowledge-Aware Dialogue Responses  & \multirow{2}*{Tensorflow} & IJCAI \\
& & ------ Code: \url{https://github.com/pku-sixing/IJCAI2020-TopicKA} & & 2020 \\
\hline \hline
\rowcolor{gray!12}\multicolumn{5}{|c|}{KG-enhanced methods} \\
\hline \hline
\multirow{6}*{\makecell[l]{Dialogue \\ system}} & \multirow{2}*{\cite{zhou2018commonsense}} & Commonsense Knowledge Aware Conversation Generation with Graph & \multirow{2}*{Tensorflow} & IJCAI \\
& & Attention ------ Code: \url{https://github.com/thu-coai/ccm} &  & 2018 \\
\cline{2-5}
& \multirow{2}*{\cite{tuan2019dykgchat}} & DyKgChat: Benchmarking Dialogue Generation Grounding on Dynamic & \multirow{2}*{Tensorflow} & EMNLP \\
& &  Knowledge Graphs ------ Code: \url{https://github.com/Pascalson/DyKGChat} & & 2019 \\
\cline{2-5}
& \multirow{2}*{\cite{zhang2020grounded}} & Grounded Conversation Generation as Guided Traverses in Commonsense & \multirow{2}*{PyTorch} & ACL \\
& & Knowledge Graphs ------ Code: \url{https://github.com/thunlp/ConceptFlow} & & 2020 \\
\hline \hline
\multirow{4}*{\makecell[l]{Scientific \\ writing}} & \multirow{2}*{\cite{koncel2019text}} & Text Generation from Knowledge Graphs with Graph Transformers & \multirow{2}*{PyTorch} & NAACL \\
& & ------ Code: \url{https://github.com/rikdz/GraphWriter} & & 2019 \\
\cline{2-5}
& \multirow{2}*{\cite{wang2019paperrobot}} & PaperRobot: Incremental Draft Generation of Scientific Ideas & \multirow{2}*{PyTorch} & ACL \\
& & ------ Code: \url{https://github.com/EagleW/PaperRobot} & & 2019 \\
\hline \hline
\multirow{8}*{\makecell[l]{Commonsense \\ reasoning \\ \&\\ Story \\ generation }} & \multirow{2}*{\cite{guan2019story}} & Story Ending Generation with Incremental Encoding and Commonsense & \multirow{2}*{Tensorflow} & AAAI \\
& & Knowledge ------ Code: \url{https://github.com/JianGuanTHU/StoryEndGen} & & 2019 \\
\cline{2-5}
& \multirow{2}*{\cite{ji2020language}} & Language Generation with Multi-Hop Reasoning on Commonsense  & \multirow{2}*{PyTorch} & EMNLP \\
& & Knowledge Graph ------ Code: \url{https://github.com/cdjhz/multigen} & & 2020 \\
\cline{2-5}
& \multirow{2}*{\cite{liu2021kg}} & KG-BART: Knowledge Graph-Augmented BART for Generative   & \multirow{2}*{PyTorch} & AAAI \\
& & Commonsense Reasoning ------ Code: \url{https://github.com/yeliu918/KG-BART} & & 2021 \\
\cline{2-5}
& \multirow{2}*{\cite{cheng2020entdesc}} & ENT-DESC: Entity Description Generation by Exploring Knowledge Graph & \multirow{2}*{MXNet} & EMNLP \\
& & ------ Code: \url{https://github.com/LiyingCheng95/EntityDescriptionGeneration} & & 2020 \\
% \cline{2-5}
% & \multirow{2}*{\cite{guan2020knowledge}} & A Knowledge-Enhanced Pretraining Model for Commonsense Story Generation & \multirow{2}*{Tensorflow} & TACL \\
% & &  ------ Code: \url{https://github.com/JianGuanTHU/CommonsenseStoryGen} & & 2020 \\
\hline \hline
\multirow{2}*{\makecell[l]{Question \\ answering}} & \multirow{2}*{\cite{bauer2018commonsense}} & Commonsense for Generative Multi-Hop Question Answering Tasks & \multirow{2}*{Tensorflow} & EMNLP \\
& & ------ Code: \url{https://github.com/yicheng-w/CommonSenseMultiHopQA} & & 2018 \\
\hline \hline
\rowcolor{gray!12}\multicolumn{5}{|c|}{Ground text-enhanced methods} \\ 
\hline \hline
\multirow{8}*{\makecell[l]{Dialogue \\ system}} & \multirow{2}*{\cite{dinan2019wizard}} & Wizard of Wikipedia: Knowledge-Powered Conversational agents & \multirow{2}*{PyTorch} & ICLR \\
& & ------ Code: \url{https://github.com/facebookresearch/ParlAI} &  & 2019 \\
\cline{2-5}
& \multirow{2}*{\cite{qin2019conversing}} & Conversing by Reading: Contentful Neural Conversation with On-demand & \multirow{2}*{PyTorch} & ACL \\
& & Machine Reading ------ Code: \url{https://github.com/qkaren/converse_reading_cmr} & & 2019 \\
\cline{2-5}
& \multirow{2}*{\cite{kim2020sequential}} & Sequential Latent Knowledge Selection for Knowledge-Grounded Dialogue & \multirow{2}*{Tensorflow} & ICLR \\
& & ------ Code: \url{https://github.com/bckim92/sequential-knowledge-transformer} & & 2020 \\
\cline{2-5}
& \multirow{2}*{\cite{meng2020refnet}} & RefNet: A Reference-aware Network for Background Based  & \multirow{2}*{Tensorflow} & AAAI \\
& & Conversation ------ Code: \url{https://github.com/ChuanMeng/RefNet} & & 2020 \\
\hline \hline
\multirow{2}*{\makecell[l]{Summarization}} & \multirow{2}*{\cite{wang2019biset}} & BiSET: Bi-directional Selective Encoding with Template for & \multirow{2}*{PyTorch} & ACL \\
& & Abstractive Summarization ------ Code: \url{https://github.com/InitialBug/BiSET} & & 2019 \\
\hline \hline
\multirow{4}*{\makecell[l]{Question \\ answering}} & \multirow{2}*{\cite{lewis2020retrieval}} & Retrieval-Augmented Generation for Knowledge-Intensive NLP Tasks & \multirow{2}*{PyTorch} & Neurips \\
& & ------ Code in \url{https://github.com/huggingface/transformers} & & 2020 \\
\cline{2-5}
& \multirow{2}*{\cite{petroni2021kilt}} & KILT: a Benchmark for Knowledge Intensive Language Tasks & \multirow{2}*{PyTorch} & NAACL \\
& & ------ Code: \url{https://github.com/facebookresearch/KILT} & & 2021 \\
\hline
\end{tabular}}
\end{center}
\label{tab:code}
\end{table*}

% If your work has an appendix, this is the place to put it.

\end{document}